\documentclass{article} % For LaTeX2e
\usepackage[preprint]{arxiv/neurips}

\usepackage[utf8]{inputenc} % allow utf-8 input
\usepackage[T1]{fontenc}    % use 8-bit T1 fonts
\usepackage{hyperref}       % hyperlinks
\usepackage{url}            % simple URL typesetting
\usepackage{booktabs}       % professional-quality tables
\usepackage{amsfonts}       % blackboard math symbols
\usepackage{nicefrac}       % compact symbols for 1/2, etc.
\usepackage{microtype}      % microtypography
\usepackage{xcolor}         % colors

\usepackage{amsthm}
\theoremstyle{plain}
\newtheorem{theorem}{Theorem}[section]

\newtheorem{lemma}[theorem]{Lemma}

\theoremstyle{definition}

\theoremstyle{remark}
\newtheorem{remark}[theorem]{Remark}
\usepackage{mathtools}
\usepackage{subcaption}
\usepackage{cleveref}
\usepackage{enumitem}
\usepackage{multirow}
\usepackage{wrapfig}
\usepackage{xspace}
\usepackage{pifont}

\usepackage{algorithm}
\usepackage{algpseudocode}
\usepackage{listings}

\definecolor{codeblue}{RGB}{31, 119, 180}
\definecolor{codegray}{RGB}{80, 80, 80}
\definecolor{codered}{RGB}{180, 40, 40}
\definecolor{lightgray}{RGB}{248, 248, 248}
\definecolor{titlegray}{RGB}{30, 60, 100}
\definecolor{linkred}{RGB}{240, 0, 0}
\definecolor{ourswincolor}{HTML}{1B7837}
\definecolor{ourslosscolor}{HTML}{9A6D65}
\definecolor{sitcolor}{HTML}{236FA8}
\definecolor{contaminatedcolor}{HTML}{B52E17}
\newcommand{\ourswin}[1]{\textcolor{ourswincolor}{#1}}
\newcommand{\oursloss}[1]{\textcolor{ourslosscolor}{#1}}

\hypersetup{colorlinks=true, linkcolor=linkred, urlcolor=codeblue}

\newcommand{\model}[1]{\texttt{#1}\xspace}
\newcommand{\Moment}{\model{MOMENT}}
\newcommand{\PatchTST}{\model{PatchTST}}
\newcommand{\LagLLama}{\model{LagLLama}}
\newcommand{\TimeGPTone}{\model{TimeGPT-1}}
\newcommand{\TimeLLM}{\model{Time-LLM}}
\newcommand{\Moirai}{\model{Moirai}}
\newcommand{\Moiraitwo}{\model{Moirai-2.0}}

\newcommand{\Chronostwo}{\model{Chronos2}}

\newcommand{\TimesFMone}{\model{TimesFM}}
\newcommand{\TimesFMtwofive}{\model{TimesFM-2.5}}
\newcommand{\TimesFMthree}{\model{TimesFM-3}}
\newcommand{\TOTO}{\model{Toto}}
\newcommand{\TOTOtwo}{\model{Toto-2.0}}
\newcommand{\Sundial}{\model{Sundial}}
\newcommand{\Kairos}{\model{Kairos}}
\newcommand{\TiRex}{\model{TiRex}}
\newcommand{\TiRexTwo}{\model{TiRex-2}}
\newcommand{\NHITS}{\model{NHITS}}
\newcommand{\NBEATS}{\model{NBEATS}}
\newcommand{\YingLong}{\model{YingLong}}
\newcommand{\FalconTwo}{\model{Falcon 2}}
\newcommand{\Zeus}{\model{Zeus}}
\newcommand{\TimerSone}{\model{Timer-S1}}
\newcommand{\TimeMoE}{\model{Time-MoE}}
\newcommand{\TTM}{\model{TTM}}
\newcommand{\FlowState}{\model{FlowState}}
\newcommand{\PatchTSTFM}{\model{PatchTST-FM}}

\newcommand{\TimesFMtwo}{\model{TimesFM2.5}}

\newcommand{\ETS}{\model{ETS}}

\newcommand{\ARIMA}{\model{ARIMA}}
\newcommand{\SNaive}{\model{S.Naive}}

\newcommand{\ReVIN}{\model{ReVIN}}

\newcommand{\SIT}{\textcolor{sitcolor}{\textsc{\textbf{ScaleIn}}}\xspace}
\newcommand{\SITShort}{\textcolor{sitcolor}{\textsc{\textbf{Sc.In}}}\xspace}
\newcommand{\ours}{\SIT}
\newcommand{\oursShort}{\SITShort}

\newcommand{\contaminated}{%
  \textcolor{contaminatedcolor}{\textsc{\textbf{ScaleCon}}}\xspace}
\newcommand{\contaminatedShort}{%
  \textcolor{contaminatedcolor}{\textsc{\textbf{Sc.Con}}\xspace}}
\newcommand{\SCL}{\contaminated}

\newcounter{tsfmtotal}
\newcounter{tsfmpaperreported}
\newcounter{tsfmpaperinferred}
\newcounter{tsfmcodeinspected}
\newcounter{tsfmunverified}
\newcounter{tsfmverified}
\newcounter{tsfmcontaminated}
\newcounter{tsfminvariant}

\newcommand{\TSFMCount}[2]{%
  \stepcounter{tsfmtotal}%
  #1#2}
\newcommand{\TSFMReportedInPaper}{%
  \stepcounter{tsfmpaperreported}%
  \stepcounter{tsfmverified}}
\newcommand{\TSFMInferredFromPaper}{%
  \stepcounter{tsfmpaperinferred}%
  \stepcounter{tsfmverified}}
\newcommand{\TSFMRecoveredFromCode}{%
  \stepcounter{tsfmcodeinspected}%
  \stepcounter{tsfmverified}}
\newcommand{\TSFMUnverified}{\stepcounter{tsfmunverified}}
\newcommand{\TSFMScaleContaminated}{\stepcounter{tsfmcontaminated}}
\newcommand{\TSFMScaleInvariant}{\stepcounter{tsfminvariant}}
\newcommand{\TSFMLossUnverified}{}

\makeatletter
\newcommand{\TSFMStoreCount}[2]{%
  \begingroup
  \edef\@currentlabel{\arabic{#1}}%
  \label{#2}%
  \endgroup}
\makeatother

\newcommand{\TSFMStoreCounts}{%
  \TSFMStoreCount{tsfmtotal}{count:tsfm-total}%
  \TSFMStoreCount{tsfmpaperreported}{count:tsfm-paper-reported}%
  \TSFMStoreCount{tsfmpaperinferred}{count:tsfm-paper-inferred}%
  \TSFMStoreCount{tsfmcodeinspected}{count:tsfm-code-inspected}%
  \TSFMStoreCount{tsfmunverified}{count:tsfm-unverified}%
  \TSFMStoreCount{tsfmverified}{count:tsfm-verified}%
  \TSFMStoreCount{tsfmcontaminated}{count:tsfm-contaminated}%
  \TSFMStoreCount{tsfminvariant}{count:tsfm-invariant}}

\newcommand{\TSFMTotalCount}{\ref*{count:tsfm-total}}
\newcommand{\TSFMPaperReportedCount}{\ref*{count:tsfm-paper-reported}}
\newcommand{\TSFMPaperInferredCount}{\ref*{count:tsfm-paper-inferred}}
\newcommand{\TSFMCodeInspectedCount}{\ref*{count:tsfm-code-inspected}}
\newcommand{\TSFMUnverifiedCount}{\ref*{count:tsfm-unverified}}
\newcommand{\TSFMVerifiedCount}{\ref*{count:tsfm-verified}}
\newcommand{\TSFMContaminatedCount}{\ref*{count:tsfm-contaminated}}
\newcommand{\TSFMInvariantCount}{\ref*{count:tsfm-invariant}}

\newcommand{\dataset}[1]{\texttt{#1}\xspace}
\newcommand{\Mone}{\dataset{M1}}
\newcommand{\Mfour}{\dataset{M4}}

\newcommand{\Mthree}{\dataset{M3}}
\newcommand{\Tourism}{\dataset{Tourism}}

\newcommand{\Favorita}{\dataset{Favorita}}

\title{
Scale-Invariant Training \\ for Time Series Foundation Models}

\author{
Ignacy Stepka$^{1}$,
Willa Potosnak$^{1,2}$,
Kin G. Olivares$^{2,3}$, 
Artur Dubrawski$^{1}$ \\
$^{1}$Carnegie Mellon University \qquad
$^{2}$Amazon\thanks{All experiments were conducted in a controlled research setting and do not represent production deployments.} \qquad
$^{3}$Nixtla
}

\begin{document}

\maketitle

\begin{abstract}
Time series foundation models (TSFMs) are trained on large collections of time series datasets that span various morphologies and domains. This setting exposes models to series whose scales -- typical magnitudes of their values -- can differ substantially. Affine scaling methods such as Reversible Instance Normalization (ReVIN) scale model inputs and reverse the transform before computing the loss. We show that this inversion multiplies each series' gradient by $b^p$ relative to loss on scaled targets, where $b$ is the scaling denominator (e.g., standard deviation) and $p$ is the loss degree. We call this \emph{scale-contaminated training} (\contaminated), because the scale of each series consequently becomes an importance weight, causing high-scale series to dominate training. For any scale-equivariant scaler and residual loss that is homogeneous of degree $p$, including MSE, MAE, and Quantile Loss, we prove that computing loss on scaled targets makes every mini-batch gradient and, consequently, the full optimization trajectory invariant to arbitrary independent rescaling of the training series, yielding \emph{scale-invariant training} (\ours). Notably, existing TSFMs use both objectives, with neither consistent reporting nor a common convention on how to compute training loss. We isolate the convergence disparity induced by \contaminated and its correction under \ours in controlled studies on synthetic and real data. In pretraining across four TSFM architectures, \ours lowers MASE in all 24 architecture-benchmark comparisons, with average reductions across TSFMs of 18.8\% on GIFT-Eval and 21.9\% on the M-competitions. The gains extend to supervised neural forecasting, where it lowers MASE in 16 of 20 matched settings. Most existing time series forecasting pipelines can adopt \ours with a one-line code change.
\end{abstract}

\section{Introduction}

Scale differences across time series pose a challenge for any forecasting model. Individual series can exhibit substantial scale variation across sliding windows due to nonstationarity, level shifts, or trends, and, in datasets containing multiple series, across series themselves. This challenge is particularly relevant in Time Series Foundation Models (TSFMs)~\citep{garza2023timegpt1, das2024timesfm, ansari2025chronos2, salesforce2023moirai, goswami2024moment, cohen2025toto, vanness2023crossfrequencytimeseriesmetaforecasting}, where pretraining combines series from diverse domains, so scale differences across series within a training batch arise by construction and can span many orders of magnitude. A common approach to handling scale heterogeneity is to normalize each series using statistics computed from its observed history \citep{ulyanov2016instancenorm, salinas2020deepar, kim2022reversible}. A large body of work has proposed scaling methods for time series focused on \emph{what statistics to use} and \emph{how to compute them} (\ReVIN~\citep{kim2022reversible}, DAIN~\citep{passalis2019deep}, GAS-Norm~\citep{Urettini_2024}, Non-stationary Transformers~\citep{liu2023nonstationarytransformersexploringstationarity}).
An orthogonal design choice is \emph{where the scaling boundary sits relative to the loss}, that is, whether the loss is computed after inverse scaling of model predictions or after scaling the target. 
Prior empirical studies highlight related scale sensitivities. \citet{wolff2025spades} finds that models underperform on low-magnitude series, \citet{ahmed2025comparative} finds that \ReVIN's effectiveness depends on where the loss is computed, \citet{berthelier2026rolereversibleinstancenormalization} finds empirically that computing loss on scaled targets improves model performance, and \citet{olivares2025cftl} reveals accuracy struggles of TSFMs and traces them to scale heterogeneity.
While recent work has empirically compared these loss spaces, we formalize the consequences in training and show that this boundary determines whether training is \textbf{scale-invariant} or \textbf{scale-contaminated}.

\begin{figure}[t] %!ht
    \centering
    \includegraphics[width=.92\linewidth]{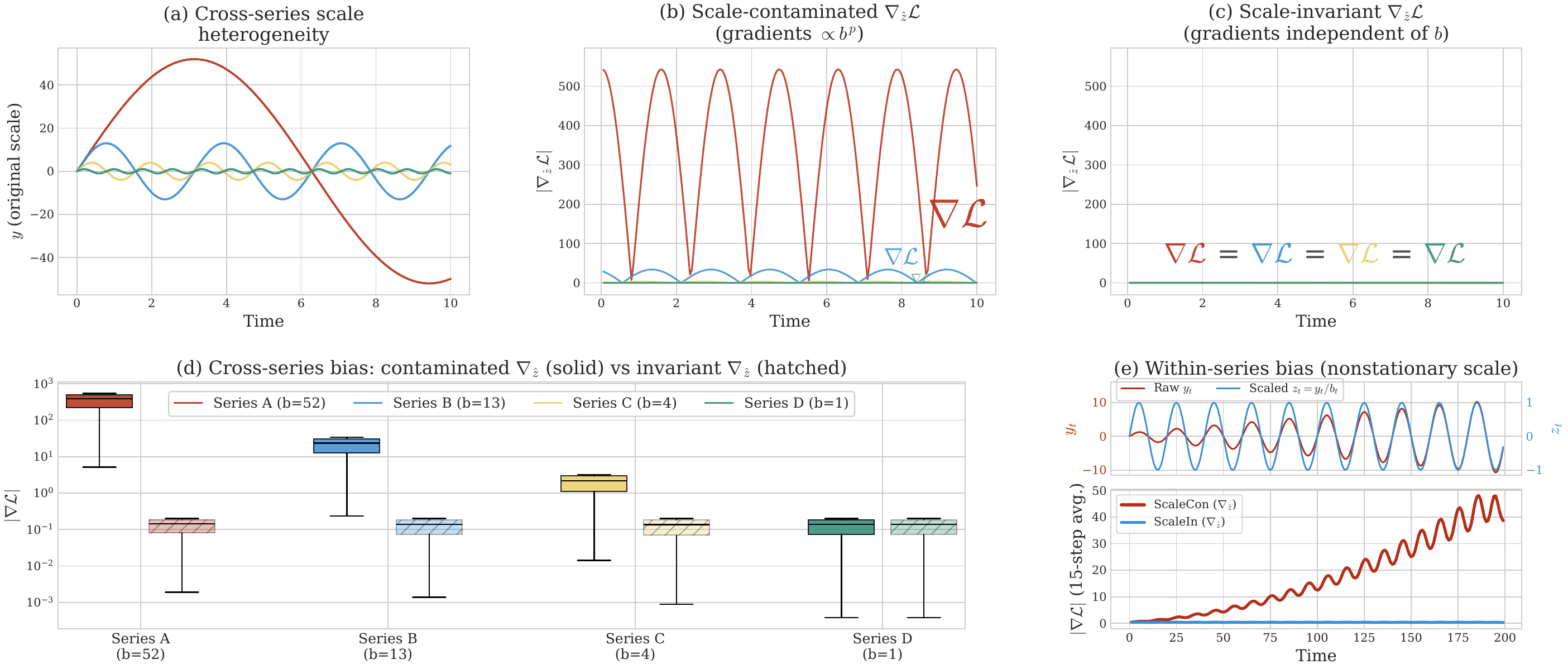}
    \caption{
    \ours removes scale-driven gradient bias (Theorem~\ref{thm:cross_series_invariance}).
    (a)~Synthetic series scales span $b\in\{1,4,13,52\}$.
    (b)~Under \contaminated, MSE gradients with respect to $\hat z$ scale as $b^2$, giving $b{=}52$ approximately $2{,}700\times$ the gradient magnitude of $b{=}1$ for equal scaled residuals.
    (c)~\ours leaves gradients unchanged by rescaling, on the same $y$-axis.
    (d)~Gradient distributions under \contaminated (solid) and \ours (hatched).
    (e)~Growing scale induces gradient drift under \contaminated but not \ours. Lower-row curves use a 15-step moving average of $|\nabla\mathcal{L}|$.
    } 
    \label{fig:motivation}
\end{figure}

When the loss is computed after rescaling predictions back to the original scale, as prescribed by \ReVIN~\citep{kim2022reversible}, the full series scale is reintroduced into the loss function. We refer to this phenomenon as scale-contaminated training (\contaminated). Under a degree-$p$ loss, each series' gradient is weighted by $b^p$, creating an implicit and potentially unintended importance weighting as shown in Fig.~\ref{fig:motivation} and discussed in \Cref{sec:forecasting_task}.
Nonstationarity within a single series causes this weighting to vary across forecast creation dates even in traditional single-series training.
In TSFMs, cross-dataset training compounds the effect across series within a batch.
In this work, we show that computing the loss directly on scaled targets yields Scale-Invariant Training (\ours) and improves forecast accuracy in TSFM pretraining and supervised forecasting.

Yet despite its impact on TSFMs, loss space is rarely discussed or documented in published work. Tables~\ref{tab:tsfm_comparison} and~\ref{tab:codebase-properties} show inconsistent reporting across existing TSFMs and popular forecasting codebases. Of the \TSFMTotalCount{} TSFMs surveyed (\Cref{tab:tsfm_comparison}), only
\TSFMPaperReportedCount{} explicitly document their loss space. For another
\TSFMPaperInferredCount{}, we infer the likely loss space by tracing how the paper
describes normalization, model outputs, and loss computation, although the paper
does not clearly establish it. Furthermore, inspecting the released code allows us to classify \TSFMCodeInspectedCount{} more models, while
\TSFMUnverifiedCount{} remain unresolved from either source. Among the \TSFMVerifiedCount{} models assigned a loss space from these sources, \TSFMContaminatedCount{} use \contaminated and \TSFMInvariantCount{} use \ours, showing that no prevailing convention exists.
Moreover, among popular forecasting codebases (\Cref{tab:codebase-properties}), only 2 support scale-invariant loss in some capacity but none make loss space user-configurable or document it explicitly. The practical change for training pipelines is one line of code that computes the loss on scaled targets instead of inverse scaling predictions, using existing context statistics (see Appendix~\ref{sec:apd_code_block}). We advocate for explicit reporting of loss space as a training design choice that affects convergence and downstream accuracy.

Our main contributions are as follows:
\begin{enumerate}[label=(\roman*)]
    \item We formally characterize the implicit $b^p$ weighting induced by inverse scaling and prove that Scale-Invariant Training (\ours) leaves every mini-batch gradient and the full optimization trajectory unchanged under arbitrary positive series-wise rescaling (Theorem~\ref{thm:cross_series_invariance}).

    \item We document inconsistent reporting of loss space across TSFM papers and forecasting codebases, establishing the need to report it explicitly as a design choice. The practical change is lightweight and computes the loss on scaled targets using existing context statistics.

    \item We validate scale-invariance theory in four settings using controlled synthetic data, a real-world scale-swap intervention, TSFM pretraining, and supervised training. These experiments provide empirical evidence that \ours removes scale-driven convergence disparities and improves forecast accuracy in both foundation model pretraining and supervised forecasting settings.
\end{enumerate}
 %\clearpage

\begin{table}[ht]
\centering
\caption{A survey of loss spaces used in time series foundation models. \textbf{Explicit}: the paper states the loss space explicitly in text or in equations that define the normalized or denormalized quantities entering the loss.  \textbf{Inferred}: it is possible to trace the target from the input through the normalization to the loss and infer with high likelihood what is the loss space. \textbf{Code}: the loss space is determined by the publicly released code. A dash means that we were not able to determine the loss space
(or loss function) from that source.}

\label{tab:tsfm_comparison}
\renewcommand{\arraystretch}{1.2}
\resizebox{0.95\textwidth}{!}{
\begin{tabular}{llllll}
\toprule
\textbf{Model}
    &
    & \textbf{Loss Function}
    & \textbf{Explicit}
    & \textbf{Inferred}
    & \textbf{Code} \\
\midrule
\TSFMCount{\TSFMUnverified}{\TSFMLossUnverified}
\TimeGPTone
    & \cite{garza2023timegpt1} & -- & -- & -- & -- \\
% confirmed with authors: \ours
\TSFMCount{\TSFMInferredFromPaper}{\TSFMScaleContaminated}
\LagLLama
    & \cite{rasul2023lag} & Student-t NLL & --
    & \contaminated$^{\P}$ & \contaminated$^{\P}$ \\

\TSFMCount{\TSFMInferredFromPaper}{\TSFMScaleContaminated}
\TimeLLM
    & \cite{jin2023time} & MSE & -- & \contaminated & \contaminated \\

\TSFMCount{\TSFMInferredFromPaper}{\TSFMScaleContaminated}
\Moirai
    & \cite{salesforce2023moirai} & Mixture NLL & --
    & \contaminated$^{\P}$ & \contaminated$^{\P}$ \\

\TSFMCount{\TSFMInferredFromPaper}{\TSFMScaleContaminated}
\Moment
    & \cite{goswami2024moment} & MSE & -- & \contaminated
    & \contaminated$^{\ddagger}$ \\

\TSFMCount{\TSFMInferredFromPaper}{\TSFMScaleContaminated}
\TimesFMone
    & \cite{das2024timesfm} & MSE & -- & \contaminated
    & \contaminated$^{*\ddagger}$ \\

\TSFMCount{\TSFMRecoveredFromCode}{\TSFMScaleContaminated}
\TimesFMtwofive
    & \cite{das2024timesfm} & Quantile & -- & -- & \contaminated$^{*}$ \\

\TSFMCount{\TSFMUnverified}{\TSFMLossUnverified}
\TimesFMthree
    & \cite{jain2026timesfm3} & Quantile  & -- & -- & -- \\

\TSFMCount{\TSFMInferredFromPaper}{\TSFMScaleContaminated}
\Moiraitwo
    & \cite{liu2025moirai2} & Quantile & -- & \contaminated & \ours$^{*}$ \\

\TSFMCount{\TSFMReportedInPaper}{\TSFMScaleInvariant}
\Chronostwo
    & \cite{ansari2025chronos2} & Quantile & \ours & \ours & \ours \\

\TSFMCount{\TSFMRecoveredFromCode}{\TSFMScaleInvariant}
\TOTO
    & \cite{cohen2025toto} & NLL + Cauchy & -- & -- & \ours$^{*}$ \\

\TSFMCount{\TSFMReportedInPaper}{\TSFMScaleInvariant}
\TOTOtwo
    & \cite{khwaja2026toto20timeseries} & Quantile & \ours & \ours
    & \ours$^{*}$ \\

\TSFMCount{\TSFMReportedInPaper}{\TSFMScaleContaminated}
\TiRex
    & \cite{auer2025tirex} & Quantile & \contaminated & \contaminated & -- \\

\TSFMCount{\TSFMReportedInPaper}{\TSFMScaleContaminated}
\TiRexTwo
    & \cite{podest2026tirex2} & Quantile & \contaminated & \contaminated
    & \contaminated$^{*}$ \\

\TSFMCount{\TSFMRecoveredFromCode}{\TSFMScaleInvariant}
\Kairos
    & \cite{feng2025kairos} & Weighted quantile & -- & -- & \ours \\

\TSFMCount{\TSFMUnverified}{\TSFMLossUnverified}
\Sundial
    & \cite{liu2025sundial} & TimeFlow & -- & -- & -- \\

\TSFMCount{\TSFMReportedInPaper}{\TSFMScaleContaminated}
\YingLong
    & \cite{wang2025yinglong} & Weighted quantile & \contaminated$^{\dagger}$
    & \contaminated$^{\dagger}$ & \contaminated$^{\dagger}$ \\

\TSFMCount{\TSFMReportedInPaper}{\TSFMScaleInvariant}
\FalconTwo
    & \cite{xia2026falcon} & Quantile & \ours & \ours & -- \\

\TSFMCount{\TSFMRecoveredFromCode}{\TSFMScaleInvariant}
\Zeus
    & \cite{fu2026zeus} & Quantile & -- & -- & \ours$^{*}$ \\

\TSFMCount{\TSFMReportedInPaper}{\TSFMScaleContaminated}
\TimerSone
    & \cite{liu2026timers1} & Weighted quantile & \contaminated$^{\dagger}$
    & \contaminated$^{\dagger}$ & -- \\

\TSFMCount{\TSFMRecoveredFromCode}{\TSFMScaleContaminated}
\TimeMoE
    & \cite{shi2025timemoe} & Huber & -- & -- & \contaminated$^{\S}$ \\

\TSFMCount{\TSFMReportedInPaper}{\TSFMScaleContaminated}
\TTM
    & \cite{ekambaram2024ttms} & MSE & \contaminated & \contaminated
    & \contaminated \\

\TSFMCount{\TSFMReportedInPaper}{\TSFMScaleContaminated}
\FlowState
    & \cite{graf2025flowstate} & Quantile & \contaminated & \contaminated & -- \\

\TSFMCount{\TSFMInferredFromPaper}{\TSFMScaleInvariant}
\PatchTSTFM
    & \cite{wen2026revisitingtransformer} & Quantile & -- & \ours & \ours \\

\bottomrule
\end{tabular}
}

\vspace{2pt}
\begin{minipage}{0.95\textwidth}
\footnotesize
$^{*}$Inferred from released model, fine-tuning, or inference code, as the pretraining
code is not released.
$^{\P}$If NLL is the loss function, the original-space loss differs from
the normalized one only by a constant $\log\sigma$ which does not affect gradients.
$^{\dagger}$Pinball loss divided by the mean absolute target.
$^{\ddagger}$Additionally standardization globally per dataset and channel.
$^{\S}$Series are globally standardized, with no per-window scaling.
\end{minipage}

\TSFMStoreCounts
\end{table}

\begin{table}[ht]
\centering
\caption{Training properties of time series forecasting codebases.}
\label{tab:codebase-properties}
\renewcommand{\arraystretch}{1.5}
\resizebox{0.95\textwidth}{!}{
\begin{tabular}{llp{5cm}p{5cm}c}
\toprule
\textbf{Codebase} & \textbf{Reference} & \contaminated & \textbf{Invariant (\ours)} & \textbf{Fixed/User-specified} \\
\midrule
GluonTS
    & \cite{alexandrov2020gluonts}
    & NLL, CRPS, EnergyScore
    & ---
    & Fixed \\[6pt]
PyTorch Forecasting
    & \cite{beitner2020pytorch_forecasting}
    & MAE, RMSE, MAPE, SMAPE,
      QuantileLoss, NormalDist,
      LogNormal, NegBinomial,
      Beta, MQF2
    & ---
    & Fixed \\[6pt]
TSLib
    & \cite{wu2023tslib}
    & ---
    & MSE, MAPE, MASE, SMAPE
    & Fixed \\[6pt]
NeuralForecast
    & \cite{olivares2022neuralforecast}
    & NLL (Normal, StudentT,
      NegBinomial, Poisson,
      Tweedie)
    & MAE, MSE, RMSE, MAPE,
      SMAPE, MASE,
      QuantileLoss, MQLoss
    & Fixed \\
\bottomrule
\end{tabular}
}
\end{table}

\section{Related Work}
Training time series models on heterogeneous data requires accounting for series whose scales differ by orders of magnitude. The most common approach is to apply instance normalization \citep{ulyanov2016instancenorm, salinas2020deepar, kim2022reversible} to scale inputs using normalization statistics computed over the context window. Frequently cited in TSFM papers is Reversible Instance Normalization (\ReVIN)~\citep{kim2022reversible}, which scales each context window inside the model's forward pass and restores predictions to the original scale before computing the loss.
However, these TSFMs differ in where they compute the loss (see the summary in \Cref{tab:tsfm_comparison}). 
One group~\citep{rasul2023lag, goswami2024moment, jin2023time, salesforce2023moirai, auer2025tirex, das2024timesfm, liu2026timers1, graf2025flowstate} reverses the scaling before computing the loss, as prescribed by \ReVIN. 
Another group~\citep{ansari2025chronos2,cohen2025toto,khwaja2026toto20timeseries,xia2026falcon} scales the targets and computes a scale-invariant loss.
DAIN \citep{passalis2019deep} learns adaptive input shifts, scales, and feature gates.
Other approaches including Non-stationary Transformers~\citep{liu2023nonstationarytransformersexploringstationarity}, GAS-Norm~\citep{Urettini_2024}, and \citet{vilhes2026normalizationchoice}, differ in how they estimate scaling statistics but, like \ReVIN, restore predictions before computing the loss.

Recent empirical studies examine some of these design choices. \citet{ahmed2025comparative} compare scaling schemes across TSFM architectures, including a comparison of loss computed on scaled targets with clipping and loss computed after returning predictions to the original scale. \citet{berthelier2026rolereversibleinstancenormalization} ablate the components of \ReVIN on two supervised forecasters across four benchmark datasets. They observe that computing loss on scaled targets improves accuracy and hypothesize that the benefit comes from equal weighting of low- and high-scale series. We formally characterize this mechanism for homogeneous residual losses, deriving the scale-dependent gradient weights and establishing conditions for invariance of the full optimization trajectory. Moreover, our controlled experiments isolate the effect of loss space.
On the empirical side, we evaluate TSFM pretraining and supervised forecasting on the M1, M3, M4, Tourism, Favorita, and GIFT-Eval datasets~\citep{makridakis1982m1_competition,makridakis2000m3_competition,makridakis2020m4_competition,athanosopoulus2011tourism_competition,favorita,aksu2024giftevalbenchmarkgeneraltime}. Our synthetic studies isolate the $b^p$ mechanism, while pretraining and supervised experiments test its practical consequences (\Cref{sec:experiments}). These results establish why the loss space should be deliberately chosen and reported explicitly in forecasting papers as a design choice.

% \newpage
\section{Forecasting Task and Scale-Invariant Training Losses}\label{sec:forecasting_task}

We consider supervised univariate forecasting with real-valued targets.
At forecast creation date $t$, the model receives a context $\mathbf{y}_{t-L+1:t}\in\mathbb{R}^{L}$, where $L$ is the supported context length, and predicts the next $H$ values $\mathbf{y}_{t+1:t+H}\in\mathbb{R}^{H}$.
Before the model sees the context, a scaling (or normalization) method computes an offset $a$ and a positive denominator $b$ from observations available by time $t$ and transforms each value as:
\begin{equation}
z_t=\frac{y_t-a}{b}.
\label{eq:affine_scaling}
\end{equation}
After that, the model maps the scaled context $\mathbf{z}_{t-L+1:t}$ to a scaled prediction $\mathbf{\hat z}_{t+1:t+H}$,
\begin{equation}
\hat{\mathbf{z}}_{t+1:t+H}=f_{\boldsymbol{\theta}}\left(\mathbf{z}_{t-L+1:t}\right).
\label{eq:forecast_task}
\end{equation}
In this work, we study residual losses of the form $\mathcal{L}(y,\hat y)=\phi(y-\hat y)$ that satisfy $\phi(br)=b^p\phi(r)$ for $b>0$.
The three cases used throughout this manuscript and popular in related work are:
\begin{equation}
\begin{aligned}
\mathrm{MAE}(y,\hat y) &= |y-\hat y|, & p&=1, \\
\mathrm{MSE}(y,\hat y) &= (y-\hat y)^2, & p&=2, \\
\mathrm{QL}_q(y,\hat y) &= q(y-\hat y)_+ +(1-q)(\hat y-y)_+, & p&=1.
\end{aligned}
\label{eq:common_residual_losses}
\end{equation}
The central choice is whether in training the loss is computed before or after returning the prediction to the original scale.
The default pipeline with \ReVIN~\citep{kim2022reversible} first reverses the transform,
\begin{equation}
\hat{y}_{t+h}=b\hat{z}_{t+h}+a,
\label{eq:revin_rescaling}
\end{equation}
and then compares $\hat y_{t+h}$ with $y_{t+h}$.
We call this objective scale-contaminated (\contaminated),
\begin{equation}
\mathcal{L}_{\mathrm{contam}}
=\frac{1}{H}\sum_{h=1}^{H}
\phi\left(y_{t+h}-(b\hat{z}_{t+h}+a)\right).
\label{eq:scale_contaminated}
\end{equation}
The alternative is to scale the observed target with the same $a$ and $b$,
\begin{equation}
z_{t+h}=\frac{y_{t+h}-a}{b}, \qquad h=1,\ldots,H.
\label{eq:scaled_target}
\end{equation}
Note that the future observations do not contribute to $a$ or $b$.
As we show below, computing the loss before reversing the transform gives scale-invariant loss (\ours),
\begin{equation}
\mathcal{L}_{\mathrm{inv}}
=\frac{1}{H}\sum_{h=1}^{H}
\phi\left(z_{t+h}-\hat{z}_{t+h}\right).
\label{eq:scale_inv}
\end{equation}

\begin{figure}[t] %!ht
    \centering
    \includegraphics[width=\linewidth]{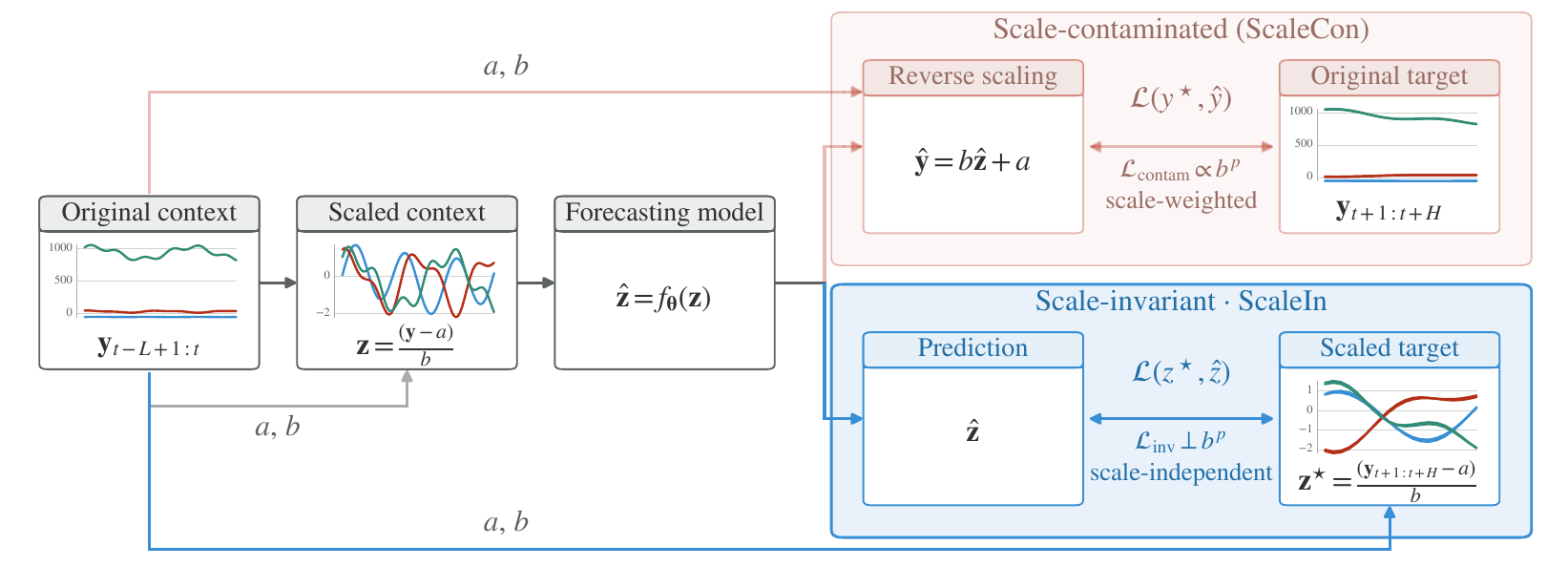}
    \caption{
    Loss placement determines whether inverse scaling introduces $b^p$ gradient weighting. Both pipelines scale inputs using context statistics $a,b$.
    \textbf{Top:} \contaminated reverses scaling before computing the loss, introducing the $b^p$ weight.
    \textbf{Bottom:} \ours computes loss on scaled targets and reverses scaling only for inference, yielding scale-invariant training (Theorem~\ref{thm:cross_series_invariance}).}
    \label{fig:method}
\end{figure}

\begin{lemma}[Inverse scaling creates importance weights]
\label{lem:scale_contamination}
For every forecasting example, the scale-contaminated and scale-invariant objectives satisfy
\begin{align}
\mathcal{L}_{\mathrm{contam}}&=b^p\mathcal{L}_{\mathrm{inv}}, \label{eq:loss_space_identity} \\
\nabla_{\boldsymbol{\theta}}\mathcal{L}_{\mathrm{contam}}&=b^p\nabla_{\boldsymbol{\theta}}\mathcal{L}_{\mathrm{inv}}. \label{eq:gradient_space_identity}
\end{align}
The gradient relation also holds for subgradients at nondifferentiable residuals.
\end{lemma}

\noindent Proof in Appendix~\ref{apd:proofs}.
For a mini-batch $\mathcal{B}$, the resulting update is therefore
\begin{equation}
\nabla_{\boldsymbol{\theta}}\mathcal{L}_{\mathrm{contam},\mathcal{B}}=\frac{1}{|\mathcal{B}|}\sum_{i\in\mathcal{B}}b_i^p\nabla_{\boldsymbol{\theta}}\mathcal{L}_{\mathrm{inv},i}.
\label{eq:contaminated_batch_gradient}
\end{equation}
The denominator $b_i$ is an importance weight inside every gradient descent update.
For instance, the weight is $b_i$ for MAE and quantile loss and $b_i^2$ for MSE.
Because $b_i$ varies across series and forecast creation dates, scale-contaminated training changes the relative importance of forecasting examples even when their scaled forecasting problems are identical.
\Cref{fig:motivation} visualizes the MSE gradient gap, and the left panel of \Cref{fig:apd:adam:grad-global} confirms the $b^2$ pattern at initialization.  In \Cref{fig:method} we visualize the difference between \contaminated and \ours in a typical forecasting pipeline.

\bigskip

\begin{theorem}[Scale-Invariant Training]
\label{thm:cross_series_invariance}
Assume that the scaling method is scale equivariant, meaning that multiplying a series by any $c>0$ multiplies its offset $a$ and denominator $b$ by the same $c$.
Under scale-invariant loss, independently multiplying each training series by any positive constant leaves the loss and parameter gradient of every forecasting example unchanged for all $\boldsymbol{\theta}$.
Consequently, it leaves every mini-batch gradient and the full parameter trajectory unchanged when the initialization, mini-batch sequence, optimizer state, and optimizer randomness are fixed.
\end{theorem}

\noindent Proof in Appendix~\ref{apd:proofs}.
The scale-equivariance assumption holds for commonly used scaling methods such as standardization ($\frac{x-\mu}{\sigma}$), min-max scaling ($\frac{x - \text{min}}{\text{max}-\text{min}}$), and robust scaling ($\frac{x-m}{\text{IQR}}$), all with positive denominators and without fixed numerical stabilizers. 
Thus, under \ours, series' scale cannot change which forecasting examples dominate training. 
Fixed additive $\epsilon$ terms or scale floors can break exact equivariance, so practical invariance may be approximate. Importantly, they do not remove the $b^p$ weighting in Lemma~\ref{lem:scale_contamination}, which holds for the actual denominator used in the affine transform.

\bigskip
\begin{remark}[Practical intuition]
Consider two series with identical scaled contexts and targets, but denominators $b=1$ and $b=10$ -- for instance, assume standardization with $b$ corresponding to the standard deviation.
If the model makes the same scaled error on both, for the second series, \contaminated yields a gradient that is $10$ times larger under MAE or pinball loss and $100$ times larger under MSE.
The optimizer treats the second series as more important solely because of its raw scale.
Under \ours, both series produce the same gradient.
Note, however, that even under scale invariance, different forecasting problems can still produce different gradients and achieve nonuniform convergence due to inherent data difficulty.
\end{remark}

Some TSFMs (e.g., \Chronostwo) additionally apply an arcsinh transform after standardization. We note that scale contamination persists when this transform is reversed, but the nonlinearity changes the residual geometry, so the two objectives are not related solely by $b^p$. Appendix~\ref{apd:nonlinear-scaling} gives the corresponding identity and explains why \ours remains scale-invariant under equivariance.

Although stated for univariate point forecasting, our result extends componentwise to multivariate objectives that sum homogeneous residual losses across channels~\citep{liu2024itransformer, ekambaram2024ttms}.
It also covers quantile forecasting because pinball loss is homogeneous of degree one at every quantile level~\citep{taylor2000quantile, ansari2025chronos2}, as detailed in Appendix~\ref{apd:forecasting_extensions}. We include an algorithm implementation of \ours in Section~\ref{sec:apd_code_block}. For clarity, we note that predictions $\hat z$ return to the original scale for inference and evaluation.

\section{Experiments}\label{sec:experiments}

We evaluate \ours in 4 complementary settings, from controlled isolation of the mechanism (Sections~\ref{sec:exp_synthetic} and \ref{sec:exp_scale_swap}) to TSFM pretraining and supervised forecasting (Sections~\ref{sec:exp_tsfm} and \ref{sec:exp_supervised}).

\begin{figure}[t!] 
    \centering
    \includegraphics[width=0.5\linewidth]{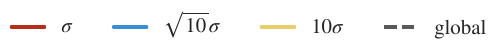}\par
    \begin{subfigure}[t]{0.276\linewidth}
        \centering
        \includegraphics[width=\linewidth]{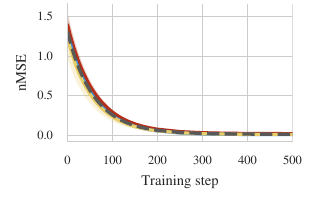}
        \caption{Scale-invariant}
        \label{fig:synthetic:normalized}
    \end{subfigure}\hfill
    \begin{subfigure}[t]{0.236\linewidth}
        \centering
        \includegraphics[width=\linewidth]{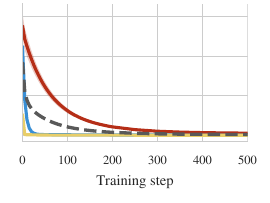}
        \caption{Scale-contaminated}
        \label{fig:synthetic:original}
    \end{subfigure}\hfill
    \begin{subfigure}[t]{0.236\linewidth}
        \centering
        \includegraphics[width=\linewidth]{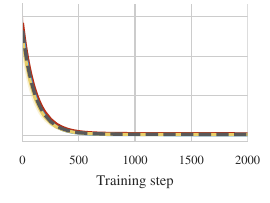}
        \caption{Equal-scale control}
        \label{fig:synthetic:original_equalvar}
    \end{subfigure}\hfill
    \begin{subfigure}[t]{0.236\linewidth}
        \centering
        \includegraphics[width=\linewidth]{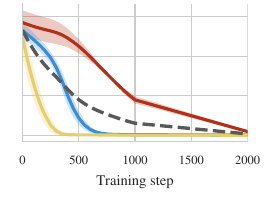}
        \caption{Unit gradient}
        \label{fig:synthetic:original_gradmatch}
    \end{subfigure}
    \caption{Per-source validation nMSE over training steps. We report the mean $\pm1$ standard deviation over five seeds. \textbf{(a)}~The scale-invariant loss yields parallel convergence. \textbf{(b)}~The scale-contaminated loss produces scale-ordered convergence. \textbf{(c)}~The disparity disappears when all sources share the same scale. \textbf{(d)}~The disparity persists after matching the global gradient norm. The dashed grey curve is global nMSE, aggregated rather than disaggregated by source.}
    \label{fig:loss_space_comp:synthetic}
\end{figure}

\subsection{Controlled Synthetic Data Experiment}\label{sec:exp_synthetic}

% In this experiment, we isolate the $b^p$ gradient bias in standard \ReVIN and expose the disparate convergence per-source it produces. The synthetic data set consists of three sinusoidal sources with scaling denominators in the approximate ratio $1\!:\!\sqrt{10}\!:\!10$. Under MSE loss function, their loss values and gradient multipliers are $b^2$ and follow the ratio $1\!:\!10\!:\!100$. We train a 2-layer \PatchTST-style Transformer with \ReVIN for 30K SGD steps, fixing the random seed, parameter initialization, and the examples seen during optimization. 
% The only knob that varies is whether the model learns on a \emph{scale-contaminated} or a \emph{scale-invariant} loss. 
In this experiment, we expose the disparate convergence per-series resulting from $b^p$ gradient bias in standard \ReVIN and demonstrate how \ours fixes this. To isolate the effect in a simple controlled setting, we generate a synthetic data set consisting of three sinusoidal sources with scaling denominators in an approximate ratio $1\!:\!\sqrt{10}\!:\!10$. As shown in Lemma~\ref{lem:scale_contamination}, \ReVIN's denormalization under MSE loss introduces $b^2$ scaling of loss values and gradient multipliers, yielding the ratio $1\!:\!10\!:\!100$ across the three sources. This scale-dependent gradient should cause the optimizer to disproportionately reduce error on high-scale series and lead to disparate convergence rates across sources. To test this, we train a 2-layer \PatchTST-style Transformer with \ReVIN for 30K SGD steps, fixing the random seed, parameter initialization, and the examples seen during optimization. With all model and training aspects held fixed, the only experimental parameter that varies is whether the model learns on a \emph{scale-contaminated} or a \emph{scale-invariant} loss. In the synthetic and scale-swap experiments (Section \ref{sec:exp_scale_swap}), we report normalized mean squared error (nMSE), computed between predictions and targets normalized using each window's context mean and standard deviation, then averaged over forecast timesteps and validation windows. Both loss conditions use this metric so errors are comparable across scales.
The results show that under \contaminated, the high-scale source converges $50\times$ faster than the low-scale source (Figure~\ref{fig:synthetic:original}). In comparison, with \ours in Figure~\ref{fig:synthetic:normalized}, all three scale sources converge in parallel at a similar pace. This confirms that the $b^p$ gradient bias characterized by Lemma~\ref{lem:scale_contamination} does indeed produce disparate per-series convergence, as hypothesized. Appendix~\ref{apd:synthetic-reproducibility} provides the full experimental settings.

In addition, we controlled the experiments to better attribute the gap to $b^p$ and eliminate confounders. First, to isolate the effect of series scale, we give all sources unit amplitude and train again with the scale-contaminated loss (Figure~\ref{fig:synthetic:original_equalvar}). The gap vanishes compared with Figure~\ref{fig:synthetic:normalized}, confirming that it stems from scale heterogeneity rather than intrinsic source difficulty. Second, to eliminate gradient-magnitude confounders, we rescale the global gradient to unit norm (Figure~\ref{fig:synthetic:original_gradmatch}). The convergence gap persists, matching the dynamics of  \contaminated training in Figure~\ref{fig:synthetic:original}. Because $b^2$ scaling is applied item-wise within a batch, a single global scalar cannot fix the between-source ratio: it rescales all per-item gradients uniformly and preserves their relative magnitudes. These findings also hold under Adam~\citep{kingma2014adam}, where the gap narrows from approximately $50\times$ to $10\times$, consistent with adapting gradient magnitude but not its per-source composition (Appendix~\ref{apd:adam-loss-space}).

\begin{figure}[t]
    \centering
    \includegraphics[width=0.4\linewidth]{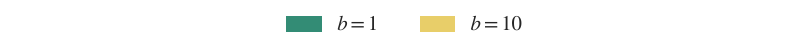}
    \hspace{5cm}
    \begin{subfigure}[t]{0.48\linewidth}
        \centering
        \includegraphics[width=\linewidth]{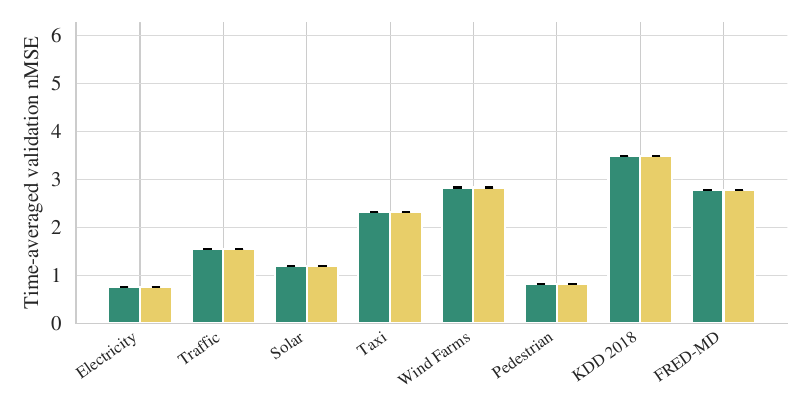}
        \caption{Scale-invariant loss (\ours)}
        \label{fig:loss-space-scale-swap-main:normalized}
    \end{subfigure}
    \hfill
    \begin{subfigure}[t]{0.48\linewidth}
        \centering
        \includegraphics[width=\linewidth]{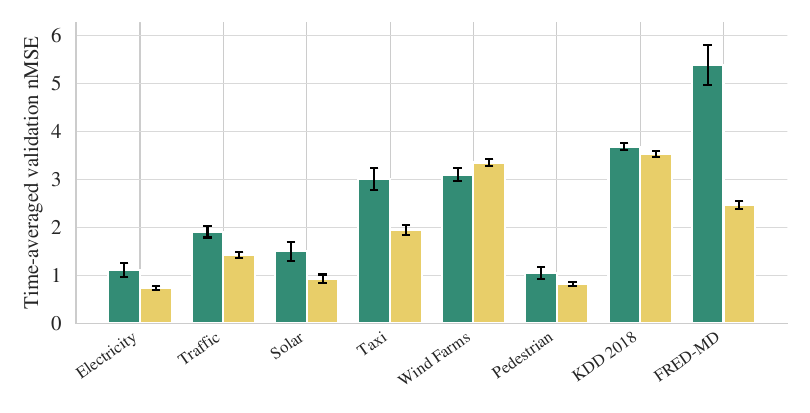}
        \caption{Scale-contaminated loss (\contaminated)}
        \label{fig:loss-space-scale-swap-main:lr-adjusted}
    \end{subfigure}
    \caption{
    Inflating scale leaves trajectory-level error nearly unchanged under \ours~\textbf{(a)}, but lowers it on average under \contaminated~\textbf{(b)}. Bars show $A^{(k)}_{b,c}$ averaged over five seeds and 15 complementary assignment pairs. Error bars show 95\% Student-$t$ intervals across assignment pairs.
    }
    \label{fig:loss-space-scale-swap-main}
\end{figure}

\subsection{Scale-Swap Intervention Experiment}\label{sec:exp_scale_swap}

In this experiment, we confirm the effect of loss-space gradient bias on disparate per-series convergence using real-world data. We selectively rescale series $10\times$ as a diagnostic: under \ours, Theorem~\ref{thm:cross_series_invariance} predicts rescaling leaves convergence unchanged, while under \contaminated, the $b^p$ bias should cause the optimizer to favor the inflated series. We selected $K=8$ datasets from different domains in GluonTS~\citep{alexandrov2020gluonts} (Electricity, Traffic, Solar, Taxi, Wind Farms, Pedestrian, KDD Cup 2018, FRED-MD) and applied global standard scaling ($\frac{x-\mu}{\sigma}$). Then, for each dataset $D^{(k)}$ we create a matched pair $(D^{(k)}_{b=1},\, D^{(k)}_{b=10})$: the original series at scale $b{=}1$ and a copy multiplied by $b{=}10$. Because both copies of a dataset differ only in scale, any convergence difference between them is attributable to the $b^p$-scaled gradient bias rather than to intrinsic difficulty.

%\paragraph{Matched-pair scale-augmented corpus.}

\begin{figure}[t]
    \centering
    \begin{subfigure}[t]{0.42\linewidth}
        \centering
        \includegraphics[width=\linewidth]{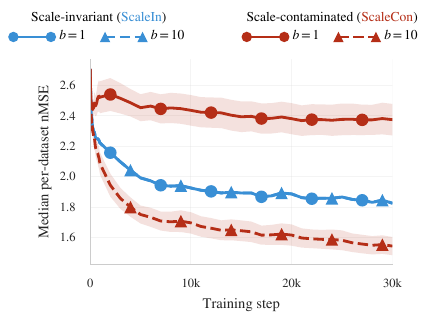}
        \caption{Scale-swap corpus, $K{=}8$ datasets}
        \label{fig:tsfm-natural-mase-convergence:scaleswap}
    \end{subfigure}\hfill
    \begin{subfigure}[t]{0.5\linewidth}
        \centering
        \includegraphics[width=\linewidth]{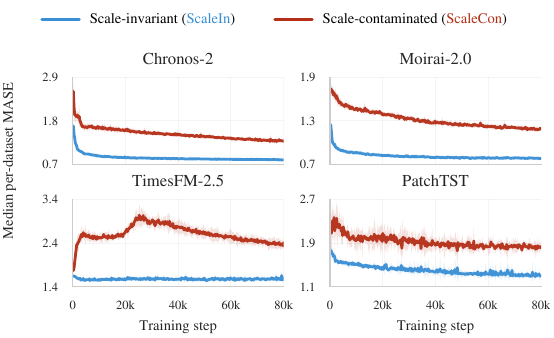}
        \caption{Full TSFM pretraining, four architectures}
        \label{fig:tsfm-natural-mase-convergence:pretrain}
    \end{subfigure}
    \caption{
    \ours removes scale-driven convergence disparities and improves TSFM convergence. \textbf{(a)}~Median per-dataset nMSE over 8 scale-swap datasets, aggregated across 15 assignment pairs and five seeds. The $b{=}1$ and $b{=}10$ curves overlap under \ours and separate under \contaminated.
    \textbf{(b)}~Median per-dataset MASE for four architectures over $80{,}000$ updates. Plotted are averages across 4 seeds, with 95\% Student-$t$ intervals.
    }
    \label{fig:tsfm-natural-mase-convergence}
\end{figure}

We construct 15 configurations
$D_c = \{D^{(k)}_b\}_{k=1}^{8}$, each assigning four datasets to $b{=}1$ and
four to $b{=}10$, plus a complementary run that reverses every assignment.
Across the $30$ resulting runs, every dataset appears 15 times at each scale, and we average over configurations to minimize the impact of batch composition. 
We use the \PatchTST-style Transformer from Section~\ref{sec:exp_synthetic} with four layers instead of two, trained under scale-invariant and scale-contaminated losses with SGD on $S=5$ seeds per condition.
Because scale-contaminated MSE multiplies the gradient of a $b{=}10$ window by $100$ while leaving a $b{=}1$ window unchanged, the two conditions cannot share a learning rate without confounding the diagnostic with raw update magnitude. We therefore divide the contaminated learning rate by the nominal average multiplier of a balanced batch, $(1+100)/2=50.5$, giving $10^{-4}$ for the scale-invariant condition and $1.9802\times10^{-6}$ for the scale-contaminated one. This approximately compensates for the overall gradient amplification while preserving the relative $100\!:\!1$ weighting \emph{within} each contaminated batch, which is the quantity our diagnostic tests.
The comparison of interest is in either case within a condition, between its own $b{=}1$ and $b{=}10$ copies.
Figure~\ref{fig:tsfm-natural-mase-convergence:scaleswap} shows the paired convergence trajectories, and the weighted-average analysis below quantifies it. Appendix~\ref{apd:scale-swap-reproducibility} provides the full experimental settings.

To validate our hypothesis from Section~\ref{sec:exp_synthetic} on real-world data, we compare each dataset's average validation nMSE over training when assigned $b{=}1$ versus $b{=}10$. We compute this average as the area under the validation curve divided by training duration, then average over 5 seeds and 15 complementary assignment pairs. Figure~\ref{fig:loss-space-scale-swap-main} shows that inflating scale leaves this error essentially unchanged under \ours, with a mean difference across datasets below $0.002\%$. Under \contaminated, the mean error falls from $2.6$ to $1.9$, a $26.8\%$ reduction, and a one-sided paired $t$-test across the dataset-level differences is significant at $\alpha=0.05$. This shows that the $b^p$ gradient bias of Lemma~\ref{lem:scale_contamination} extends to real-world data. Full metric definitions, test statistics, and dataset-level trajectories are in Appendix~\ref{apd:scale-swap-details}.

\subsection{Time Series Foundation Model Pretraining}\label{sec:exp_tsfm}

In this experiment, we empirically analyze the difference between using \ours and \contaminated in TSFM pretraining. We employ 5 forecasting benchmarks drawn from well-known forecasting competitions: \Mone~\citep{makridakis1982m1_competition}, \Mthree~\citep{makridakis2000m3_competition}, \Mfour~\citep{makridakis2020m4_competition}, \Favorita~\citep{favorita},  and \Tourism~\citep{athanosopoulus2011tourism_competition}. We additionally evaluate on GIFT-Eval, a popular benchmark in the TSFM literature~\citep{aksu2024giftevalbenchmarkgeneraltime}.
We adopt the data handling and pre-processing practices established in prior work leveraging heterogeneous-dataset pretraining~\citep{alexandrov2020gluonts,olivares2025cftl,potosnak2025forkingsequences}. 
Benchmark training prefixes exclude reserved validation and test regions (Appendix~\ref{apd:tsfm-details}).

We pretrain four architectures for $80{,}000$ updates at batch size $512$, $4$ seeds each, on GiftEvalPretrain~\citep{aksu2024giftevalbenchmarkgeneraltime} with the \emph{training regions} of M1, M3, M4, Tourism, and Favorita included. Two are recent foundation models in their small configurations, \Chronostwo{} ($27.9$M parameters, \citet{ansari2025chronos2}) and \Moiraitwo{} ($5.0$M parameters, \citet{liu2025moirai2}). The third is \TimesFMtwo{} ($49.6$M parameters, \citet{das2024timesfm}), and the fourth is a supervised \PatchTST{} ($1.5$M parameters, \citet{nie2023patchtst}). These architectures span a representative range of recent TSFMs in terms of scale ($1.5$M--$49.6$M parameters) and design (decoder-only and encoder-only). \PatchTST was not originally proposed as a foundation model, but recent work has shown its effectiveness in multi-dataset pretraining~\citep{wen2026revisitingtransformer}. Its inclusion highlights that this work is applicable to any general forecaster, particularly those trained across heterogeneous datasets. Each model keeps the temporal scaling method of its own published implementation.
Within each model, \ours and \SCL runs see identical data in the same order and share initialization, optimization schedule, and learning rate, with no per-condition adjustment. AdamW normalizes each coordinate by the running root-mean-square of its own gradient, so a uniform rescaling of the objective leaves its updates unchanged (note, it does not wash out within batch item-wise $b^p$ scaling). 

We score the final checkpoint on one official test horizon per series for M1, M3, M4, and Tourism, 16 day holdout for Favorita, and GIFT-Eval's non-overlapping short-term windows across 55 configurations. Models receive up to 512 historical observations, with shorter contexts padded and masked. Within each benchmark, we average scores between frequency groups (dataset configurations for GIFT-Eval), weighted by the number of forecasts evaluated, then average across 4 training seeds (Appendix~\ref{apd:tsfm-details}).
We include Seasonal Na\"ive (\SNaive) as a reference, which repeats the last observed season using the benchmark's seasonal period and repeats the last observation when fewer than two seasons are available~\citep{hyndman2026forecasting}.

We report two scale-independent metrics. The Mean Absolute Scaled Error (MASE) ~\citep{hyndman2006another} normalizes the forecast error by the in-sample seasonal naive error,
\begin{equation}
\mathrm{MASE} = \frac{1}{H}\sum_{h=1}^{H} \frac{|y_{t+h} - \hat{y}_{t+h}|}{\frac{1}{T-m}\sum_{j=m+1}^{T}|y_j - y_{j-m}|},
\end{equation}
where $m$ is the seasonal period. The Weighted Quantile Loss (WQL) averages twice the quantile loss over the quantile levels $\mathcal{Q}$ supplied by each model,
\begin{equation}
\mathrm{WQL} = \frac{2}{|\mathcal{Q}|}\sum_{q \in \mathcal{Q}} \frac{\sum_{h=1}^{H} \mathrm{QL}_q(y_{t+h},\, \hat{y}^{(q)}_{t+h})}{\sum_{h=1}^{H} |y_{t+h}|},
\end{equation}
with $\mathrm{QL}_q$ defined in Section~\ref{sec:forecasting_task}. For models that do not supply quantile forecasts, we treat the point prediction as the median and compute WQL with $\mathcal{Q}=\{0.5\}$.

% Generated by src/scripts/render_paper_tables.py. Do not edit by hand.
\begin{table}[t!]
\centering
\caption{Forecasting performance across benchmark datasets. The results were averaged across 4 seeds. \ours uses scale-invariant loss and \SCL uses scale-contaminated loss. These are the only differences between the two runs of each architecture. $\Delta\%$ reports \ours relative to \SCL. \ourswin{Green} denotes an improvement and \oursloss{red} a regression. The better loss is bold within each architecture. References are fitted on the series they forecast and appear behind the double rule.}
\label{tab:tsfm_heldout_indomain-main}
\renewcommand{\arraystretch}{1.15}
\scriptsize
\setlength{\tabcolsep}{3pt}
\begin{tabular}{l|ccc|ccc|ccc|ccc||c}
\toprule
 & \multicolumn{3}{c}{\Chronostwo} & \multicolumn{3}{c}{\Moiraitwo} & \multicolumn{3}{c}{\TimesFMtwo} & \multicolumn{3}{c}{\PatchTST} & \multicolumn{1}{c}{Reference} \\
\cmidrule(lr){2-4}\cmidrule(lr){5-7}\cmidrule(lr){8-10}\cmidrule(lr){11-13}\cmidrule(lr){14-14}
\textbf{Benchmark} & \oursShort & \contaminatedShort & $\Delta\%$ & \oursShort & \contaminatedShort & $\Delta\%$ & \oursShort & \contaminatedShort & $\Delta\%$ & \oursShort & \contaminatedShort & $\Delta\%$ & \SNaive \\
\midrule
\multicolumn{14}{l}{\emph{MASE}} \\
M1 & \textbf{1.879} & 2.388 & \ourswin{-21.3} & \textbf{2.013} & 2.123 & \ourswin{-5.2} & \textbf{2.731} & 2.863 & \ourswin{-4.6} & \textbf{2.091} & 3.859 & \ourswin{-45.8} & 2.117 \\
M3 & \textbf{1.608} & 1.963 & \ourswin{-18.1} & \textbf{1.621} & 1.800 & \ourswin{-9.9} & \textbf{2.233} & 2.573 & \ourswin{-13.2} & \textbf{2.041} & 4.840 & \ourswin{-57.8} & 1.764 \\
M4 & \textbf{1.845} & 2.294 & \ourswin{-19.6} & \textbf{1.879} & 1.986 & \ourswin{-5.4} & \textbf{2.661} & 3.208 & \ourswin{-17.0} & \textbf{2.235} & 4.555 & \ourswin{-50.9} & 2.057 \\
Tourism & \textbf{2.350} & 2.816 & \ourswin{-16.5} & \textbf{2.316} & 2.564 & \ourswin{-9.7} & \textbf{3.820} & 4.108 & \ourswin{-7.0} & \textbf{2.509} & 3.123 & \ourswin{-19.7} & 2.412 \\
Favorita & \textbf{0.816} & 0.862 & \ourswin{-5.4} & \textbf{0.825} & 0.864 & \ourswin{-4.5} & \textbf{1.050} & 1.360 & \ourswin{-22.8} & \textbf{0.889} & 0.968 & \ourswin{-8.2} & 1.079 \\
GIFT-Eval & \textbf{1.495} & 1.782 & \ourswin{-16.1} & \textbf{1.547} & 1.676 & \ourswin{-7.7} & \textbf{2.162} & 2.687 & \ourswin{-19.5} & \textbf{1.902} & 2.813 & \ourswin{-32.4} & 1.824 \\
\midrule
\multicolumn{14}{l}{\emph{WQL}} \\
M1 & 0.133 & \textbf{0.119} & \oursloss{+11.7} & 0.174 & \textbf{0.164} & \oursloss{+6.6} & \textbf{0.208} & 0.241 & \ourswin{-13.7} & \textbf{0.227} & 0.284 & \ourswin{-20.2} & 0.150 \\
M3 & \textbf{0.084} & 0.088 & \ourswin{-4.7} & \textbf{0.109} & 0.114 & \ourswin{-4.4} & \textbf{0.158} & 0.176 & \ourswin{-10.6} & \textbf{0.153} & 0.253 & \ourswin{-39.3} & 0.110 \\
M4 & \textbf{0.084} & 0.089 & \ourswin{-5.7} & 0.110 & \textbf{0.107} & \oursloss{+2.9} & \textbf{0.161} & 0.183 & \ourswin{-11.7} & \textbf{0.147} & 0.229 & \ourswin{-36.0} & 0.115 \\
Tourism & \textbf{0.086} & 0.113 & \ourswin{-24.1} & \textbf{0.116} & 0.132 & \ourswin{-12.0} & \textbf{0.227} & 0.266 & \ourswin{-14.8} & \textbf{0.149} & 0.200 & \ourswin{-25.5} & 0.123 \\
Favorita & \textbf{0.297} & 0.326 & \ourswin{-9.1} & \textbf{0.387} & 0.422 & \ourswin{-8.3} & \textbf{0.638} & 0.848 & \ourswin{-24.8} & \textbf{0.540} & 0.595 & \ourswin{-9.3} & 0.614 \\
GIFT-Eval & \textbf{0.285} & 0.320 & \ourswin{-10.9} & \textbf{0.371} & 0.397 & \ourswin{-6.5} & \textbf{0.569} & 0.681 & \ourswin{-16.4} & \textbf{0.518} & 0.597 & \ourswin{-13.3} & 0.705 \\
\bottomrule
\end{tabular}
\end{table}

Table~\ref{tab:tsfm_heldout_indomain-main} reports MASE and WQL across all benchmarks for all architectures at the final checkpoint. \ours outperforms \contaminated in terms of lower MASE across all 24 architecture-benchmark comparisons, with  error reductions (averaged across architectures) between 10.2\% and 24.1\%. On WQL it wins 21 of 24 comparisons, with average performance gains between 3.3\% and 19.1\%. The improvement is consistent regardless of the normalization method: \TimesFMtwo computes statistics from an expanding causal window, \Chronostwo, \Moiraitwo, and \PatchTST from the context window, with \Chronostwo additionally applying an arcsinh transformation.

In an ablation study with~\Moment~\citep{goswami2024moment}, we also find that the benefit extends beyond forecasting objectives: \ours substantially lowers the reconstruction MASE of \Moment, trained on a masked reconstruction task (Appendix~\ref{apd:moment-reconstruction}). The benchmark-frequency breakdowns, standard deviations, and cross-dataset variance trajectories
can be found in Appendix~\ref{apd:tsfm-details}.

We hypothesize that these accuracy improvements stem from more stable and scale-unbiased optimization. Figure~\ref{fig:tsfm-natural-mase-convergence:pretrain} follows the median per-dataset MASE throughout the $80{,}000$ optimization steps. The \ours develops an advantage during training for all four architectures and retains it at the final checkpoint. Additionally, we observe that \contaminated loss introduces excess difficulty and \TimesFMtwo is the clearest case of that. Its scale-contaminated run degrades for the first $25{,}000$ steps before recovering, while its scale-invariant run is flat from the start.

\subsection{Supervised Neural Forecasting Models}
\label{sec:exp_supervised}

The experiments in Section~\ref{sec:exp_tsfm} establish that the scale-contamination effect persists under TSFM-scale, multi-dataset pretraining, affecting forecast error on both in-distribution and zero-shot benchmarks.
Beyond the TSFM pretraining setting, we show that forecast error improvements with \ours extend to standard deep forecasters trained across heterogeneous datasets on a smaller scale, with supervised evaluation only. We train 5 established architectures, \NHITS \citep{challu2023nhits}, \NBEATS \citep{oreshkin2020NBeats}, \PatchTST \citep{nie2023patchtst}, \model{TCN} \citep{bai2018TCN}, and \model{LSTM} \citep{hochreiter1997lstm}, on M1, M3, M4, and Tourism, fitting 6 separate models, one for each sampling frequency (i.e., yearly, monthly, quarterly, weekly, daily, and hourly) by aggregating training splits across datasets. All neural networks runs share data splits, Adam optimizer ($\text{lr}=10^{-3}$), and a $20{,}000$-step budget across 4 seeds, selecting checkpoints via minimum validation MAE. See Appendix~\ref{apd:supervised-breakdown} for complete data, standard deviation, and aggregation details. For reference, we compare with classical statistical baselines \SNaive, \ETS, and \ARIMA~\citep{hyndman2026forecasting}.

% Generated by src/scripts/render_updated_m_supervised_appendix.py. Do not edit by hand.
\begin{table}[t]
\centering
\caption{Forecasting performance by benchmark and architecture. For each seed, we select the checkpoint with the lowest validation MAE. Within each frequency, we average MASE across forecasts and compute WQL as a ratio of sums. We then macro-average both metrics across frequencies and report their means across 4 seeds. \ours uses scale-invariant loss and \SCL uses scale-contaminated loss. Lower is better, and \textbf{bold} marks the better loss within each architecture.}
\label{tab:supervised-mseries-benchmark}
\scriptsize
\setlength{\tabcolsep}{2.5pt}
\resizebox{\textwidth}{!}{%
\begin{tabular}{ll|cc|cc|cc|cc|cc||ccc}
\toprule
 &  & \multicolumn{10}{c||}{\textbf{Neural Forecast Baselines}} & \multicolumn{3}{c}{\textbf{Reference}} \\
\cmidrule(lr){3-12}\cmidrule(l){13-15}
\multirow{2}{*}{\textbf{Benchmark}} & \multirow{2}{*}{\textbf{Metric}} & \multicolumn{2}{c}{\model{N-HiTS}} & \multicolumn{2}{c}{\model{N-BEATS}} & \multicolumn{2}{c}{\PatchTST} & \multicolumn{2}{c}{\model{TCN}} & \multicolumn{2}{c}{\model{LSTM}} & \multirow{2}{*}{\SNaive} & \multirow{2}{*}{\ETS} & \multirow{2}{*}{\ARIMA} \\
\cmidrule(lr){3-4}\cmidrule(lr){5-6}\cmidrule(lr){7-8}\cmidrule(lr){9-10}\cmidrule(lr){11-12}
 &  & \oursShort & \contaminatedShort & \oursShort & \contaminatedShort & \oursShort & \contaminatedShort & \oursShort & \contaminatedShort & \oursShort & \contaminatedShort &  &  &  \\
\midrule
\multirow{2}{*}{M1} & MASE & $\textbf{1.826}$ & $1.855$ & $\textbf{1.822}$ & $1.835$ & $1.917$ & $\textbf{1.911}$ & $\textbf{1.941}$ & $1.987$ & $\textbf{1.998}$ & $2.057$ & $2.478$ & $1.904$ & $1.977$ \\
 & WQL & $0.099$ & $\textbf{0.098}$ & $0.099$ & $\textbf{0.098}$ & $\textbf{0.106}$ & $\textbf{0.106}$ & $\textbf{0.100}$ & $0.101$ & $\textbf{0.105}$ & $0.106$ & $0.167$ & $0.108$ & $0.132$ \\
\midrule
\multirow{2}{*}{M3} & MASE & $1.652$ & $\textbf{1.640}$ & $\textbf{1.677}$ & $1.767$ & $\textbf{2.802}$ & $3.083$ & $\textbf{1.896}$ & $2.060$ & $\textbf{2.133}$ & $2.134$ & $2.011$ & $1.477$ & $1.687$ \\
 & WQL & $\textbf{0.090}$ & $0.091$ & $\textbf{0.089}$ & $0.092$ & $\textbf{0.116}$ & $0.123$ & $\textbf{0.099}$ & $0.103$ & $0.109$ & $\textbf{0.107}$ & $0.103$ & $0.087$ & $0.090$ \\
\midrule
\multirow{2}{*}{M4} & MASE & $\textbf{1.991}$ & $2.007$ & $\textbf{1.987}$ & $2.036$ & $3.639$ & $\textbf{3.572}$ & $\textbf{2.090}$ & $2.198$ & $\textbf{2.264}$ & $2.268$ & $2.446$ & $2.030$ & $2.351$ \\
 & WQL & $\textbf{0.072}$ & $\textbf{0.072}$ & $\textbf{0.072}$ & $0.073$ & $\textbf{0.093}$ & $0.093$ & $\textbf{0.075}$ & $0.076$ & $\textbf{0.080}$ & $0.081$ & $0.083$ & $0.071$ & $0.079$ \\
\midrule
\multirow{2}{*}{Tourism} & MASE & $\textbf{2.346}$ & $2.420$ & $\textbf{2.307}$ & $2.392$ & $\textbf{2.884}$ & $2.960$ & $\textbf{2.668}$ & $2.688$ & $2.643$ & $\textbf{2.614}$ & $2.232$ & $2.259$ & $2.503$ \\
 & WQL & $\textbf{0.133}$ & $0.137$ & $\textbf{0.132}$ & $0.135$ & $\textbf{0.164}$ & $0.167$ & $\textbf{0.153}$ & $\textbf{0.153}$ & $0.153$ & $\textbf{0.152}$ & $0.142$ & $0.133$ & $0.156$ \\
\bottomrule
\end{tabular}%
}
\end{table}

Table~\ref{tab:supervised-mseries-benchmark} shows that the scale-invariant loss lowers MASE in 16 of 20 matched architecture-benchmark comparisons and WQL in 15 of 20. At the finer benchmark-frequency level, \ours lowers MASE in 49 of 75 comparisons and WQL in 43 of 75, with the complete breakdown and standard deviations reported in Appendix~\ref{apd:supervised-breakdown}. Averaged over all architectures, benchmarks, and seeds, \ours lowers MASE by $1.9\%$ and WQL by $1.1\%$ The effect is therefore broad, but rather modest, and it varies between architectures and benchmarks. The fitted statistical references remain competitive and lead the neural models on several benchmark-metric pairs.  %\clearpage
\section{Conclusion}\label{sec:conclusion}

In this work, we show that reversing an affine scaling before computing a degree-$p$ homogeneous loss injects a $b^p$ factor into the gradient that grows with series scale, imposing an implicit curriculum that favors high-scale series. 
The remedy is a one-line code change: compute the loss directly on scaled targets. We confirmed this across the full spectrum from controlled synthetic studies to real-world pretraining. 
Scale-invariant loss (\ours) closes the convergence gap on synthetic data and is invariant to artificial scale assignments on real data (Sections~\ref{sec:exp_synthetic} and~\ref{sec:exp_scale_swap}).
Employing \ours in TSFM pre-training leads on average to substantial improvements in MASE and WQL performance metrics, reaching 18.8\% on GIFT-Eval and 21.9\% on M-competitions (Section~\ref{sec:exp_tsfm}). The effect on neural forecasting models is broadly positive, but expectedly modest (Section~\ref{sec:exp_supervised}).
Based on the results, we thus posit that loss space is a crucial training decision -- especially for TSFMs -- that should be made deliberately and reported explicitly in any forecasting paper to promote reproducibility.

Our findings motivate a few directions for future work.
While the same scale bias persists in multivariate settings with independent scaling of each series, its practical impact and the effects of more general normalization schemes that jointly transform correlated channels remain to be studied.
Another direction is to characterize how loss-space choices interact with probabilistic objectives and their parameterizations, beyond the homogeneous losses studied here.

\bibliographystyle{abbrvnat}
\bibliography{references}

\appendix
\newpage
\section{Scale-Invariance and Contamination
in the TSFM Open-Source Ecosystem}

Training loss space is a fundamental design choice that directly affects what a model learns, yet it is not consistently documented in papers or codebases. Tables~\ref{tab:tsfm_comparison} and~\ref{tab:codebase-properties} survey the loss space used by prominent TSFMs and open-source forecasting codebases respectively. Across the \TSFMTotalCount{} models in the survey,
we distinguish explicit reporting from conclusions drawn indirectly from the
available sources. The papers of \TSFMPaperReportedCount{} models explicitly
specify the loss space. For \TSFMPaperInferredCount{} others, the descriptions of
normalization, outputs, and the objective suggest a likely loss space without
establishing it explicitly. Inspecting released code provides a classification
for a further \TSFMCodeInspectedCount{} models, leaving \TSFMUnverifiedCount{}
unresolved. Under these classifications, \TSFMInvariantCount{} models use \ours.
The survey thus identifies both scale-contaminated and scale-invariant training
in current practice, with substantial gaps in how these choices are reported.

 We classify each model from its paper and its public code separately. A paper counts as \emph{explicit} if it states the loss space in text, or if its equations define the normalized or denormalized quantities that enter the loss. A paper counts as \emph{inferred} if it gives no such statement, but its description of the normalization, the model outputs, and the loss lets us trace the target from the input to the loss and infer the loss space with high likelihood. We note, that citing RevIN ~\citep{kim2022reversible} or writing the loss on a generic target $y$ is not enough on its own to infer either \ours or \contaminated. In code column, we mark TSFMs that released code which allows to determine where the loss is computed, and we flag cases where this follows from model, fine-tuning, or inference code rather than a released pretraining loop.

Furthermore, prominent open-source codebases do not explicitly document loss space choice, nor support user specification of the loss space. The lack of documentation raises reproducibility concerns because practitioners who train, fine-tune or extend models may inadvertently change the loss space without realizing it, producing results that are not directly comparable to the original. More broadly, without awareness of the implications, practitioners may simply propagate the use of scale-contaminated losses, which we show has consequences for training convergence and downstream accuracy in cross-frequency training scenarios. We hypothesize that the lack of proper loss space reporting is a result of the widespread adoption of instance normalization schemes such as RevIN, which (by default) denormalize predictions before returning them. However, as we show in this work with both theoretical and empirical evidence, loss space selection is an important design choice.  
\section{Scale-Contamination and Scale-Invariance Proofs} \label{apd:proofs}

\paragraph{Proof of Lemma~\ref{lem:scale_contamination}.}
For one series contribution, the observed and predicted targets on the original scale are $y_{t+h}=bz_{t+h}+a$ and $\hat{y}_{t+h}=b\hat{z}_{t+h}+a$.
Their residual is therefore
\begin{equation}
y_{t+h}-\hat{y}_{t+h}=b\left(z_{t+h}-\hat{z}_{t+h}\right).
\label{eq:proof_residual_scaling}
\end{equation}
Substituting Equation~\ref{eq:proof_residual_scaling} into the scale-contaminated objective and applying $\phi(br)=b^p\phi(r)$ at every horizon step gives
\begin{align}
\mathcal{L}_{\mathrm{contam}}
&=\frac{1}{H}\sum_{h=1}^{H}\phi\left(b\left(z_{t+h}-\hat{z}_{t+h}\right)\right) \\
&=\frac{b^p}{H}\sum_{h=1}^{H}\phi\left(z_{t+h}-\hat{z}_{t+h}\right) \\
&=b^p\mathcal{L}_{\mathrm{inv}}.
\end{align}
The scaling offset $a$ cancels, while the scaling denominator remains as the multiplicative factor $b^p$.
The statistics $a$ and $b$ depend only on the observed history, so they are constant with respect to the model parameters $\boldsymbol{\theta}$.
Differentiating the loss identity therefore yields
\begin{equation}
\nabla_{\boldsymbol{\theta}}\mathcal{L}_{\mathrm{contam}}=b^p\nabla_{\boldsymbol{\theta}}\mathcal{L}_{\mathrm{inv}}.
\end{equation}
When $\phi$ is nondifferentiable, multiplication by the positive constant $b^p$ scales every valid subgradient by the same factor.
For a mini-batch $\mathcal{B}$, linearity of summation then gives
\begin{equation}
\nabla_{\boldsymbol{\theta}}\mathcal{L}_{\mathrm{contam},\mathcal{B}}=\frac{1}{|\mathcal{B}|}\sum_{i\in\mathcal{B}}b_i^p\nabla_{\boldsymbol{\theta}}\mathcal{L}_{\mathrm{inv},i},
\end{equation}
which is Equation~\ref{eq:contaminated_batch_gradient}. \hfill$\square$

\paragraph{Proof of Theorem~\ref{thm:cross_series_invariance}.}
Consider multiplying every observation of one series by an arbitrary constant $c>0$.
Write $y_s^{(c)}=cy_s$ for the rescaled observation.
Scale equivariance gives $a^{(c)}=ca$ and $b^{(c)}=cb$, so every scaled context and target value satisfies
\begin{equation}
z_s^{(c)}=\frac{y_s^{(c)}-a^{(c)}}{b^{(c)}}=\frac{cy_s-ca}{cb}=\frac{y_s-a}{b}=z_s.
\label{eq:proof_scaled_value_invariance}
\end{equation}
The model consequently receives the same scaled context and produces the same scaled prediction for every $\boldsymbol{\theta}$,
\begin{equation}
\hat{\mathbf{z}}_{t+1:t+H}^{(c)}=f_{\boldsymbol{\theta}}\left(\mathbf{z}_{t-L+1:t}^{(c)}\right)=f_{\boldsymbol{\theta}}\left(\mathbf{z}_{t-L+1:t}\right)=\hat{\mathbf{z}}_{t+1:t+H}.
\label{eq:proof_prediction_invariance}
\end{equation}
Equations~\ref{eq:proof_scaled_value_invariance} and~\ref{eq:proof_prediction_invariance} make the scale-invariant objective identical before and after the rescaling,
\begin{align}
\mathcal{L}_{\mathrm{inv}}^{(c)}(\boldsymbol{\theta})
&=\frac{1}{H}\sum_{h=1}^{H}\phi\left(z_{t+h}^{(c)}-\hat{z}_{t+h}^{(c)}\right) \\
&=\frac{1}{H}\sum_{h=1}^{H}\phi\left(z_{t+h}-\hat{z}_{t+h}\right) \\
&=\mathcal{L}_{\mathrm{inv}}(\boldsymbol{\theta}).
\end{align}
Because this equality holds as a function of $\boldsymbol{\theta}$, the parameter gradients are also identical,
\begin{equation}
\nabla_{\boldsymbol{\theta}}\mathcal{L}_{\mathrm{inv}}^{(c)}(\boldsymbol{\theta})=\nabla_{\boldsymbol{\theta}}\mathcal{L}_{\mathrm{inv}}(\boldsymbol{\theta}).
\label{eq:proof_gradient_invariance}
\end{equation}
The rescaling constants may differ across series because Equation~\ref{eq:proof_gradient_invariance} holds for every $c>0$ separately.
If series $i$ is multiplied by $c_i>0$, the scale-invariant mini-batch gradient therefore satisfies
\begin{equation}
\nabla_{\boldsymbol{\theta}}\mathcal{L}_{\mathrm{inv},\mathcal{B}}^{(\mathbf{c})}=\frac{1}{|\mathcal{B}|}\sum_{i\in\mathcal{B}}\nabla_{\boldsymbol{\theta}}\mathcal{L}_{\mathrm{inv},i}^{(c_i)}=\frac{1}{|\mathcal{B}|}\sum_{i\in\mathcal{B}}\nabla_{\boldsymbol{\theta}}\mathcal{L}_{\mathrm{inv},i}=\nabla_{\boldsymbol{\theta}}\mathcal{L}_{\mathrm{inv},\mathcal{B}}.
\label{eq:proof_batch_gradient_invariance}
\end{equation}
For contrast, the same constant leaks directly into a scale-contaminated gradient.
After rescaling, the scaling denominator in the inverse transform is $b^{(c)}=cb$, so Lemma~\ref{lem:scale_contamination} gives
\begin{align}
\nabla_{\boldsymbol{\theta}}\mathcal{L}_{\mathrm{contam}}^{(c)}
&=(b^{(c)})^p\nabla_{\boldsymbol{\theta}}\mathcal{L}_{\mathrm{inv}}^{(c)} \\
&=(cb)^p\nabla_{\boldsymbol{\theta}}\mathcal{L}_{\mathrm{inv}} \\
&=c^p\nabla_{\boldsymbol{\theta}}\mathcal{L}_{\mathrm{contam}}.
\label{eq:proof_c_leaks_into_gradient}
\end{align}
Thus \ours cancels $c$ before the loss is evaluated, while a scale-contaminated loss multiplies the same series contribution by $c^p$.
For the same independent rescalings, the scale-contaminated mini-batch gradient becomes
\begin{equation}
\nabla_{\boldsymbol{\theta}}\mathcal{L}_{\mathrm{contam},\mathcal{B}}^{(\mathbf{c})}=\frac{1}{|\mathcal{B}|}\sum_{i\in\mathcal{B}}(c_i b_i)^p\nabla_{\boldsymbol{\theta}}\mathcal{L}_{\mathrm{inv},i},
\label{eq:proof_batch_c_leak}
\end{equation}
Comparing Equations~\ref{eq:proof_batch_gradient_invariance} and~\ref{eq:proof_batch_c_leak} shows that only the scale-contaminated update changes with the chosen $c_i$ values.
Fix the initialization, mini-batch sequence, optimizer state, and optimizer randomness as stated in the theorem.
At the first step, Equation~\ref{eq:proof_batch_gradient_invariance} gives both scale-invariant runs the same gradient, so the optimizer produces the same next parameters and state.
Induction over training steps proves that the complete scale-invariant optimization trajectory is unchanged. \hfill$\square$

\subsection{Extensions to Multivariate and Quantile Forecasting}
\label{apd:forecasting_extensions}

\paragraph{Multivariate forecasting.}
Suppose the target has $D$ channels and channel $j$ uses scaling offset $a_j$ and denominator $b_j$.
For a loss $\phi$ homogeneous of degree $p$, the scale-invariant and scale-contaminated objectives are
\begin{align}
\mathcal{L}_{\mathrm{inv}}^{\mathrm{multi}}
&=\frac{1}{HD}\sum_{h=1}^{H}\sum_{j=1}^{D}\phi\left(z_{t+h,j}-\hat{z}_{t+h,j}\right), \\
\mathcal{L}_{\mathrm{contam}}^{\mathrm{multi}}
&=\frac{1}{HD}\sum_{h=1}^{H}\sum_{j=1}^{D}b_j^p\phi\left(z_{t+h,j}-\hat{z}_{t+h,j}\right).
\end{align}
The derivation applies Lemma~\ref{lem:scale_contamination} separately to each channel.
Multiplying channel $j$ by any $c_j>0$ leaves its scaled values unchanged under a channel-wise scale-equivariant scaling method, so the proof of Theorem~\ref{thm:cross_series_invariance} applies to the complete scaled input and target tensors.
Cross-channel dependencies inside the forecasting model do not affect this conclusion because the model receives the same scaled tensor after rescaling.

\paragraph{Quantile forecasting.}
Let $\mathcal{Q}$ be a set of quantile levels and let $\hat{z}_{t+h}^{(q)}$ be the scaled prediction for level $q$.
Writing the pinball loss as $\mathrm{QL}_q(r)=q(r)_+ +(1-q)(-r)_+$, the quantile objective is
\begin{equation}
\mathcal{L}_{\mathrm{inv}}^{\mathrm{quant}}=\frac{1}{H|\mathcal{Q}|}\sum_{h=1}^{H}\sum_{q\in\mathcal{Q}}\mathrm{QL}_q\left(z_{t+h}-\hat{z}_{t+h}^{(q)}\right).
\end{equation}
This objective averages pinball loss across all $H$ horizon steps and all requested quantile levels, so the model learns the quantile forecasts jointly.
For every $b>0$, pinball loss satisfies $\mathrm{QL}_q(br)=b\mathrm{QL}_q(r)$.
The same $b$ applies to every quantile level of a series, which gives
\begin{equation}
\mathcal{L}_{\mathrm{contam}}^{\mathrm{quant}}=b\mathcal{L}_{\mathrm{inv}}^{\mathrm{quant}},
\qquad
\nabla_{\boldsymbol{\theta}}\mathcal{L}_{\mathrm{contam}}^{\mathrm{quant}}=b\nabla_{\boldsymbol{\theta}}\mathcal{L}_{\mathrm{inv}}^{\mathrm{quant}}.
\end{equation}
Computing pinball loss on scaled targets removes this factor jointly across all quantile levels, so Theorem~\ref{thm:cross_series_invariance} applies without changing the model or the quantile set.

\subsection{Nonlinear Scaling in \Chronostwo}
\label{apd:nonlinear-scaling}
Write $u=\operatorname{arcsinh}((y-a)/b)$ and let $\hat u$ be the model output. Reversing the transform gives $\hat y=b\sinh(\hat u)+a$, so
\begin{equation}
    \phi(y-\hat y)
    = b^p\phi\bigl(\sinh(u)-\sinh(\hat u)\bigr).
\end{equation}
The $b^p$ factor still weights the loss and parameter gradient, but the remaining loss differs from $\phi(u-\hat u)$ because of the nonlinear inverse. This comparison therefore changes residual geometry as well as scale weighting. Under scale-equivariant statistics, independently rescaling a series leaves $u$ unchanged, so training on transformed targets retains scale invariance. Fixed numerical stabilizers can qualify this invariance as discussed in Section~\ref{sec:forecasting_task}.
 
\section{Algorithm}\label{sec:apd_code_block}

Algorithm~\ref{alg:sil} contrasts the two training paradigms. Both begin identically: a scaling offset $a$ and denominator $b$ are computed from the context window, the context is normalized, and the model predicts in scaled space. The two approaches diverge only in how the loss is computed. In scale-contaminated training, predictions are denormalized back to the original scale before the loss is applied, coupling the gradient magnitude to 
$b^p$ and thereby making optimization sensitive to the absolute scale of each series. In scale-invariant training, the targets are instead normalized to match the prediction space, and the loss is computed entirely in scaled space, making the gradient independent of $b$. The key observation is that the difference between the two paradigms reduces to a single step: whether the targets are denormalized predictions or normalized targets are passed to the loss function, a change that amounts to one line of code in practice.

\begin{algorithm}[ht]
\caption{Scale-\contaminated and scale-invariant (\ours) objectives}
\label{alg:sil}
\begin{algorithmic}[1]
\Require Context $\mathbf{y}_{t-L+1:t}$, targets $\mathbf{y}_{t+1:t+H}$, model $f_{\boldsymbol{\theta}}$, loss $\phi$, scaling method $(a, b)$
\State Compute offset $a$ and denominator $b$ from context
\State Scale context: $z_{t-l} \gets (y_{t-l} - a) / b$ for $l = 0, \ldots, L-1$
\State Predict scaled outputs: $\hat{\mathbf{z}}_{t+1:t+H} \gets f_{\boldsymbol{\theta}}\left(\mathbf{z}_{t-L+1:t}\right)$
\vspace{0.5em}
\State \textbf{// Scale-\contaminated objective}
\State Reverse scaling: $\hat{y}_{t+h} \gets b\hat{z}_{t+h} + a$ for $h = 1, \ldots, H$
\State $\mathcal{L}_{\mathrm{contam}} \gets \frac{1}{H}\sum_{h=1}^{H} \phi\left(y_{t+h} - \hat{y}_{t+h}\right)$ \Comment{Loss in original scale; gradient $\propto b^p$}
\vspace{0.5em}
\State \textbf{// Scale-invariant (\ours) objective}
\State Scale targets: $z_{t+h} \gets (y_{t+h} - a) / b$ for $h = 1, \ldots, H$
\State $\mathcal{L}_{\mathrm{inv}} \gets \frac{1}{H}\sum_{h=1}^{H} \phi\left(z_{t+h} - \hat{z}_{t+h}\right)$ \Comment{Loss in scaled space; gradient invariant to $b$}
\vspace{0.5em}
\State Backpropagate chosen loss and update $\boldsymbol{\theta}$
\end{algorithmic}
\end{algorithm}
 \clearpage
\section{Dataset details}
\subsection{M-series Benchmark Details} \label{appendix:dataset_details}

\begin{figure}[h!]
    \centering
    \includegraphics[width=0.55\linewidth]{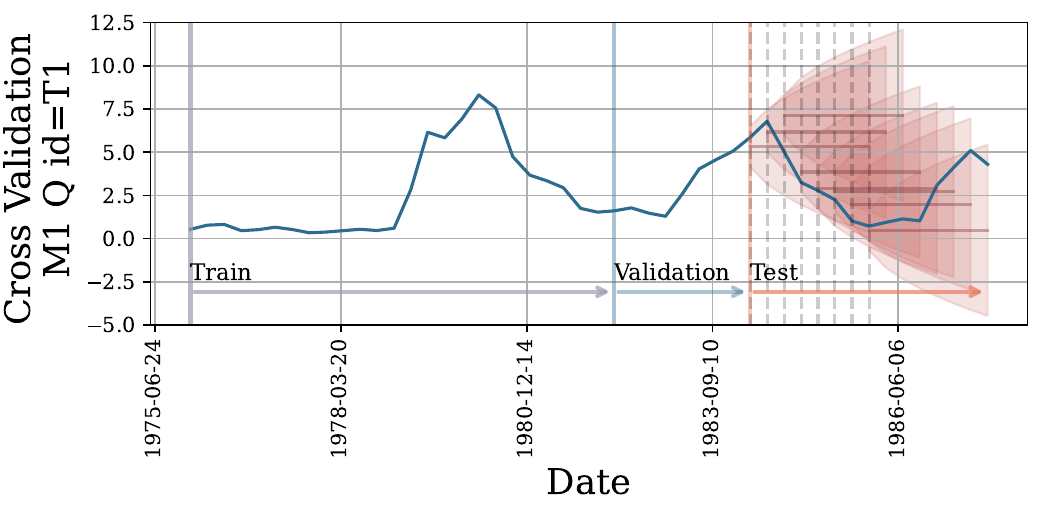}
    \vspace{-0.2cm}
    \caption{
    Example time series from the M1 competition dataset. The train and validation sets consist of all observations preceding the first dotted line. The cross validation test set is comprised of a set of forecast creation dates with lengths defined in Table~\ref{table:datasets}. 
    }
    \label{fig:training_methodology}
\end{figure}

\vspace{-0.5cm}
% Please add the following required packages to your document preamble:
\begin{table}[!ht]
\caption{Summary of forecasting datasets used in our empirical study.} \label{table:datasets}
\centering
% \tiny
% \scriptsize
\footnotesize
\resizebox{0.9\textwidth}{!}{
\begin{tabular}{llcccccc}
\toprule
                         & Frequency & Seasonality & Horizon                     & Series                     & Min Length                     & Max Length                     & \% Erratic                     \\ \midrule
\multirow{3}{*}{M1}      & Monthly   & 12          & 18                          & 617                        & 48                             & 150                            & 0                              \\
                         & Quarterly & 4           & 8                           & 203                        & 18                             & 114                            & 0                              \\
                         & Yearly    & 1           & 6                           & 181                        & 15                             & 58                             & 0                              \\ \midrule
\multirow{4}{*}{M3}      & Other     & 4           & 8                           & 174                        & 71                             & 104                            & 0                              \\
                         & Monthly   & 12          & 18                          & 1428                       & 66                             & 144                            & 2                              \\
                         & Quarterly & 4           & 8                           & 756                        & 24                             & 72                             & 1                              \\
                         & Yearly    & 1           & 6                           & 645                        & 20                             & 47                             & 10                             \\ \midrule
\multirow{6}{*}{M4}      & Hourly    & 24          & 48                          & 414                        & 748                            & 1008                           & 17                             \\ 
                         & Daily     & 1           & 14                          & 4,227                      & 107                            & 9933                           & 2                              \\
                         & Weekly    & 1           & 13                          & 359                        & 93                             & 2610                           & 16                             \\
                         & Monthly   & 12          & 18                          & 48,000                     & 60                             & 2812                           & 6                              \\
                         & Quarterly & 4           & 8                           & 24,000                     & 24                             & 874                            & 11                             \\
                         & Yearly    & 1           & 6                           & 23,000                     & 19                             & 841                            & 18                             \\ \midrule
\multirow{3}{*}{Tourism} & Monthly   & 12          & 18                          & 366                        & 91                             & 333                            & 51                             \\
                         & Quarterly & 4           & 8                           & 427                        & 30                             & 130                            & 39                             \\
                         & Yearly    & 1           & 6                           & 518                        & 11                             & 47                             & 23                             \\ \bottomrule
\end{tabular}
}
\end{table}

The early  \Mone\ competition~\citep{makridakis1982m1_competition} competition features 1,001 demographic, industrial, and economic time series spanning varying frequencies with lengths from 9 to 132 observations. Its key finding, that simple methods like ETS~\citep{holt1957exponential_smoothing} often outperform complex approaches, shaped a legacy emphasizing accuracy, automation, and overfitting caution while shifting focus from traditional forecasting.

Held two decades after \Mone, the \Mthree competition~\citep{makridakis2000m3_competition} featured 3,003 positive-valued time series across business, demographic, financial, and economic domains with lengths from 14 to 126 observations. Reinforcing the trend that simple methods often outperform complex alternatives, the Theta method~\citep{hyndman2003unmasking_theta} emerged as the top-performing approach.

The \Mfour competition~\citep{makridakis2020m4_competition} massively scaled up size and diversity to 100,000 time series across six frequencies and multiple domains. Expanding evaluation beyond point forecasts to include prediction intervals, it featured 18\% erratic series~\citep{syntetos2005syntetos_categorization} and marked the first time a neural model, ESRNN~\citep{smyl2019esrnn}, outperformed traditional methods.

The \Tourism\ dataset~\citep{athanosopoulus2011tourism_competition} comprises 1,311 monthly, quarterly, and yearly series designed to evaluate demand forecasting. It used Mean Absolute Scaled Error (MASE) for scaled point forecasts and evaluated forecast intervals, with 36\% of series classified as erratic or intermittent, making \SNaive difficult to beat annually.

\clearpage
\subsection{GIFT-eval Benchmark Details} \label{appendix:gifteval_details}

\begin{table}[!ht]
\caption{Summary of GiftEval forecasting datasets used in our empirical study.
} \label{table:gifteval_datasets}
\centering
\footnotesize
\resizebox{\textwidth}{!}{
\begin{tabular}{lllrrrrrrrr}
\hline
Dataset & Source & Domain & Frequency & \# Series & Avg & Min & Max & \# Obs \\ \hline
Jena Weather & Autoformer \citep{wu2021autoformer} & Nature & 10T & 1 & 52,704 & 52,704 & 52,704 & 52,704 \\
Jena Weather & Autoformer \citep{wu2021autoformer} & Nature & H & 1 & 8,784 & 8,784 & 8,784 & 8,784 \\
Jena Weather & Autoformer \citep{wu2021autoformer} & Nature & D & 1 & 366 & 366 & 366 & 366 \\ %\hline
BizITObs - Application & AutoMixer \citep{palaskar2024automixer} & Web/CloudOps & 10S & 1 & 8,834 & 8,834 & 8,834 & 8,834 \\
BizITObs - Service & AutoMixer \citep{palaskar2024automixer} & Web/CloudOps & 10S & 21 & 8,835 & 8,835 & 8,835 & 185,535 \\
BizITObs - L2C & AutoMixer \citep{palaskar2024automixer} & Web/CloudOps & 5T & 1 & 31,968 & 31,968 & 31,968 & 31,968 \\
BizITObs - L2C & AutoMixer \citep{palaskar2024automixer} & Web/CloudOps & H & 1 & 2,664 & 2,664 & 2,664 & 2,664 \\ %\hline
Bitbrains - Fast Storage & Grid Workloads Archive \citep{shen2015bitbrains} & Web/CloudOps & 5T & 1,250 & 8,640 & 8,640 & 8,640 & 10,800,000 \\
Bitbrains - Fast Storage & Grid Workloads Archive \citep{shen2015bitbrains} & Web/CloudOps & H & 1,250 & 721 & 721 & 721 & 901,250 \\
Bitbrains - rnd & Grid Workloads Archive \citep{shen2015bitbrains} & Web/CloudOps & 5T & 500 & 8,640 & 8,640 & 8,640 & 4,320,000 \\
Bitbrains - rnd & Grid Workloads Archive \citep{shen2015bitbrains} & Web/CloudOps & H & 500 & 720 & 720 & 720 & 360,000 \\ %\hline
Restaurant & Recruit Rest. Comp. \citep{howard2017recruit} & Sales & D & 807 & 358 & 67 & 478 & 289,303 \\ %\hline
ETT1 & Informer \citep{zhou2021informer} & Energy & 15T & 1 & 69,680 & 69,680 & 69,680 & 69,680 \\ 
ETT1 & Informer \citep{zhou2021informer} & Energy & H & 1 & 17,420 & 17,420 & 17,420 & 17,420 \\
ETT1 & Informer \citep{zhou2021informer} & Energy & D & 1 & 725 & 725 & 725 & 725 \\
ETT1 & Informer \citep{zhou2021informer} & Energy & W-THU & 1 & 103 & 103 & 103 & 103 \\
ETT2 & Informer \citep{zhou2021informer} & Energy & 15T & 1 & 69,680 & 69,680 & 69,680 & 69,680 \\
ETT2 & Informer \citep{zhou2021informer} & Energy & H & 1 & 17,420 & 17,420 & 17,420 & 17,420 \\
ETT2 & Informer \citep{zhou2021informer} & Energy & D & 1 & 725 & 725 & 725 & 725 \\
ETT2 & Informer \citep{zhou2021informer} & Energy & W-THU & 1 & 103 & 103 & 103 & 103 \\ %\hline
Loop Seattle & LibCity \citep{wang2023libcity} & Transport & 5T & 323 & 105,120 & 105,120 & 105,120 & 33,953,760 \\
Loop Seattle & LibCity \citep{wang2023libcity} & Transport & H & 323 & 8,760 & 8,760 & 8,760 & 2,829,480 \\
Loop Seattle & LibCity \citep{wang2023libcity} & Transport & D & 323 & 365 & 365 & 365 & 117,895 \\ %\hline
SZ-Taxi & LibCity \citep{wang2023libcity} & Transport & 15T & 156 & 2,976 & 2,976 & 2,976 & 464,256 \\
SZ-Taxi & LibCity \citep{wang2023libcity} & Transport & H & 156 & 744 & 744 & 744 & 116,064 \\ %\hline
M\_DENSE & LibCity \citep{wang2023libcity} & Transport & H & 30 & 17,520 & 17,520 & 17,520 & 525,600 \\
M\_DENSE & LibCity \citep{wang2023libcity} & Transport & D & 30 & 730 & 730 & 730 & 21,900 \\ %\hline
Solar & LSTNet \citep{lai2018lstnet} & Energy & 10T & 137 & 52,560 & 52,560 & 52,560 & 7,200,720 \\
Solar & LSTNet \citep{lai2018lstnet} & Energy & H & 137 & 8,760 & 8,760 & 8,760 & 1,200,120 \\
Solar & LSTNet \citep{lai2018lstnet} & Energy & D & 137 & 365 & 365 & 365 & 50,005 \\
Solar & LSTNet \citep{lai2018lstnet} & Energy & W-FRI & 137 & 52 & 52 & 52 & 7,124 \\ %\hline
Hierarchical Sales & \citet{mancuso2020machine} & Sales & D & 118 & 1,825 & 1,825 & 1,825 & 215,350 \\
Hierarchical Sales & \citet{mancuso2020machine} & Sales & W-WED & 118 & 260 & 260 & 260 & 30,680 \\ %\hline
M4 Yearly & Monash \citep{godahewa2021monash} & Econ/Fin & A-DEC & 22,974 & 37 & 19 & 284 & 845,109 \\
M4 Quarterly & Monash \citep{godahewa2021monash} & Econ/Fin & Q-DEC & 24,000 & 100 & 24 & 874 & 2,406,108 \\
M4 Monthly & Monash \citep{godahewa2021monash} & Econ/Fin & M & 48,000 & 234 & 60 & 2,812 & 11,246,411 \\
M4 Weekly & Monash \citep{godahewa2021monash} & Econ/Fin & W-SUN & 359 & 1,035 & 93 & 2,610 & 371,579 \\
M4 Daily & Monash \citep{godahewa2021monash} & Econ/Fin & D & 4,227 & 2,371 & 107 & 9,933 & 10,023,836 \\
M4 Hourly & Monash \citep{godahewa2021monash} & Econ/Fin & H & 414 & 902 & 748 & 1,008 & 373,372 \\ %\hline
Hospital & Monash \citep{godahewa2021monash} & Healthcare & M & 767 & 84 & 84 & 84 & 64,428 \\ %\hline
COVID Deaths & Monash \citep{godahewa2021monash} & Healthcare & D & 266 & 212 & 212 & 212 & 56,392 \\ %\hline
US Births & Monash \citep{godahewa2021monash} & Healthcare & D & 1 & 7,305 & 7,305 & 7,305 & 7,305 \\
US Births & Monash \citep{godahewa2021monash} & Healthcare & W-TUE & 1 & 1,043 & 1,043 & 1,043 & 1,043 \\
US Births & Monash \citep{godahewa2021monash} & Healthcare & M & 1 & 240 & 240 & 240 & 240 \\ %\hline
Saugeen & Monash \citep{godahewa2021monash} & Nature & D & 1 & 23,741 & 23,741 & 23,741 & 23,741 \\ 
Saugeen & Monash \citep{godahewa2021monash} & Nature & W-THU & 1 & 3,391 & 3,391 & 3,391 & 3,391 \\
Saugeen & Monash \citep{godahewa2021monash} & Nature & M & 1 & 780 & 780 & 780 & --- \\ %\hline
Temperature Rain & Monash \citep{godahewa2021monash} & Nature & D & 32,072 & 725 & 725 & 725 & 780 \\ %\hline
KDD Cup 2018 & Monash \citep{godahewa2021monash} & Nature & H & 270 & 10,898 & 9,504 & 10,920 & 2,942,364 \\
KDD Cup 2018 & Monash \citep{godahewa2021monash} & Nature & D & 270 & 455 & 396 & 455 & 122,791 \\ %\hline
Car Parts & Monash \citep{godahewa2021monash} & Sales & M & 2,674 & 51 & 51 & 51 & 136,374 \\ %\hline
Electricity & UCI ML Archive \citep{trindade2015electricity} & Energy & 15T & 370 & 140,256 & 140,256 & 140,256 & 51,894,720 \\
Electricity & UCI ML Archive \citep{trindade2015electricity} & Energy & H & 370 & 35,064 & 35,064 & 35,064 & 12,973,680 \\
Electricity & UCI ML Archive \citep{trindade2015electricity} & Energy & D & 370 & 1,461 & 1,461 & 1,461 & 540,570 \\
Electricity & UCI ML Archive \citep{trindade2015electricity} & Energy & W-FRI & 370 & 208 & 208 & 208 & 76,960 \\ \hline
\end{tabular}
}
\end{table}

The GiftEval \citep{aksu2024giftevalbenchmarkgeneraltime} benchmark leverages 10 distinct data sources to construct a comprehensive testing suite. This incorporates the Jena Weather collection from Autoformer~\citep{wu2021autoformer}, business and IT infrastructure observability data  AutoMixer~\citep{palaskar2024automixer}, cloud workloads sourced from the Grid Workloads Archive via Bitbrains~\citep{shen2015bitbrains}, temperature profiles from Informer~\citep{zhou2021informer}, solar generation series from LSTNet~\citep{lai2018lstnet}, sales histories~\citep{mancuso2020machine}, recruit restaurant visitors \citep{howard2017recruit}, urban traffic logs from LibCity~\citep{wang2023libcity}, selected non-overlapping subsets from the Monash repository~\citep{godahewa2021monash}, and smart-meter electricity tracking from the UCI Machine Learning Archive~\citep{trindade2015electricity}. Summary statistics for each source including frequencies, prediction lengths, variate counts, and scale, are detailed in Table \ref{table:gifteval_datasets}.

% In this work we prioritized meas
 \clearpage
\section{Controlled Synthetic Data Experiment Details} \label{apd:synthetic-experiments}

\subsection{Reproducibility Details}\label{apd:synthetic-reproducibility}

\paragraph{Data.} We generate $3$ noiseless sinusoidal sources. For source $k$ and training phase $s$,
\begin{equation}
    y_t^{(k,s)} = 1 + \sqrt{2v_k}\sin\!\left(\frac{2\pi f_k t}{32} + \phi_k + \frac{2\pi s}{8}\right),
\end{equation}
where $(v_k,f_k,\phi_k)$ is $(1,1,0)$, $(10,2,\pi/3)$, or $(100,3,2\pi/3)$. Each 64-point context has population variance $v_k$, giving scaling denominators in the approximate ratio $1\!:\!\sqrt{10}\!:\!10$ and MSE multipliers $b^2$ in the ratio $1\!:\!10\!:\!100$. Each source has $8$ training series of length $2{,}048$. We use stride-$1$ windows with context length $64$ and horizon $16$. Validation contains $512$ fixed windows per source sampled from $4$ series on a phase grid shifted by $\pi/8$ from training.

\paragraph{Model and scaling.} The univariate patch Transformer uses $8$ nonoverlapping patches of length $8$, embedding dimension $64$, $2$ encoder layers, $4$ attention heads, feedforward width $128$, dropout $0$, and a direct linear head for the $16$-step horizon. Each context is scaled by its sample mean and sample standard deviation with $\epsilon=10^{-5}$. We compute MSE either on scaled targets for the scale-invariant condition or after reversing the affine transform for the scale-contaminated condition.

\paragraph{Optimization.} We train for $30{,}000$ updates with batch size $24$ ($8$ windows per source) and seeds $\{0,1,2,3,4\}$. SGD uses learning rate $10^{-3}$, momentum $0$, and weight decay $0$. Adam uses the same learning rate with $(\beta_1,\beta_2)=(0.9,0.999)$, $\epsilon=10^{-8}$, and weight decay $0$. Learning rates are constant. Within each seed, all conditions share the initialization and sampled batches.

\paragraph{Controls and evaluation.} The equal-scale control sets all amplitudes to $1$. The gradient-matched control rescales the full parameter gradient to unit Euclidean norm after every backward pass. We report validation nMSE on scaled targets as the mean $\pm 1$ standard deviation over $5$ seeds. Evaluation occurs every $5$ updates through step $200$, every $50$ through step $1{,}000$, and every $1{,}000$ thereafter.

\subsection{Scale-invariance Comparison Under Adam}\label{apd:adam-loss-space}

A natural objection to the controlled synthetic experiment of \Cref{sec:exp_synthetic} is that an adaptive optimizer should absorb the $b^p$ factor. Adam \citep{kingma2014adam} is invariant to a global rescaling of the loss: scaling every gradient by $c$ scales the first moment by $c$ and the second by $c^2$, leaving the update unchanged. If the $b^p$ factor applied uniformly to the objective, Adam would cancel it. It does not apply uniformly. Each source enters the batch gradient weighted by its own $b^p$, and the mean reduction fixes those relative weights before the optimizer acts, so Adam rescales the mixture without changing its composition.

We repeat the synthetic loss-space comparison of \Cref{fig:synthetic:normalized,fig:synthetic:original}, with Adam in place of SGD, keeping all other knobs identical. Adam normalizes each coordinate by the running root-mean-square of its own gradient. This cancels the \emph{global} magnitude inflation of the scale-contaminated gradient, so scale-contaminated training no longer risks divergence and converges faster overall, but it does not change which source dominates training, and the step-0 per-source gradient ratio is identical to SGD.

\begin{figure}[htbp]
    \centering
    \includegraphics[width=\linewidth]{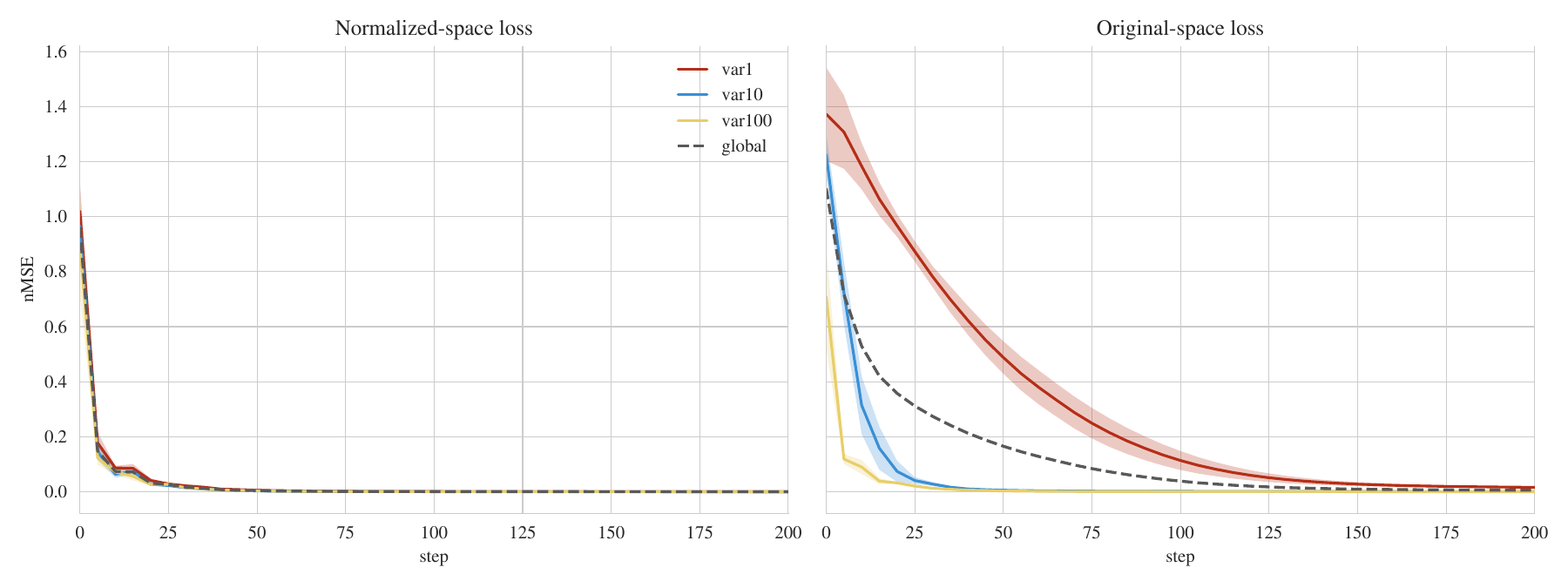}
    \caption{Adam, per-source nMSE over the first $200$ steps (linear axis, mean over 5 seeds with $\pm 1$ std bands; dashed grey = global nMSE). \textbf{Left:} under the scale-invariant loss all three sources converge together. \textbf{Right:} under the scale-contaminated loss the curves spread out in scale order (high-scale first, low-scale last), the same disparity SGD shows but compressed into the first ${\sim}150$ steps.}
    \label{fig:apd:adam:core}
\end{figure}

\begin{figure}[htbp]
    \centering
    \includegraphics[width=\linewidth]{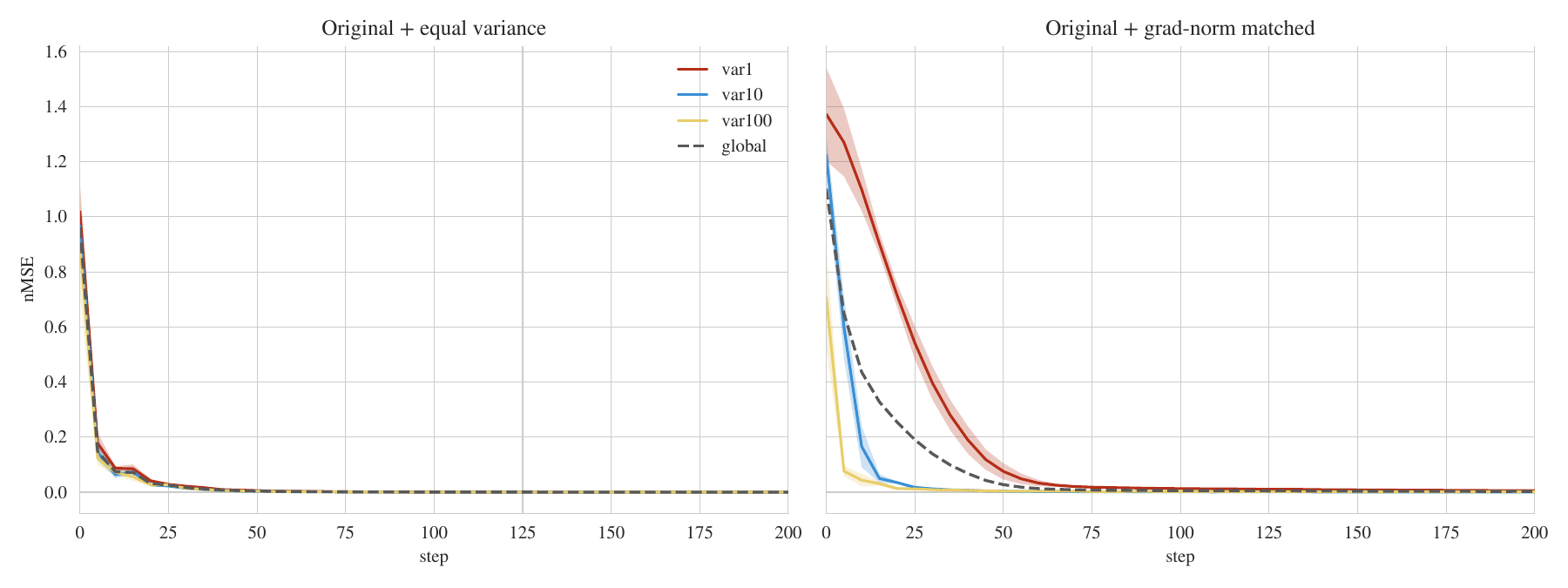}
    \caption{Adam control experiments (first $200$ steps, linear axis, mean $\pm 1$ std over 5 seeds). \textbf{Left:} on equal-scale data the scale-contaminated spread collapses to the scale-invariant baseline. \textbf{Right:} under gradient-norm-matched scale-contaminated training the scale-ordered spread persists.}
    \label{fig:apd:adam:controls}
\end{figure}

Under the scale-contaminated loss the highest-scale source reaches $\text{nMSE}<0.1$ within ${\sim}10$ steps (\Cref{fig:apd:adam:core}), the next within ${\sim}20$, and the lowest within ${\sim}105$. The time-to-threshold gap is down to ${\sim}10\times$ from the ${\sim}50\times$ under SGD, but it remains clearly scale-ordered. Under the scale-invariant loss all three sources reach the threshold together in about $10$ steps. The equal-scale and gradient-norm-matched controls (\Cref{fig:apd:adam:controls}), as under SGD, confirm the spread is not an artifact of a confounding factor. Together with the gradient-norm-matched control in the main text, this places the effect in the composition of the batch gradient rather than in its magnitude, which is what the learning rate, gradient clipping, and adaptive preconditioning all act on.

\begin{figure}[htbp]
    \centering
    \begin{subfigure}[t]{0.49\linewidth}
    \includegraphics[width=\linewidth]{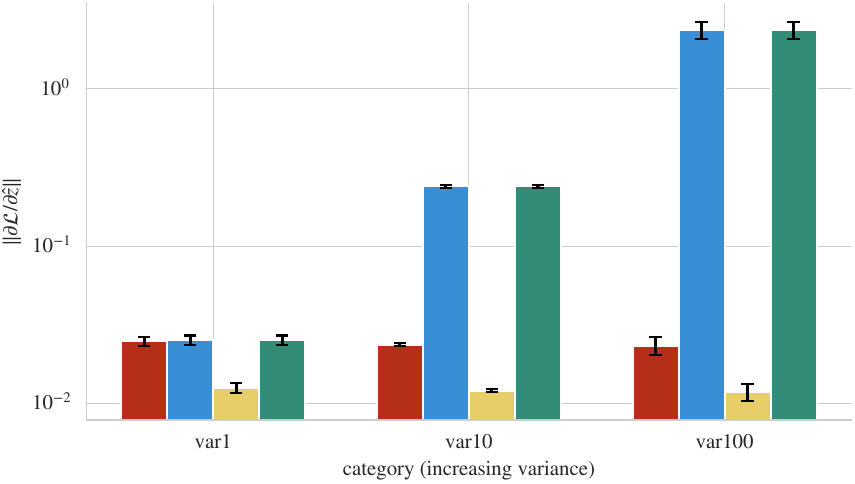}
    \caption{Per-source gradient magnitude at initialization.}
    \end{subfigure}
    \hfill
    \begin{subfigure}[t]{0.49\linewidth}
    \includegraphics[width=\linewidth]{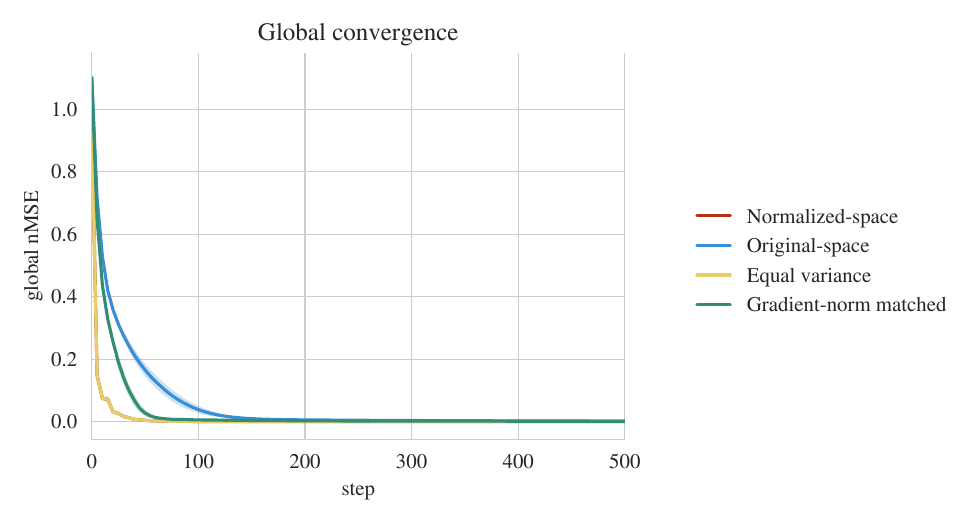}
    \caption{Global nMSE (first $500$ steps).}
    \end{subfigure}
    \caption{Adam, gradient magnitude and global convergence (mean $\pm 1$ std over 5 seeds). The initialization gradient ratio is unchanged from SGD and scales as $b^p$ under the scale-contaminated loss. Adam rescales coordinate magnitude but not the scale-driven ordering of convergence.}
    \label{fig:apd:adam:grad-global}
\end{figure}

\begin{figure}[htbp]
    \centering
    \begin{subfigure}[t]{0.8\linewidth}
    \includegraphics[width=\linewidth]{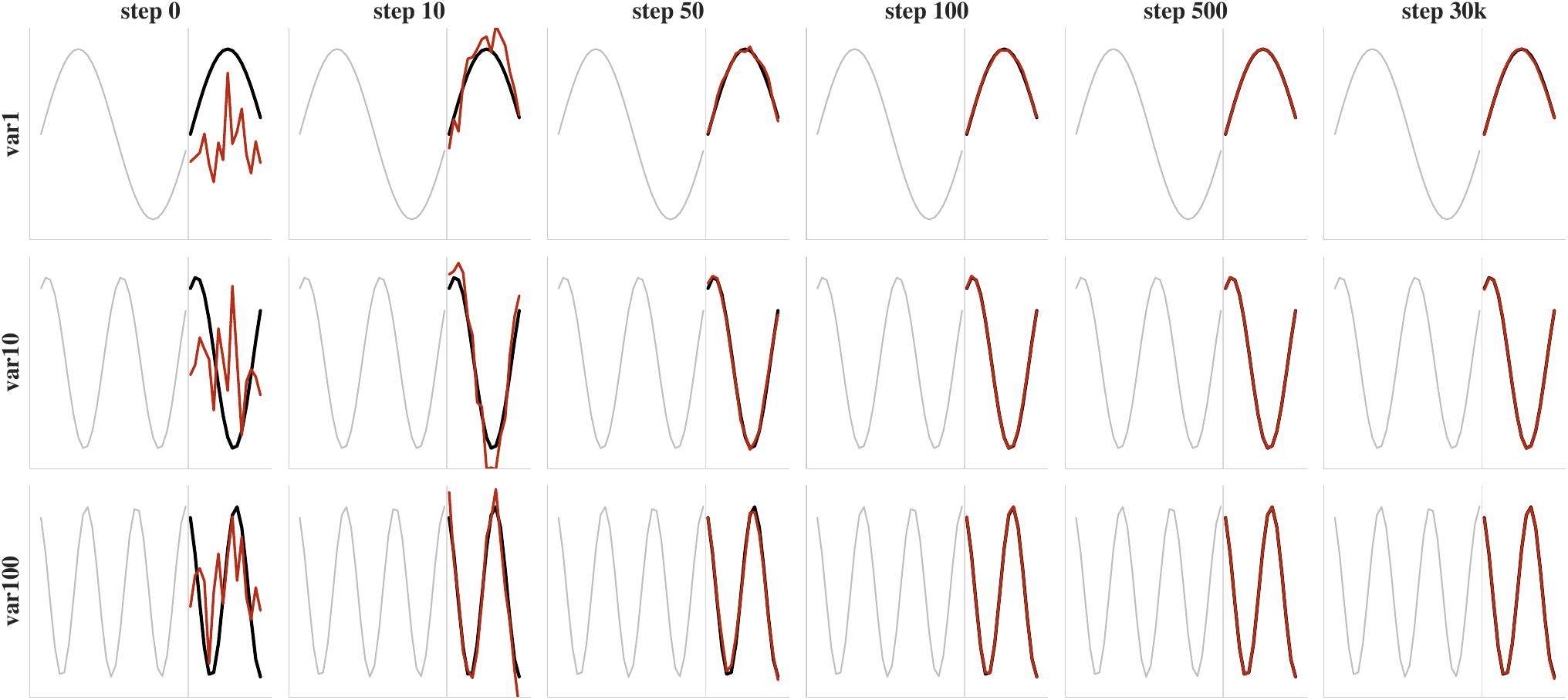}
    \caption{Scale-invariant loss (\ours)}
    \end{subfigure}
    \vfill
    \begin{subfigure}[t]{0.8\linewidth}
    \includegraphics[width=\linewidth]{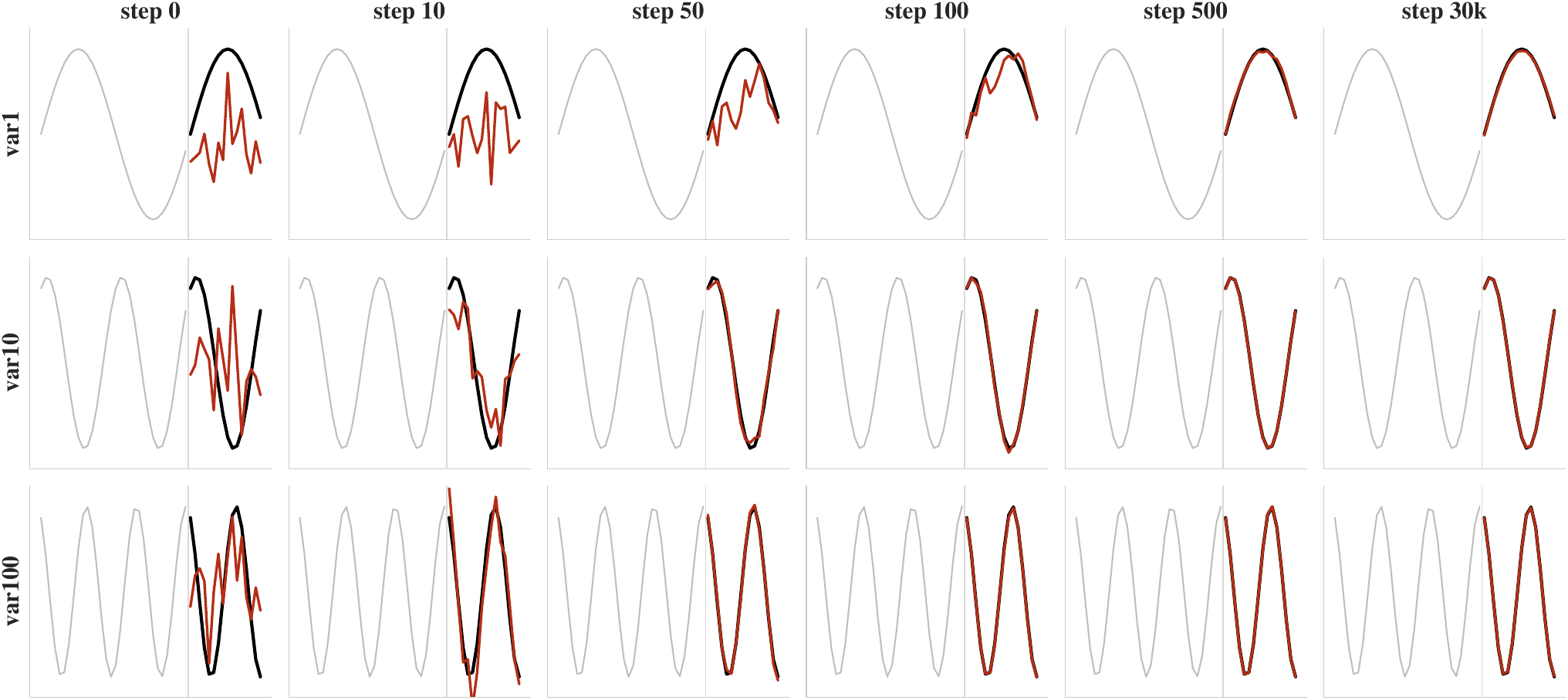}
    \caption{Scale-\contaminated loss}
    \end{subfigure}
    \caption{Adam, qualitative per-source forecasts at training steps $0$, $10$, $50$, $100$, $500$, and $30{,}000$.
    }
    \label{fig:apd:adam:forecasts}
\end{figure}
  \clearpage
\section{Scale-Swap Intervention Experiment}
\label{apd:scale-swap-details}

\subsection{Reproducibility Details}
\label{apd:scale-swap-reproducibility}

\paragraph{Data.} We use the $8$ datasets listed in \Cref{sec:exp_scale_swap}, represented as univariate series of length $128$. We split across complete series, assigning a fixed random $80\%$ of the series to training and the remaining $20\%$ to validation. No series appears in both splits. We form stride-$1$ windows only after this split, using context length $64$ and horizon $16$ and removing windows with nonfinite or constant contexts. Each window is scaled to zero context mean and unit context standard deviation, then multiplied by its assigned $b\in\{1,10\}$. Validation uses $512$ fixed windows per dataset.

\paragraph{Assignments.} Each configuration assigns $4$ datasets to each scale. Its complement reverses all $8$ assignments. We use $15$ complementary pairs, giving $30$ configurations in which every dataset appears $15$ times at each scale. Each pair of data sets shares the larger scale in $6$ or $7$ configurations.

\paragraph{Model and optimization.} The patch Transformer uses $8$ nonoverlapping patches of length $8$, embedding dimension $128$, $4$ encoder layers, $8$ attention heads, feedforward width $256$, dropout $0$, and a direct linear head for the $16$-step horizon. Scale each context using its sample mean and sample standard deviation with $\epsilon=10^{-5}$. We train for $30{,}000$ SGD updates with batch size $32$ ($4$ windows per dataset), momentum $0$, weight decay $0$, and seeds $\{0,1,2,3,4\}$. Within each seed and assignment pair, conditions share the initialization and sampled batches.

\paragraph{Learning-rate control.} Both conditions use constant learning rates. The scale-invariant condition uses $10^{-4}$. Under scale-contaminated MSE, $b=10$ multiplies a window's gradient by $100$, while $b=1$ leaves it unchanged. Because every batch contains equal numbers from the $2$ scales, the nominal average multiplier is $(1+100)/2=50.5$. We therefore use $10^{-4}/50.5=1.9802\times10^{-6}$ for the scale-contaminated condition. This keeps the average update magnitude comparable without removing the relative $100\!:\!1$ weighting between scales.

\paragraph{Evaluation.} We report validation nMSE on scaled targets. Evaluation occurs every $5$ updates through step $200$, every $50$ through step $1{,}000$, and every $1{,}000$ thereafter. The trajectory summary and paired test are defined below.

\subsection{Convergence Statistic and Test}
\label{apd:scale-swap-metric}

Let $\tau_1 < \cdots < \tau_J$ be the training steps at which we evaluate held-out nMSE. For dataset $D^{(k)}$, scale $b$, complementary assignment pair $c$, seed $s$, and checkpoint $j$, write this error as $e^{(k)}_{b,c,s,j}$. We summarize the complete validation trajectory by integrating the straight lines between consecutive checkpoints and dividing by the training duration,
\begin{equation}
\begin{split}
    A^{(k)}_{b,c}
    = \frac{1}{S}\sum_{s=1}^{S}
    \frac{1}{\tau_J-\tau_1}
    \sum_{j=1}^{J-1}
    \frac{\tau_{j+1}-\tau_j}{2}
    \left(e^{(k)}_{b,c,s,j}+e^{(k)}_{b,c,s,j+1}\right).
\end{split}
\label{eq:scale-swap-trajectory-error}
\end{equation}
This weighted average remains on the nMSE scale. It gives more weight to checkpoint intervals that cover more training steps. Because all runs share the same checkpoints, it compares their full trajectories on the same basis.

Each complementary assignment pair places every dataset at $b{=}1$ in one run and at $b{=}10$ in the other. For each dataset, we average over the $C=15$ paired assignments,
\begin{equation}
    \bar{A}^{(k)}_b
    = \frac{1}{C}\sum_{c=1}^{C} A^{(k)}_{b,c},
    \qquad
    \Delta_k
    = \bar{A}^{(k)}_1-\bar{A}^{(k)}_{10}.
\label{eq:scale-swap-dataset-effect}
\end{equation}
A positive $\Delta_k$ means dataset $D^{(k)}$ has lower validation error over training when assigned the larger scale. The aggregate effect is
\begin{equation}
    \bar{\Delta} = \frac{1}{K}\sum_{k=1}^{K}\Delta_k.
\label{eq:scale-swap-aggregate-effect}
\end{equation}

The one-sided paired $t$-test uses the $K=8$ dataset-level differences $d_k=\Delta_k$ as its statistical units. With sample mean $\bar{d}$ and sample standard deviation $s_d$, the test statistic is
\begin{equation}
    t = \frac{\bar{d}}{s_d/\sqrt{K}},
    \qquad
    s_d^2 = \frac{1}{K-1}\sum_{k=1}^{K}(d_k-\bar{d})^2.
\label{eq:scale-swap-t-test}
\end{equation}
The null hypothesis is $\mathbb{E}[d_k]\leq 0$ and the one-sided alternative is $\mathbb{E}[d_k]>0$. Under the scale-invariant loss, the mean trajectory-level errors across datasets are $1.964752$ at $b{=}1$ and $1.964783$ at $b{=}10$, a relative difference below $0.002\%$. For the scale-contaminated loss, the test gives $t_7{=}2.013$ and $p{=}0.042$. We compute the reported percentage from its means across datasets at each scale,
\begin{equation}
    \bar{A}_b = \frac{1}{K}\sum_{k=1}^{K}\bar{A}^{(k)}_b,
    \qquad
    \bar{A}_1=2.5925,
    \qquad
    \bar{A}_{10}=1.8990,
\end{equation}
which gives
\begin{equation}
    100\times\frac{\bar{A}_1-\bar{A}_{10}}{\bar{A}_1}=26.8\%.
\label{eq:scale-swap-percent-reduction}
\end{equation}
We repeat the trajectory and paired-effect calculations separately for the scale-invariant and scale-contaminated losses.

\begin{figure}[htbp]
    \centering
    \includegraphics[width=\linewidth]{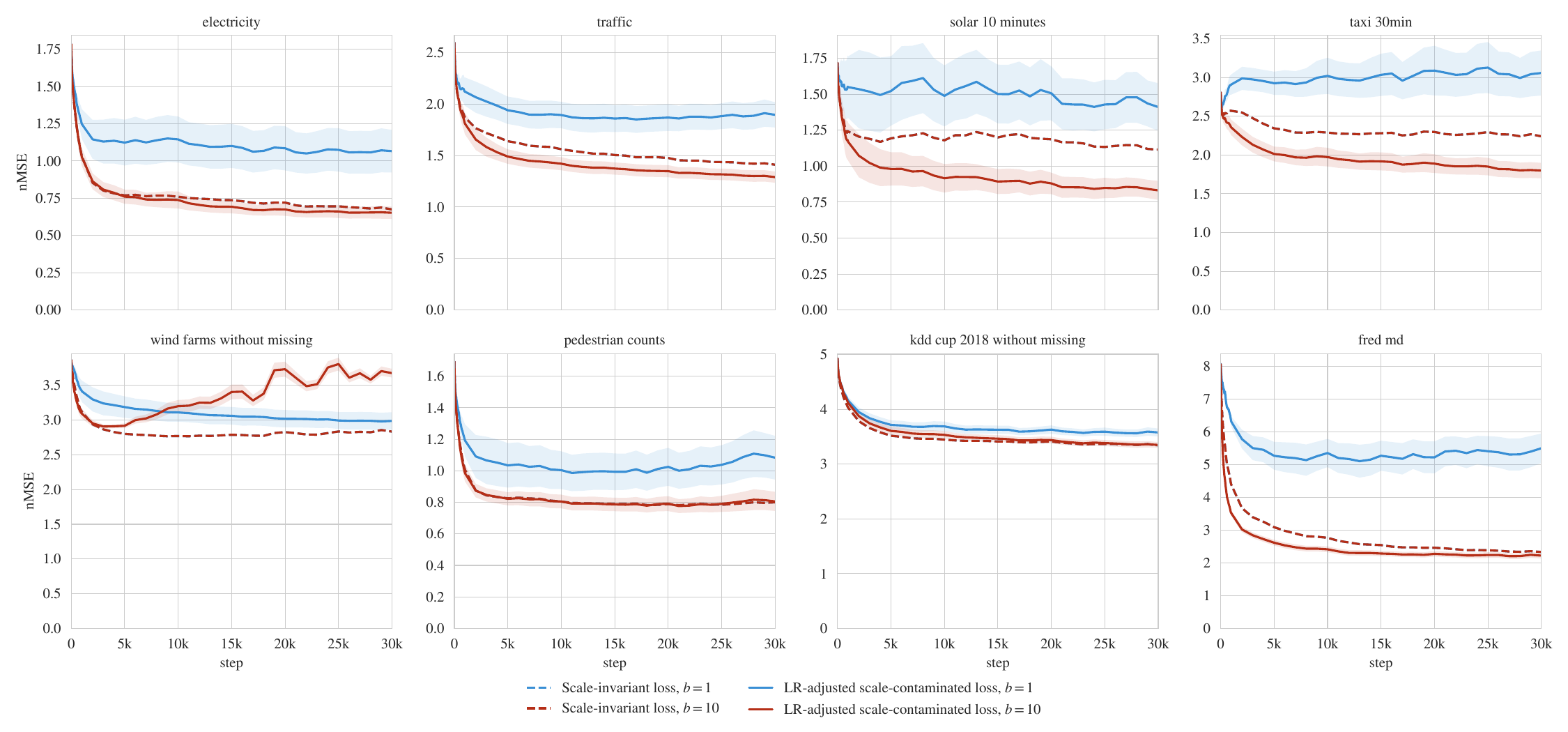}
    \caption{Dataset-level trajectories under scale-invariant loss and learning-rate-adjusted scale-contaminated loss. Under scale-invariant loss, the $b=1$ and $b=10$ trajectories overlap because the loss is computed on scaled targets. Lines and bands show the mean and 95\% Student-$t$ interval across $15$ assignment contexts after averaging $5$ seeds within each context.}
    \label{fig:loss-space-scale-swap-datasets}
\end{figure}

\Cref{fig:loss-space-scale-swap-datasets} disaggregates the aggregate scale-swap trajectory in \Cref{fig:tsfm-natural-mase-convergence:scaleswap}. The $2$ scales overlap under the scale-invariant loss for every dataset. Under the scale-contaminated loss, the inflated copy converges faster in $7$ of $8$ datasets. Wind Farms is the only exception. These are the same dataset-level trajectories used for the paired analysis in \Cref{sec:exp_scale_swap}.

 \clearpage

\section{TSFM Pretraining Details}
\label{apd:tsfm-details}

\subsection{Reproducibility Details}
\label{apd:tsfm-reproducibility}

\paragraph{Corpus and windows.} We pretrain on GiftEvalPretrain together with the training prefixes of M1, M3, M4, Tourism, and Favorita. All benchmark validation and test observations are excluded. The resulting corpus contains $73$ datasets with usable windows. Before constructing windows, a fixed series-level split assigns $90\%$ of the series in each dataset to training and $10\%$ to validation, so no series appears in both splits. The maximum context and prediction lengths are $512$ and $128$, with minima of $64$ and $8$. For a series of length $n$, we set $H=\min(128,\max(8,\lfloor n/5\rfloor))$ and $L=\min(512,n-H)$. If $L<64$, we instead set $L=64$ and $H=n-64$. Series shorter than $72$ observations are excluded. We use stride $512$, retain windows with at least $90\%$ observed values, and pad shorter windows to the maximum shape with masks.

\paragraph{Sampling.} Each training example first samples a dataset uniformly, then a series uniformly within that dataset, and finally a window uniformly within that series. Thus, datasets and series with more available windows do not automatically receive more updates. The paired loss conditions use the same fixed sequence of sampled batches.

\paragraph{Models and objectives.} All four architectures start from random weights. The two foundation models use their Small configurations. \Chronostwo Small uses patch size $16$, hidden dimension $512$, feed-forward dimension $2048$, $6$ layers, $8$ attention heads, and dropout $0.1$. It applies its native context standardization and arcsinh transform, then optimizes its native CRPS-style objective over $13$ quantiles in $\{0.01,0.05,0.1,0.2,\ldots,0.9,0.95,0.99\}$. \Moiraitwo Small uses patch size $16$, hidden dimension $384$, feed-forward dimension $1024$, $6$ layers, and dropout $0.1$. It applies its native context scaling and optimizes mean quantile loss over the $9$ deciles from $0.1$ to $0.9$ on up to $64$ target steps. \TimesFMtwo is reduced to $49.6$M parameters from the released $200$M definition, keeping its head dimension of $80$ and cutting width to $800$ and depth to $12$ layers. It keeps the input and output patch lengths of $32$ and $128$, so a single forward pass covers the $128$-step horizon and no autoregressive decoding is needed. Its scaling is the per-patch expanding-window statistic that \TimesFMtwo introduced, where patch $i$ is scaled by the masked running mean and standard deviation of every observed point up to and including itself. We train the point head with MSE, as the released model does, and drop the separate $1024$-step quantile head, which is unused at this horizon. \PatchTST follows its original supervised configuration at $1.5$M parameters, with patch length $16$, stride $8$, $3$ encoder layers, $16$ heads, hidden dimension $128$, and dropout $0.2$. It scales by RevIN with the affine parameters disabled, which is the setting its own released code defaults to, and optimizes MSE. Within each model, scale-invariant training computes the native objective on scaled targets, while scale-contaminated training reverses the model's scaling before computing the same objective. Everything else is held fixed. The four therefore differ in where their scaling statistics come from, which is what lets the loss-space comparison be read as independent of that choice.

\paragraph{Optimization and validation.} We train for $80{,}000$ updates at batch size $512$ with seeds $\{0,1,2,3\}$. AdamW uses a constant learning rate of $10^{-4}$, weight decay $0.01$, and global gradient clipping at norm $1.0$. Conditions within each model and seed share their initialization, batches, and optimizer settings. Every $250$ updates, the training curves are evaluated on a fixed set of $64$ validation windows per dataset.

\paragraph{Test splits and inference.} M1, M3, and Tourism reserve the final official horizon, while M4 uses the released train/test files. Favorita uses the final $16$ days of series ending on August 15, 2017, with missing sales filled by zero. For each GIFT-Eval short-term configuration, we use $W=\min(20,\max(1,\lceil 0.1L_{\min}/H\rceil))$ non-overlapping windows of horizon $H$ from each series' tail, where $L_{\min}$ is the shortest series length in that configuration. Its M4 configurations use $W=1$, and multivariate series are evaluated channel-wise. Each origin uses only preceding observations, including observations from earlier windows when available. Models receive the most recent $512$ observations, with shorter histories left-padded and masked, and their outputs are truncated to $H$ without autoregressive rollout or benchmark-specific fine-tuning.

\paragraph{Metric aggregation.} Within each subset, MASE is averaged over valid forecast cases using each case's full preceding history for the seasonal-naive denominator. WQL pools its numerator and denominator across cases before division, including the factor of two and averaging over supplied quantile levels. Missing targets and cases with undefined metric denominators are excluded from the corresponding metric. Benchmark scores average subset scores weighted by their total forecast-case counts, then average over four seeds. Thus, benchmark WQL is a weighted average of subset ratios. The overlapping supervised test origins in Appendix~\ref{apd:supervised-breakdown} constitute a separate evaluation protocol.

\paragraph{Compute.} All runs were performed with a single NVIDIA H200 GPU, where each 80k-step run took between 3.6 and 6.5 GPU-hours depending on the architecture. Generating Table~\ref{tab:tsfm_heldout_indomain-main} used 166 GPU-hours in total.

\subsection{Cross-Dataset Dispersion and Pooled Accuracy During Training}

\begin{figure}[htbp]
    \centering
    \includegraphics[width=\linewidth]{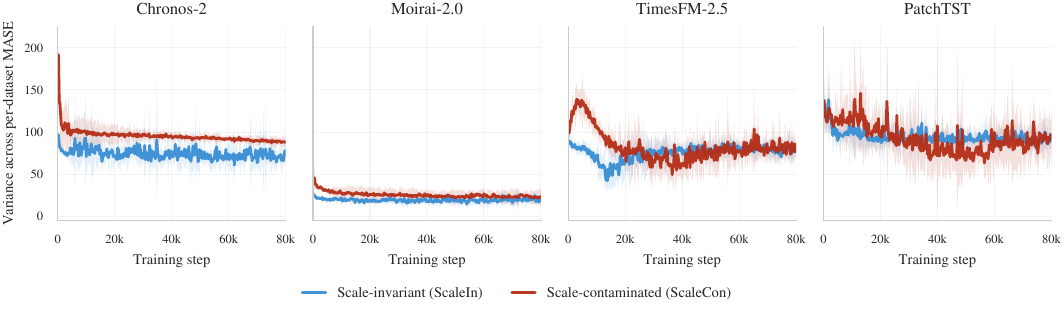}
    \caption{Cross-dataset variance of per-dataset MASE over $80{,}000$ updates for all four architectures. Lines show the mean across $4$ seeds and bands show 95\% Student-$t$ intervals. Lower variance indicates more uniform accuracy across datasets, but does not by itself imply lower forecast error.}
    \label{fig:tsfm-cross-dataset-mase-variance}
\end{figure}

\begin{figure}[htbp]
    \centering
    \includegraphics[width=\linewidth]{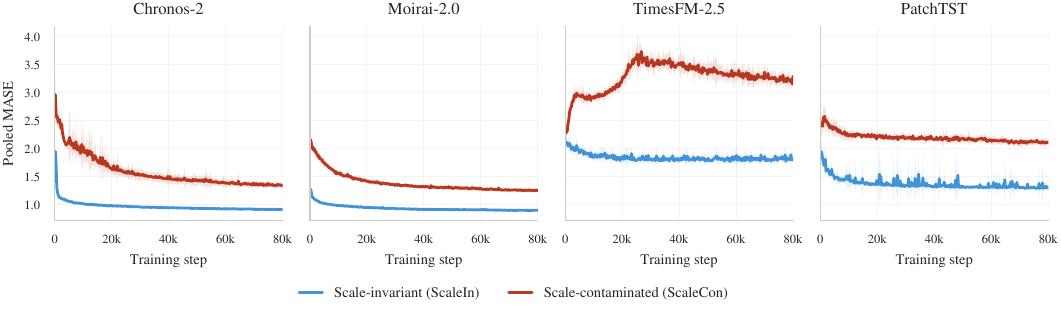}
    \caption{Pooled validation MASE over $80{,}000$ updates for all four architectures. The pooled mean weights usable forecast cases in their natural dataset proportions, unlike the dataset-balanced median in \Cref{fig:tsfm-natural-mase-convergence:pretrain}. Lines show the mean across $4$ seeds and bands show 95\% Student-$t$ intervals. Lower is better.}
    \label{fig:tsfm-pooled-mase-convergence}
\end{figure}

\Cref{fig:tsfm-cross-dataset-mase-variance} complements the median MASE trajectory in \Cref{fig:tsfm-natural-mase-convergence:pretrain} for all four architectures. For each seed and checkpoint, it computes the variance across per-dataset MASE values before averaging across the $4$ seeds. Scale-invariant training clearly reduces dispersion for \Chronostwo and modestly reduces it for \Moiraitwo, while the two losses converge to similar dispersion for \TimesFMtwo and \PatchTST. Low dispersion alone does not imply accurate forecasts, so \Cref{fig:tsfm-pooled-mase-convergence} also reports MASE pooled over all usable validation forecast cases in their natural dataset proportions. Scale-invariant training achieves lower pooled MASE throughout training for every architecture, including the two architectures whose final cross-dataset variance is similar across losses.

\clearpage
\subsection{Benchmark-Frequency Results}

Tables~\ref{tab:tsfm_heldout_indomain-by-frequency-mase} and \ref{tab:tsfm_heldout_indomain-by-frequency-wql} give the full breakdown behind \Cref{tab:tsfm_heldout_indomain-main}. GIFT-Eval reports some frequencies with multipliers or anchors. We pool \texttt{A-DEC} into Y, \texttt{Q-DEC} into Q, \texttt{10S} into S, \texttt{5T}, \texttt{10T}, and \texttt{15T} into T, and the $5$ anchored weekly codes into W. Stars mark these pooled rows. The $6$ GIFT-Eval rows derived from M4 duplicate the official M4 evaluation, so the paper does not average the two benchmarks together.

% Generated by src/scripts/render_paper_tables.py. Do not edit by hand.
\begin{table}[ht]
\centering
\caption{In-domain MASE by benchmark and frequency after 80,000 updates, means with sample standard deviations ($\pm$, below) across 4 seeds. A star marks a base frequency that pools anchored or multiplied GIFT-Eval aliases. \ours uses scale-invariant loss and \SCL uses scale-contaminated loss. These are the only differences between the two runs of each architecture. $\Delta\%$ reports \ours relative to \SCL. \ourswin{Green} denotes an improvement and \oursloss{red} a regression. The better loss is bold within each architecture. References are fitted on the series they forecast and appear behind the double rule.}
\label{tab:tsfm_heldout_indomain-by-frequency-mase}
\renewcommand{\arraystretch}{1.15}
\scriptsize
\setlength{\tabcolsep}{4pt}
\resizebox{\textwidth}{!}{%
\begin{tabular}{llcccccccccccc||c}
\toprule
 &  & \multicolumn{3}{c}{\Chronostwo} & \multicolumn{3}{c}{\Moiraitwo} & \multicolumn{3}{c}{\TimesFMtwo} & \multicolumn{3}{c}{\PatchTST} & \multicolumn{1}{c}{Reference} \\
\cmidrule(lr){3-5}\cmidrule(lr){6-8}\cmidrule(lr){9-11}\cmidrule(lr){12-14}\cmidrule(lr){15-15}
\textbf{Benchmark} & \textbf{Freq.} & \oursShort & \contaminatedShort & $\Delta\%$ & \oursShort & \contaminatedShort & $\Delta\%$ & \oursShort & \contaminatedShort & $\Delta\%$ & \oursShort & \contaminatedShort & $\Delta\%$ & \SNaive \\
\midrule
M1 & Y & \shortstack{\textbf{4.124}\\$\pm 0.445$} & \shortstack{6.492\\$\pm 0.450$} & \ourswin{-36.5} & \shortstack{\textbf{4.894}\\$\pm 0.150$} & \shortstack{5.175\\$\pm 0.406$} & \ourswin{-5.4} & \shortstack{8.067\\$\pm 0.509$} & \shortstack{\textbf{7.765}\\$\pm 0.291$} & \oursloss{+3.9} & \shortstack{\textbf{4.707}\\$\pm 0.104$} & \shortstack{7.600\\$\pm 0.823$} & \ourswin{-38.1} & 4.894 \\
 & Q & \shortstack{\textbf{1.827}\\$\pm 0.056$} & \shortstack{1.997\\$\pm 0.035$} & \ourswin{-8.5} & \shortstack{\textbf{1.814}\\$\pm 0.056$} & \shortstack{1.885\\$\pm 0.053$} & \ourswin{-3.8} & \shortstack{\textbf{1.943}\\$\pm 0.042$} & \shortstack{2.394\\$\pm 0.088$} & \ourswin{-18.8} & \shortstack{\textbf{1.963}\\$\pm 0.078$} & \shortstack{3.625\\$\pm 0.917$} & \ourswin{-45.9} & 2.078 \\
 & M & \shortstack{\textbf{1.238}\\$\pm 0.019$} & \shortstack{1.312\\$\pm 0.013$} & \ourswin{-5.6} & \shortstack{\textbf{1.234}\\$\pm 0.011$} & \shortstack{1.306\\$\pm 0.024$} & \ourswin{-5.5} & \shortstack{\textbf{1.425}\\$\pm 0.018$} & \shortstack{1.579\\$\pm 0.015$} & \ourswin{-9.8} & \shortstack{\textbf{1.366}\\$\pm 0.018$} & \shortstack{2.838\\$\pm 1.050$} & \ourswin{-51.9} & 1.314 \\
M3 & Y & \shortstack{\textbf{3.070}\\$\pm 0.045$} & \shortstack{4.085\\$\pm 0.241$} & \ourswin{-24.8} & \shortstack{\textbf{3.119}\\$\pm 0.127$} & \shortstack{3.643\\$\pm 0.171$} & \ourswin{-14.4} & \shortstack{\textbf{4.880}\\$\pm 0.198$} & \shortstack{5.309\\$\pm 0.188$} & \ourswin{-8.1} & \shortstack{\textbf{3.889}\\$\pm 0.465$} & \shortstack{9.529\\$\pm 2.353$} & \ourswin{-59.2} & 3.154 \\
 & Q & \shortstack{\textbf{1.258}\\$\pm 0.060$} & \shortstack{1.461\\$\pm 0.072$} & \ourswin{-13.9} & \shortstack{\textbf{1.255}\\$\pm 0.039$} & \shortstack{1.295\\$\pm 0.077$} & \ourswin{-3.1} & \shortstack{\textbf{1.476}\\$\pm 0.016$} & \shortstack{1.846\\$\pm 0.110$} & \ourswin{-20.1} & \shortstack{\textbf{1.590}\\$\pm 0.142$} & \shortstack{3.867\\$\pm 1.043$} & \ourswin{-58.9} & 1.425 \\
 & M & \shortstack{\textbf{0.955}\\$\pm 0.018$} & \shortstack{1.012\\$\pm 0.009$} & \ourswin{-5.6} & \shortstack{\textbf{0.956}\\$\pm 0.014$} & \shortstack{1.011\\$\pm 0.020$} & \ourswin{-5.4} & \shortstack{\textbf{1.116}\\$\pm 0.014$} & \shortstack{1.389\\$\pm 0.024$} & \ourswin{-19.6} & \shortstack{\textbf{1.220}\\$\pm 0.007$} & \shortstack{2.666\\$\pm 0.799$} & \ourswin{-54.2} & 1.146 \\
M4 & Y & \shortstack{\textbf{3.537}\\$\pm 0.163$} & \shortstack{4.607\\$\pm 0.401$} & \ourswin{-23.2} & \shortstack{\textbf{3.672}\\$\pm 0.166$} & \shortstack{3.899\\$\pm 0.150$} & \ourswin{-5.8} & \shortstack{\textbf{6.294}\\$\pm 0.350$} & \shortstack{7.422\\$\pm 0.432$} & \ourswin{-15.2} & \shortstack{\textbf{4.044}\\$\pm 0.056$} & \shortstack{7.474\\$\pm 0.569$} & \ourswin{-45.9} & 3.974 \\
 & Q & \shortstack{\textbf{1.327}\\$\pm 0.040$} & \shortstack{1.553\\$\pm 0.038$} & \ourswin{-14.6} & \shortstack{\textbf{1.333}\\$\pm 0.024$} & \shortstack{1.360\\$\pm 0.027$} & \ourswin{-2.0} & \shortstack{\textbf{1.559}\\$\pm 0.029$} & \shortstack{1.956\\$\pm 0.015$} & \ourswin{-20.3} & \shortstack{\textbf{1.603}\\$\pm 0.047$} & \shortstack{3.039\\$\pm 0.697$} & \ourswin{-47.2} & 1.602 \\
 & M & \shortstack{\textbf{1.123}\\$\pm 0.035$} & \shortstack{1.248\\$\pm 0.042$} & \ourswin{-10.0} & \shortstack{\textbf{1.131}\\$\pm 0.028$} & \shortstack{1.160\\$\pm 0.008$} & \ourswin{-2.5} & \shortstack{\textbf{1.294}\\$\pm 0.024$} & \shortstack{1.545\\$\pm 0.016$} & \ourswin{-16.3} & \shortstack{\textbf{1.378}\\$\pm 0.065$} & \shortstack{3.299\\$\pm 0.835$} & \ourswin{-58.2} & 1.260 \\
 & W & \shortstack{\textbf{2.793}\\$\pm 0.210$} & \shortstack{4.823\\$\pm 0.694$} & \ourswin{-42.1} & \shortstack{\textbf{2.615}\\$\pm 0.073$} & \shortstack{3.516\\$\pm 0.149$} & \ourswin{-25.6} & \shortstack{\textbf{3.906}\\$\pm 0.385$} & \shortstack{5.805\\$\pm 0.083$} & \ourswin{-32.7} & \shortstack{\textbf{5.135}\\$\pm 1.455$} & \shortstack{8.625\\$\pm 0.881$} & \ourswin{-40.5} & 2.777 \\
 & D & \shortstack{\textbf{3.765}\\$\pm 0.114$} & \shortstack{5.502\\$\pm 0.406$} & \ourswin{-31.6} & \shortstack{\textbf{3.656}\\$\pm 0.106$} & \shortstack{4.354\\$\pm 0.243$} & \ourswin{-16.0} & \shortstack{\textbf{4.451}\\$\pm 0.236$} & \shortstack{5.743\\$\pm 0.029$} & \ourswin{-22.5} & \shortstack{\textbf{5.375}\\$\pm 0.745$} & \shortstack{11.379\\$\pm 1.590$} & \ourswin{-52.8} & 3.278 \\
 & H & \shortstack{\textbf{1.166}\\$\pm 0.043$} & \shortstack{3.175\\$\pm 0.698$} & \ourswin{-63.3} & \shortstack{\textbf{1.784}\\$\pm 0.076$} & \shortstack{2.302\\$\pm 0.098$} & \ourswin{-22.5} & \shortstack{\textbf{3.894}\\$\pm 1.002$} & \shortstack{6.283\\$\pm 0.832$} & \ourswin{-38.0} & \shortstack{3.066\\$\pm 0.253$} & \shortstack{\textbf{2.786}\\$\pm 0.179$} & \oursloss{+10.0} & 1.193 \\
Tourism & Y & \shortstack{\textbf{3.104}\\$\pm 0.139$} & \shortstack{3.654\\$\pm 0.367$} & \ourswin{-15.0} & \shortstack{\textbf{3.019}\\$\pm 0.050$} & \shortstack{3.173\\$\pm 0.117$} & \ourswin{-4.8} & \shortstack{\textbf{5.873}\\$\pm 0.805$} & \shortstack{6.250\\$\pm 0.512$} & \ourswin{-6.0} & \shortstack{\textbf{3.005}\\$\pm 0.062$} & \shortstack{3.844\\$\pm 0.126$} & \ourswin{-21.8} & 3.552 \\
 & Q & \shortstack{\textbf{1.868}\\$\pm 0.015$} & \shortstack{2.269\\$\pm 0.067$} & \ourswin{-17.7} & \shortstack{\textbf{1.875}\\$\pm 0.022$} & \shortstack{2.067\\$\pm 0.042$} & \ourswin{-9.3} & \shortstack{\textbf{2.363}\\$\pm 0.065$} & \shortstack{2.403\\$\pm 0.032$} & \ourswin{-1.7} & \shortstack{\textbf{2.182}\\$\pm 0.084$} & \shortstack{2.653\\$\pm 0.161$} & \ourswin{-17.8} & 1.699 \\
 & M & \shortstack{\textbf{1.845}\\$\pm 0.030$} & \shortstack{2.269\\$\pm 0.015$} & \ourswin{-18.7} & \shortstack{\textbf{1.834}\\$\pm 0.038$} & \shortstack{2.284\\$\pm 0.039$} & \ourswin{-19.7} & \shortstack{\textbf{2.616}\\$\pm 0.149$} & \shortstack{3.068\\$\pm 0.138$} & \ourswin{-14.7} & \shortstack{\textbf{2.189}\\$\pm 0.056$} & \shortstack{2.650\\$\pm 0.084$} & \ourswin{-17.4} & 1.631 \\
Favorita & D & \shortstack{\textbf{0.816}\\$\pm 0.004$} & \shortstack{0.862\\$\pm 0.011$} & \ourswin{-5.4} & \shortstack{\textbf{0.825}\\$\pm 0.001$} & \shortstack{0.864\\$\pm 0.007$} & \ourswin{-4.5} & \shortstack{\textbf{1.050}\\$\pm 0.015$} & \shortstack{1.360\\$\pm 0.010$} & \ourswin{-22.8} & \shortstack{\textbf{0.889}\\$\pm 0.005$} & \shortstack{0.968\\$\pm 0.026$} & \ourswin{-8.2} & 1.079 \\
GIFT-Eval & Y$^{*}$ & \shortstack{\textbf{3.530}\\$\pm 0.162$} & \shortstack{4.593\\$\pm 0.402$} & \ourswin{-23.2} & \shortstack{\textbf{3.663}\\$\pm 0.165$} & \shortstack{3.888\\$\pm 0.150$} & \ourswin{-5.8} & \shortstack{\textbf{6.278}\\$\pm 0.350$} & \shortstack{7.386\\$\pm 0.435$} & \ourswin{-15.0} & \shortstack{\textbf{4.028}\\$\pm 0.057$} & \shortstack{7.451\\$\pm 0.569$} & \ourswin{-45.9} & 3.966 \\
 & Q$^{*}$ & \shortstack{\textbf{1.327}\\$\pm 0.040$} & \shortstack{1.553\\$\pm 0.038$} & \ourswin{-14.6} & \shortstack{\textbf{1.333}\\$\pm 0.024$} & \shortstack{1.360\\$\pm 0.027$} & \ourswin{-2.0} & \shortstack{\textbf{1.559}\\$\pm 0.029$} & \shortstack{1.956\\$\pm 0.015$} & \ourswin{-20.3} & \shortstack{\textbf{1.603}\\$\pm 0.047$} & \shortstack{3.039\\$\pm 0.697$} & \ourswin{-47.2} & 1.602 \\
 & M & \shortstack{\textbf{1.124}\\$\pm 0.034$} & \shortstack{1.233\\$\pm 0.038$} & \ourswin{-8.9} & \shortstack{\textbf{1.119}\\$\pm 0.025$} & \shortstack{1.141\\$\pm 0.008$} & \ourswin{-1.9} & \shortstack{\textbf{1.288}\\$\pm 0.021$} & \shortstack{1.531\\$\pm 0.015$} & \ourswin{-15.9} & \shortstack{\textbf{1.357}\\$\pm 0.060$} & \shortstack{3.150\\$\pm 0.782$} & \ourswin{-56.9} & 1.249 \\
 & W$^{*}$ & \shortstack{\textbf{1.861}\\$\pm 0.056$} & \shortstack{2.332\\$\pm 0.168$} & \ourswin{-20.2} & \shortstack{\textbf{1.884}\\$\pm 0.057$} & \shortstack{2.032\\$\pm 0.034$} & \ourswin{-7.3} & \shortstack{\textbf{2.217}\\$\pm 0.051$} & \shortstack{2.752\\$\pm 0.070$} & \ourswin{-19.5} & \shortstack{\textbf{2.449}\\$\pm 0.313$} & \shortstack{3.920\\$\pm 0.384$} & \ourswin{-37.5} & 1.916 \\
 & D & \shortstack{\textbf{1.572}\\$\pm 0.012$} & \shortstack{1.798\\$\pm 0.021$} & \ourswin{-12.6} & \shortstack{\textbf{1.598}\\$\pm 0.013$} & \shortstack{1.763\\$\pm 0.016$} & \ourswin{-9.3} & \shortstack{\textbf{2.299}\\$\pm 0.021$} & \shortstack{2.870\\$\pm 0.023$} & \ourswin{-19.9} & \shortstack{\textbf{2.051}\\$\pm 0.017$} & \shortstack{2.497\\$\pm 0.108$} & \ourswin{-17.8} & 2.153 \\
 & H & \shortstack{\textbf{1.393}\\$\pm 0.025$} & \shortstack{1.650\\$\pm 0.038$} & \ourswin{-15.6} & \shortstack{\textbf{1.456}\\$\pm 0.014$} & \shortstack{1.677\\$\pm 0.008$} & \ourswin{-13.2} & \shortstack{\textbf{1.895}\\$\pm 0.022$} & \shortstack{2.571\\$\pm 0.084$} & \ourswin{-26.3} & \shortstack{\textbf{1.640}\\$\pm 0.015$} & \shortstack{1.857\\$\pm 0.057$} & \ourswin{-11.7} & 1.572 \\
 & T$^{*}$ & \shortstack{\textbf{1.129}\\$\pm 0.022$} & \shortstack{1.402\\$\pm 0.015$} & \ourswin{-19.4} & \shortstack{\textbf{1.239}\\$\pm 0.059$} & \shortstack{1.355\\$\pm 0.009$} & \ourswin{-8.5} & \shortstack{\textbf{1.650}\\$\pm 0.039$} & \shortstack{2.103\\$\pm 0.023$} & \ourswin{-21.6} & \shortstack{\textbf{1.610}\\$\pm 0.032$} & \shortstack{1.952\\$\pm 0.071$} & \ourswin{-17.5} & 1.320 \\
 & S$^{*}$ & \shortstack{\textbf{2.241}\\$\pm 0.337$} & \shortstack{3.568\\$\pm 0.443$} & \ourswin{-37.2} & \shortstack{\textbf{2.090}\\$\pm 0.199$} & \shortstack{2.595\\$\pm 0.135$} & \ourswin{-19.5} & \shortstack{\textbf{2.466}\\$\pm 0.139$} & \shortstack{3.807\\$\pm 0.121$} & \ourswin{-35.2} & \shortstack{\textbf{3.808}\\$\pm 0.223$} & \shortstack{4.733\\$\pm 0.262$} & \ourswin{-19.5} & 1.272 \\
\bottomrule
\end{tabular}
}
\end{table}

% Generated by src/scripts/render_paper_tables.py. Do not edit by hand.
\begin{table}[ht]
\centering
\caption{In-domain WQL by benchmark and frequency after 80,000 updates, means with sample standard deviations ($\pm$, below) across 4 seeds. A star marks a base frequency that pools anchored or multiplied GIFT-Eval aliases. \ours uses scale-invariant loss and \SCL uses scale-contaminated loss. These are the only differences between the two runs of each architecture. $\Delta\%$ reports \ours relative to \SCL. \ourswin{Green} denotes an improvement and \oursloss{red} a regression. The better loss is bold within each architecture. References are fitted on the series they forecast and appear behind the double rule.}
\label{tab:tsfm_heldout_indomain-by-frequency-wql}
\renewcommand{\arraystretch}{1.15}
\scriptsize
\setlength{\tabcolsep}{4pt}
\resizebox{\textwidth}{!}{%
\begin{tabular}{llcccccccccccc||c}
\toprule
 &  & \multicolumn{3}{c}{\Chronostwo} & \multicolumn{3}{c}{\Moiraitwo} & \multicolumn{3}{c}{\TimesFMtwo} & \multicolumn{3}{c}{\PatchTST} & \multicolumn{1}{c}{Reference} \\
\cmidrule(lr){3-5}\cmidrule(lr){6-8}\cmidrule(lr){9-11}\cmidrule(lr){12-14}\cmidrule(lr){15-15}
\textbf{Benchmark} & \textbf{Freq.} & \oursShort & \contaminatedShort & $\Delta\%$ & \oursShort & \contaminatedShort & $\Delta\%$ & \oursShort & \contaminatedShort & $\Delta\%$ & \oursShort & \contaminatedShort & $\Delta\%$ & \SNaive \\
\midrule
M1 & Y & \shortstack{\textbf{0.110}\\$\pm 0.006$} & \shortstack{0.126\\$\pm 0.018$} & \ourswin{-12.8} & \shortstack{\textbf{0.154}\\$\pm 0.007$} & \shortstack{0.157\\$\pm 0.006$} & \ourswin{-2.1} & \shortstack{\textbf{0.248}\\$\pm 0.009$} & \shortstack{0.287\\$\pm 0.018$} & \ourswin{-13.6} & \shortstack{\textbf{0.196}\\$\pm 0.009$} & \shortstack{0.307\\$\pm 0.036$} & \ourswin{-36.1} & 0.184 \\
 & Q & \shortstack{\textbf{0.065}\\$\pm 0.008$} & \shortstack{0.076\\$\pm 0.007$} & \ourswin{-13.7} & \shortstack{\textbf{0.094}\\$\pm 0.004$} & \shortstack{0.095\\$\pm 0.008$} & \ourswin{-1.2} & \shortstack{\textbf{0.115}\\$\pm 0.013$} & \shortstack{0.142\\$\pm 0.008$} & \ourswin{-18.9} & \shortstack{\textbf{0.118}\\$\pm 0.009$} & \shortstack{0.148\\$\pm 0.024$} & \ourswin{-20.2} & 0.117 \\
 & M & \shortstack{0.162\\$\pm 0.016$} & \shortstack{\textbf{0.131}\\$\pm 0.007$} & \oursloss{+23.5} & \shortstack{0.207\\$\pm 0.024$} & \shortstack{\textbf{0.188}\\$\pm 0.021$} & \oursloss{+10.0} & \shortstack{\textbf{0.227}\\$\pm 0.006$} & \shortstack{0.261\\$\pm 0.021$} & \ourswin{-12.9} & \shortstack{\textbf{0.272}\\$\pm 0.015$} & \shortstack{0.322\\$\pm 0.066$} & \ourswin{-15.7} & 0.150 \\
M3 & Y & \shortstack{\textbf{0.101}\\$\pm 0.001$} & \shortstack{0.112\\$\pm 0.006$} & \ourswin{-10.1} & \shortstack{\textbf{0.132}\\$\pm 0.004$} & \shortstack{0.143\\$\pm 0.004$} & \ourswin{-7.3} & \shortstack{\textbf{0.218}\\$\pm 0.009$} & \shortstack{0.221\\$\pm 0.006$} & \ourswin{-1.3} & \shortstack{\textbf{0.170}\\$\pm 0.010$} & \shortstack{0.283\\$\pm 0.039$} & \ourswin{-40.0} & 0.118 \\
 & Q & \shortstack{\textbf{0.063}\\$\pm 0.003$} & \shortstack{0.066\\$\pm 0.002$} & \ourswin{-4.8} & \shortstack{\textbf{0.079}\\$\pm 0.003$} & \shortstack{0.080\\$\pm 0.003$} & \ourswin{-0.5} & \shortstack{\textbf{0.104}\\$\pm 0.001$} & \shortstack{0.126\\$\pm 0.007$} & \ourswin{-17.7} & \shortstack{\textbf{0.110}\\$\pm 0.006$} & \shortstack{0.202\\$\pm 0.034$} & \ourswin{-45.4} & 0.082 \\
 & M & \shortstack{\textbf{0.085}\\$\pm 0.001$} & \shortstack{0.086\\$\pm 0.002$} & \ourswin{-0.5} & \shortstack{\textbf{0.110}\\$\pm 0.002$} & \shortstack{0.115\\$\pm 0.002$} & \ourswin{-3.8} & \shortstack{\textbf{0.151}\\$\pm 0.002$} & \shortstack{0.177\\$\pm 0.003$} & \ourswin{-14.6} & \shortstack{\textbf{0.167}\\$\pm 0.005$} & \shortstack{0.262\\$\pm 0.055$} & \ourswin{-36.4} & 0.121 \\
M4 & Y & \shortstack{\textbf{0.101}\\$\pm 0.003$} & \shortstack{0.119\\$\pm 0.005$} & \ourswin{-15.2} & \shortstack{0.136\\$\pm 0.004$} & \shortstack{\textbf{0.134}\\$\pm 0.004$} & \oursloss{+1.5} & \shortstack{\textbf{0.256}\\$\pm 0.013$} & \shortstack{0.287\\$\pm 0.015$} & \ourswin{-11.0} & \shortstack{\textbf{0.166}\\$\pm 0.001$} & \shortstack{0.258\\$\pm 0.005$} & \ourswin{-35.5} & 0.137 \\
 & Q & \shortstack{\textbf{0.067}\\$\pm 0.002$} & \shortstack{0.074\\$\pm 0.002$} & \ourswin{-8.3} & \shortstack{0.089\\$\pm 0.002$} & \shortstack{\textbf{0.088}\\$\pm 0.001$} & \oursloss{+1.1} & \shortstack{\textbf{0.119}\\$\pm 0.001$} & \shortstack{0.141\\$\pm 0.001$} & \ourswin{-15.2} & \shortstack{\textbf{0.119}\\$\pm 0.003$} & \shortstack{0.175\\$\pm 0.021$} & \ourswin{-32.0} & 0.098 \\
 & M & \shortstack{0.090\\$\pm 0.004$} & \shortstack{\textbf{0.088}\\$\pm 0.001$} & \oursloss{+2.4} & \shortstack{0.117\\$\pm 0.004$} & \shortstack{\textbf{0.111}\\$\pm 0.002$} & \oursloss{+5.0} & \shortstack{\textbf{0.149}\\$\pm 0.001$} & \shortstack{0.167\\$\pm 0.001$} & \ourswin{-10.3} & \shortstack{\textbf{0.162}\\$\pm 0.009$} & \shortstack{0.258\\$\pm 0.039$} & \ourswin{-37.4} & 0.122 \\
 & W & \shortstack{\textbf{0.041}\\$\pm 0.000$} & \shortstack{0.055\\$\pm 0.004$} & \ourswin{-25.1} & \shortstack{\textbf{0.054}\\$\pm 0.001$} & \shortstack{0.061\\$\pm 0.003$} & \ourswin{-11.7} & \shortstack{\textbf{0.078}\\$\pm 0.004$} & \shortstack{0.100\\$\pm 0.001$} & \ourswin{-21.9} & \shortstack{\textbf{0.089}\\$\pm 0.013$} & \shortstack{0.125\\$\pm 0.007$} & \ourswin{-29.0} & 0.061 \\
 & D & \shortstack{\textbf{0.020}\\$\pm 0.001$} & \shortstack{0.029\\$\pm 0.002$} & \ourswin{-32.4} & \shortstack{\textbf{0.025}\\$\pm 0.001$} & \shortstack{0.030\\$\pm 0.002$} & \ourswin{-17.5} & \shortstack{\textbf{0.037}\\$\pm 0.002$} & \shortstack{0.048\\$\pm 0.000$} & \ourswin{-22.5} & \shortstack{\textbf{0.044}\\$\pm 0.006$} & \shortstack{0.079\\$\pm 0.007$} & \ourswin{-43.9} & 0.024 \\
 & H & \shortstack{\textbf{0.019}\\$\pm 0.001$} & \shortstack{0.023\\$\pm 0.001$} & \ourswin{-17.5} & \shortstack{\textbf{0.026}\\$\pm 0.001$} & \shortstack{0.029\\$\pm 0.004$} & \ourswin{-11.0} & \shortstack{\textbf{0.057}\\$\pm 0.010$} & \shortstack{0.076\\$\pm 0.009$} & \ourswin{-24.4} & \shortstack{\textbf{0.045}\\$\pm 0.007$} & \shortstack{0.048\\$\pm 0.003$} & \ourswin{-6.7} & 0.038 \\
Tourism & Y & \shortstack{\textbf{0.115}\\$\pm 0.010$} & \shortstack{0.151\\$\pm 0.014$} & \ourswin{-23.8} & \shortstack{\textbf{0.153}\\$\pm 0.005$} & \shortstack{0.171\\$\pm 0.010$} & \ourswin{-10.4} & \shortstack{\textbf{0.332}\\$\pm 0.029$} & \shortstack{0.342\\$\pm 0.042$} & \ourswin{-3.0} & \shortstack{\textbf{0.178}\\$\pm 0.012$} & \shortstack{0.233\\$\pm 0.009$} & \ourswin{-23.6} & 0.170 \\
 & Q & \shortstack{\textbf{0.056}\\$\pm 0.002$} & \shortstack{0.077\\$\pm 0.006$} & \ourswin{-26.3} & \shortstack{\textbf{0.083}\\$\pm 0.018$} & \shortstack{0.094\\$\pm 0.006$} & \ourswin{-11.4} & \shortstack{\textbf{0.115}\\$\pm 0.002$} & \shortstack{0.120\\$\pm 0.003$} & \ourswin{-4.0} & \shortstack{\textbf{0.110}\\$\pm 0.008$} & \shortstack{0.150\\$\pm 0.017$} & \ourswin{-27.1} & 0.098 \\
 & M & \shortstack{\textbf{0.077}\\$\pm 0.004$} & \shortstack{0.100\\$\pm 0.003$} & \ourswin{-22.7} & \shortstack{\textbf{0.102}\\$\pm 0.003$} & \shortstack{0.121\\$\pm 0.003$} & \ourswin{-15.8} & \shortstack{\textbf{0.208}\\$\pm 0.018$} & \shortstack{0.330\\$\pm 0.041$} & \ourswin{-36.9} & \shortstack{\textbf{0.153}\\$\pm 0.007$} & \shortstack{0.211\\$\pm 0.009$} & \ourswin{-27.3} & 0.086 \\
Favorita & D & \shortstack{\textbf{0.297}\\$\pm 0.004$} & \shortstack{0.326\\$\pm 0.005$} & \ourswin{-9.1} & \shortstack{\textbf{0.387}\\$\pm 0.002$} & \shortstack{0.422\\$\pm 0.004$} & \ourswin{-8.3} & \shortstack{\textbf{0.638}\\$\pm 0.011$} & \shortstack{0.848\\$\pm 0.007$} & \ourswin{-24.8} & \shortstack{\textbf{0.540}\\$\pm 0.004$} & \shortstack{0.595\\$\pm 0.019$} & \ourswin{-9.3} & 0.614 \\
GIFT-Eval & Y$^{*}$ & \shortstack{\textbf{0.101}\\$\pm 0.003$} & \shortstack{0.119\\$\pm 0.005$} & \ourswin{-15.2} & \shortstack{0.136\\$\pm 0.004$} & \shortstack{\textbf{0.134}\\$\pm 0.004$} & \oursloss{+1.5} & \shortstack{\textbf{0.256}\\$\pm 0.013$} & \shortstack{0.287\\$\pm 0.015$} & \ourswin{-11.0} & \shortstack{\textbf{0.166}\\$\pm 0.001$} & \shortstack{0.258\\$\pm 0.005$} & \ourswin{-35.5} & 0.137 \\
 & Q$^{*}$ & \shortstack{\textbf{0.067}\\$\pm 0.002$} & \shortstack{0.074\\$\pm 0.002$} & \ourswin{-8.3} & \shortstack{0.089\\$\pm 0.002$} & \shortstack{\textbf{0.088}\\$\pm 0.001$} & \oursloss{+1.1} & \shortstack{\textbf{0.119}\\$\pm 0.001$} & \shortstack{0.141\\$\pm 0.001$} & \ourswin{-15.2} & \shortstack{\textbf{0.119}\\$\pm 0.003$} & \shortstack{0.175\\$\pm 0.021$} & \ourswin{-32.0} & 0.098 \\
 & M & \shortstack{0.146\\$\pm 0.005$} & \shortstack{\textbf{0.132}\\$\pm 0.002$} & \oursloss{+11.0} & \shortstack{0.175\\$\pm 0.002$} & \shortstack{\textbf{0.160}\\$\pm 0.001$} & \oursloss{+9.5} & \shortstack{\textbf{0.227}\\$\pm 0.001$} & \shortstack{0.251\\$\pm 0.002$} & \ourswin{-9.7} & \shortstack{\textbf{0.229}\\$\pm 0.007$} & \shortstack{0.317\\$\pm 0.037$} & \ourswin{-27.5} & 0.194 \\
 & W$^{*}$ & \shortstack{\textbf{0.125}\\$\pm 0.006$} & \shortstack{0.141\\$\pm 0.004$} & \ourswin{-11.8} & \shortstack{\textbf{0.163}\\$\pm 0.004$} & \shortstack{0.172\\$\pm 0.003$} & \ourswin{-5.0} & \shortstack{\textbf{0.253}\\$\pm 0.004$} & \shortstack{0.290\\$\pm 0.009$} & \ourswin{-12.9} & \shortstack{\textbf{0.257}\\$\pm 0.007$} & \shortstack{0.331\\$\pm 0.022$} & \ourswin{-22.2} & 0.268 \\
 & D & \shortstack{\textbf{0.410}\\$\pm 0.000$} & \shortstack{0.468\\$\pm 0.007$} & \ourswin{-12.5} & \shortstack{\textbf{0.533}\\$\pm 0.002$} & \shortstack{0.592\\$\pm 0.003$} & \ourswin{-10.0} & \shortstack{\textbf{0.888}\\$\pm 0.006$} & \shortstack{1.088\\$\pm 0.008$} & \ourswin{-18.4} & \shortstack{\textbf{0.771}\\$\pm 0.003$} & \shortstack{0.847\\$\pm 0.028$} & \ourswin{-9.0} & 1.182 \\
 & H & \shortstack{\textbf{0.247}\\$\pm 0.014$} & \shortstack{0.275\\$\pm 0.006$} & \ourswin{-10.1} & \shortstack{\textbf{0.339}\\$\pm 0.012$} & \shortstack{0.360\\$\pm 0.011$} & \ourswin{-5.8} & \shortstack{\textbf{0.496}\\$\pm 0.017$} & \shortstack{0.657\\$\pm 0.029$} & \ourswin{-24.5} & \shortstack{\textbf{0.426}\\$\pm 0.011$} & \shortstack{0.461\\$\pm 0.012$} & \ourswin{-7.7} & 0.425 \\
 & T$^{*}$ & \shortstack{\textbf{0.346}\\$\pm 0.018$} & \shortstack{0.398\\$\pm 0.009$} & \ourswin{-13.0} & \shortstack{\textbf{0.454}\\$\pm 0.010$} & \shortstack{0.481\\$\pm 0.005$} & \ourswin{-5.5} & \shortstack{\textbf{0.633}\\$\pm 0.014$} & \shortstack{0.718\\$\pm 0.012$} & \ourswin{-11.8} & \shortstack{\textbf{0.632}\\$\pm 0.015$} & \shortstack{0.732\\$\pm 0.026$} & \ourswin{-13.7} & 0.868 \\
 & S$^{*}$ & \shortstack{\textbf{0.028}\\$\pm 0.005$} & \shortstack{0.039\\$\pm 0.006$} & \ourswin{-28.9} & \shortstack{\textbf{0.034}\\$\pm 0.005$} & \shortstack{0.034\\$\pm 0.004$} & \ourswin{-1.0} & \shortstack{\textbf{0.035}\\$\pm 0.002$} & \shortstack{0.069\\$\pm 0.004$} & \ourswin{-49.9} & \shortstack{\textbf{0.064}\\$\pm 0.005$} & \shortstack{0.076\\$\pm 0.004$} & \ourswin{-16.4} & 0.040 \\
\bottomrule
\end{tabular}
}
\end{table}

 \clearpage
\section{Supervised Forecasting Details}
\label{apd:supervised-breakdown}

\paragraph{Data and pooling.}
The experiment pools eligible M1, M3, M4, and Tourism series by frequency. One model is trained for each of the yearly, quarterly, monthly, weekly, daily, and hourly pools. M1, M3, and Tourism contribute only their available yearly, quarterly, and monthly series, while M4 contributes all six frequencies. The model horizon is the largest official horizon in its pool. The input context is twice that horizon. Table~\ref{tab:supervised-frequency-setup} reports the resulting dimensions and eligible population.

\begin{table}[ht]
\centering
\caption{Frequency-specific dimensions and eligible training populations in the supervised experiment. The population is fixed before comparing loss spaces and is shared by the neural and statistical methods.}
\label{tab:supervised-frequency-setup}
\small
\setlength{\tabcolsep}{5pt}
\begin{tabular}{lrrr}
\toprule
\textbf{Frequency} & \textbf{Horizon} & \textbf{Context} & \textbf{Eligible series} \\
\midrule
Yearly & 8 & 16 & 7,841 \\
Quarterly & 8 & 16 & 23,239 \\
Monthly & 24 & 48 & 34,183 \\
Weekly & 13 & 26 & 359 \\
Daily & 14 & 28 & 4,227 \\
Hourly & 48 & 96 & 414 \\
\bottomrule
\end{tabular}
\end{table}

For a series with official horizon $H$, the final $2H-1$ observations form the rolling test region. The immediately preceding model-horizon block is used for validation. A series is eligible only when a complete input context and training horizon remain before validation. Evaluation uses all $H$ rolling origins in the test region. At origin $j$, only observations through the end of that origin's input context are available to the model, scaler, and fitted references.

\paragraph{Loss-space comparison.}
Both conditions receive inputs under standard scaling whose offset $a$ and denominator $b$ are estimated from the input context. The scale-invariant loss (\ours) computes MAE between predictions and targets after this scaling. The scale-contaminated loss (\contaminated) applies the same fixed inverse transform before computing MAE in original scale. The scaler has no learnable affine parameters in either condition, so matched runs differ only in the space where MAE is evaluated.

\paragraph{Optimization and checkpointing.}
Every model trains from scratch for exactly $20{,}000$ optimizer updates. Training does not stop early. Validation is evaluated every $250$ updates, and the best checkpoint is the state with the lowest validation MAE in original scale. All results average seeds $0$, $1$, $2$, and $3$. Table~\ref{tab:supervised-shared-hyperparameters} gives the shared settings.

\begin{table}[ht]
\centering
\caption{Shared supervised training and evaluation settings.}
\label{tab:supervised-shared-hyperparameters}
\small
\setlength{\tabcolsep}{6pt}
\begin{tabular}{ll}
\toprule
\textbf{Setting} & \textbf{Value} \\
\midrule
Optimizer & Adam \\
Training updates & 20,000 \\
Series batch size & 32 \\
Sampled windows per batch & 512 \\
Learning rate & $10^{-3}$ \\
Learning-rate decays & 3, evenly spaced \\
Validation interval & 250 updates \\
Early stopping & Disabled \\
Training loss & MAE \\
Input scaler & Standard scaling from the input context \\
Start padding & Disabled \\
Forecast type & Point forecast \\
Random seeds & $0,1,2,3$ \\
\bottomrule
\end{tabular}
\end{table}

We use NeuralForecast 3.2.1 \citep{olivares2022neuralforecast} and retain its architecture defaults except for the shared settings above, the frequency-specific horizon and context. The architecture parameters that materially determine model capacity are listed in Table~\ref{tab:supervised-architecture-hyperparameters}.

\begin{table}[ht]
\centering
\caption{Architecture-specific parameters for the supervised experiment. Parameters not shown use the NeuralForecast 3.2.1 defaults.}
\label{tab:supervised-architecture-hyperparameters}
\small
\setlength{\tabcolsep}{5pt}
\begin{tabular}{lp{0.72\textwidth}}
\toprule
\textbf{Architecture} & \textbf{Parameters} \\
\midrule
\NHITS & Three identity stacks, one block per stack, two hidden layers of 512 units per block, pooling kernels $(2,2,1)$, frequency downsampling $(4,2,1)$, ReLU activation, and linear interpolation. \\
\NBEATS & Identity, trend, and seasonality stacks, one block per stack, two hidden layers of 512 units per block, ReLU activation, and no shared block weights. \\
\PatchTST & Three encoder layers, 16 attention heads, hidden width 128, feed-forward width 256, patch length 16, stride 8, dropout 0.2, GELU activation, and internal RevIN disabled. \\
\model{TCN} & Kernel size 2, dilations $(1,2,4,8,16)$, encoder width 128, context width 10, and a two-layer decoder of width 128. \\
\model{LSTM} & Two encoder layers of width 128, no encoder dropout, and a two-layer decoder of width 128. Forecasts use the direct decoder rather than recursive generation. \\
\bottomrule
\end{tabular}
\end{table}

\paragraph{Metrics and aggregation.}
MASE is computed for each series-origin forecast using the seasonal-naive in-sample MAE from all history available at that origin. WQL is the ratio of summed median quantile loss to summed absolute targets within each benchmark-frequency pool. For each neural model, metrics are first computed within a seed. The main table macro-averages eligible frequencies within each benchmark and then averages the four seed values. Statistical references have no seed dimension. \SNaive{}, \ETS~\citep{holt1957exponential_smoothing}, and \ARIMA~\citep{hyndman2008automatic_arima, hyndman2026forecasting} are refitted at every rolling origin using the same available history and exactly the same eligible series as the neural models. We use the baselines implementation available in StatsForecast 2.1.1~\citep{garza2022statsforecast}.

\paragraph{Compute.} All runs were performed with a single NVIDIA GeForce RTX 2080 Ti GPU. Generating
Table~\ref{tab:supervised-mseries-benchmark} used 51 GPU-hours in total.

\paragraph{Benchmark-frequency results.}
Tables~\ref{tab:supervised-mseries-benchmark-frequency-mase} and~\ref{tab:supervised-mseries-benchmark-frequency-wql} give the full breakdown underlying the benchmark aggregates in Table~\ref{tab:supervised-mseries-benchmark}.

The frequency-level results support the same findings that are in the main paper and somewhat strengthen them. \ours lowers MASE in 49 of 75 architecture-benchmark-frequency comparisons and WQL in 43 of 75. The effect is strongest at the lower frequencies. It wins 17 of 20 yearly and 15 of 20 quarterly comparisons on MASE, with corresponding WQL counts of 15 and 14. Monthly results split 8 to 12 for both metrics, while the weekly, daily, and hourly M4 pools are also mixed. The benchmark-level advantage is therefore broad across architectures but concentrated in the yearly and quarterly pools. The detailed tables also show why the statistical references remain important. \ETS{} or \ARIMA{} often leads on M3 and the higher-frequency M4 pools even when \ours improves the matched neural model.

% Generated by src/scripts/render_updated_m_supervised_appendix.py. Do not edit by hand.
\begin{table}[p]
\centering
\caption{Best-checkpoint MASE by benchmark and frequency in the supervised experiment. Neural values are mean $\pm$ sample standard deviation across four seeds. \ours uses scale-invariant MAE and \SCL uses scale-contaminated MAE, both with the same input-window standard scaling. References are fitted at every rolling origin on the same eligible series. Lower is better, and bold marks the better loss within each matched pair.}
\label{tab:supervised-mseries-benchmark-frequency-mase}
\tiny
\setlength{\tabcolsep}{3pt}
\begin{tabular}{lllcc||ccc}
\toprule
\textbf{Architecture} & \textbf{Benchmark} & \textbf{Frequency} & \ours & \SCL & \SNaive & \ETS & \ARIMA \\
\midrule
\multirow{15}{*}{\model{N-HiTS}} & \multirow{3}{*}{M1} & Y & \textbf{2.368} $\pm 0.051$ & 2.461 $\pm 0.171$ & 3.046 & 2.476 & 2.632 \\
 &  & Q & 1.859 $\pm 0.004$ & \textbf{1.857} $\pm 0.012$ & 2.527 & 1.899 & 1.966 \\
 &  & M & 1.251 $\pm 0.012$ & \textbf{1.248} $\pm 0.009$ & 1.859 & 1.338 & 1.333 \\
 & \multirow{3}{*}{M3} & Y & 2.972 $\pm 0.346$ & \textbf{2.934} $\pm 0.092$ & 3.333 & 2.384 & 3.030 \\
 &  & Q & \textbf{1.121} $\pm 0.006$ & 1.127 $\pm 0.014$ & 1.457 & 1.115 & 1.130 \\
 &  & M & 0.863 $\pm 0.012$ & \textbf{0.860} $\pm 0.009$ & 1.243 & 0.932 & 0.901 \\
 & \multirow{6}{*}{M4} & Y & \textbf{3.264} $\pm 0.036$ & 3.324 $\pm 0.083$ & 4.557 & 3.250 & 4.754 \\
 &  & Q & \textbf{1.192} $\pm 0.002$ & 1.199 $\pm 0.003$ & 1.590 & 1.196 & 1.207 \\
 &  & M & \textbf{0.917} $\pm 0.005$ & 0.919 $\pm 0.003$ & 1.301 & 0.956 & 0.938 \\
 &  & W & 2.477 $\pm 0.051$ & \textbf{2.466} $\pm 0.055$ & 2.799 & 2.471 & 2.843 \\
 &  & D & \textbf{3.160} $\pm 0.006$ & 3.167 $\pm 0.007$ & 3.207 & 3.178 & 3.234 \\
 &  & H & \textbf{0.937} $\pm 0.066$ & 0.966 $\pm 0.074$ & 1.220 & 1.127 & 1.133 \\
 & \multirow{3}{*}{Tourism} & Y & \textbf{3.689} $\pm 0.112$ & 3.878 $\pm 0.346$ & 2.956 & 3.460 & 4.135 \\
 &  & Q & \textbf{1.689} $\pm 0.010$ & 1.735 $\pm 0.047$ & 1.924 & 1.701 & 1.750 \\
 &  & M & 1.658 $\pm 0.025$ & \textbf{1.647} $\pm 0.005$ & 1.817 & 1.615 & 1.623 \\
\midrule
\multirow{15}{*}{\model{N-BEATS}} & \multirow{3}{*}{M1} & Y & \textbf{2.356} $\pm 0.030$ & 2.386 $\pm 0.084$ & 3.046 & 2.476 & 2.632 \\
 &  & Q & \textbf{1.858} $\pm 0.018$ & 1.859 $\pm 0.007$ & 2.527 & 1.899 & 1.966 \\
 &  & M & \textbf{1.252} $\pm 0.005$ & 1.259 $\pm 0.012$ & 1.859 & 1.338 & 1.333 \\
 & \multirow{3}{*}{M3} & Y & \textbf{3.059} $\pm 0.146$ & 3.327 $\pm 0.449$ & 3.333 & 2.384 & 3.030 \\
 &  & Q & 1.120 $\pm 0.016$ & \textbf{1.118} $\pm 0.010$ & 1.457 & 1.115 & 1.130 \\
 &  & M & \textbf{0.853} $\pm 0.006$ & 0.856 $\pm 0.004$ & 1.243 & 0.932 & 0.901 \\
 & \multirow{6}{*}{M4} & Y & \textbf{3.246} $\pm 0.034$ & 3.448 $\pm 0.300$ & 4.557 & 3.250 & 4.754 \\
 &  & Q & \textbf{1.188} $\pm 0.001$ & 1.191 $\pm 0.002$ & 1.590 & 1.196 & 1.207 \\
 &  & M & \textbf{0.914} $\pm 0.001$ & 0.915 $\pm 0.004$ & 1.301 & 0.956 & 0.938 \\
 &  & W & \textbf{2.479} $\pm 0.021$ & 2.497 $\pm 0.045$ & 2.799 & 2.471 & 2.843 \\
 &  & D & \textbf{3.164} $\pm 0.003$ & \textbf{3.164} $\pm 0.007$ & 3.207 & 3.178 & 3.234 \\
 &  & H & \textbf{0.933} $\pm 0.015$ & 1.001 $\pm 0.046$ & 1.220 & 1.127 & 1.133 \\
 & \multirow{3}{*}{Tourism} & Y & \textbf{3.580} $\pm 0.116$ & 3.847 $\pm 0.073$ & 2.956 & 3.460 & 4.135 \\
 &  & Q & 1.704 $\pm 0.007$ & \textbf{1.695} $\pm 0.014$ & 1.924 & 1.701 & 1.750 \\
 &  & M & 1.637 $\pm 0.023$ & \textbf{1.635} $\pm 0.019$ & 1.817 & 1.615 & 1.623 \\
\midrule
\multirow{15}{*}{\PatchTST} & \multirow{3}{*}{M1} & Y & \textbf{2.167} $\pm 0.104$ & 2.194 $\pm 0.115$ & 3.046 & 2.476 & 2.632 \\
 &  & Q & 2.056 $\pm 0.079$ & \textbf{2.018} $\pm 0.040$ & 2.527 & 1.899 & 1.966 \\
 &  & M & 1.527 $\pm 0.025$ & \textbf{1.520} $\pm 0.033$ & 1.859 & 1.338 & 1.333 \\
 & \multirow{3}{*}{M3} & Y & \textbf{6.081} $\pm 0.790$ & 6.911 $\pm 1.352$ & 3.333 & 2.384 & 3.030 \\
 &  & Q & \textbf{1.337} $\pm 0.021$ & 1.351 $\pm 0.030$ & 1.457 & 1.115 & 1.130 \\
 &  & M & 0.989 $\pm 0.017$ & \textbf{0.988} $\pm 0.018$ & 1.243 & 0.932 & 0.901 \\
 & \multirow{6}{*}{M4} & Y & \textbf{3.807} $\pm 0.242$ & 3.867 $\pm 0.282$ & 4.557 & 3.250 & 4.754 \\
 &  & Q & 1.352 $\pm 0.015$ & \textbf{1.341} $\pm 0.025$ & 1.590 & 1.196 & 1.207 \\
 &  & M & 1.029 $\pm 0.019$ & \textbf{1.027} $\pm 0.022$ & 1.301 & 0.956 & 0.938 \\
 &  & W & \textbf{3.929} $\pm 0.098$ & \textbf{3.929} $\pm 0.214$ & 2.799 & 2.471 & 2.843 \\
 &  & D & 4.198 $\pm 0.145$ & \textbf{4.152} $\pm 0.189$ & 3.207 & 3.178 & 3.234 \\
 &  & H & 7.518 $\pm 1.134$ & \textbf{7.118} $\pm 2.043$ & 1.220 & 1.127 & 1.133 \\
 & \multirow{3}{*}{Tourism} & Y & \textbf{3.283} $\pm 0.500$ & 3.487 $\pm 0.483$ & 2.956 & 3.460 & 4.135 \\
 &  & Q & \textbf{3.168} $\pm 0.087$ & 3.188 $\pm 0.142$ & 1.924 & 1.701 & 1.750 \\
 &  & M & \textbf{2.200} $\pm 0.100$ & 2.204 $\pm 0.083$ & 1.817 & 1.615 & 1.623 \\
\midrule
\multirow{15}{*}{\model{TCN}} & \multirow{3}{*}{M1} & Y & \textbf{2.467} $\pm 0.127$ & 2.606 $\pm 0.123$ & 3.046 & 2.476 & 2.632 \\
 &  & Q & \textbf{1.903} $\pm 0.014$ & 1.936 $\pm 0.052$ & 2.527 & 1.899 & 1.966 \\
 &  & M & 1.452 $\pm 0.013$ & \textbf{1.419} $\pm 0.011$ & 1.859 & 1.338 & 1.333 \\
 & \multirow{3}{*}{M3} & Y & \textbf{3.570} $\pm 0.438$ & 4.035 $\pm 0.372$ & 3.333 & 2.384 & 3.030 \\
 &  & Q & \textbf{1.160} $\pm 0.029$ & 1.202 $\pm 0.042$ & 1.457 & 1.115 & 1.130 \\
 &  & M & 0.957 $\pm 0.008$ & \textbf{0.943} $\pm 0.007$ & 1.243 & 0.932 & 0.901 \\
 & \multirow{6}{*}{M4} & Y & \textbf{3.636} $\pm 0.147$ & 3.751 $\pm 0.151$ & 4.557 & 3.250 & 4.754 \\
 &  & Q & \textbf{1.229} $\pm 0.011$ & 1.268 $\pm 0.044$ & 1.590 & 1.196 & 1.207 \\
 &  & M & 1.003 $\pm 0.008$ & \textbf{0.991} $\pm 0.007$ & 1.301 & 0.956 & 0.938 \\
 &  & W & \textbf{2.468} $\pm 0.024$ & 2.545 $\pm 0.100$ & 2.799 & 2.471 & 2.843 \\
 &  & D & 3.171 $\pm 0.006$ & \textbf{3.169} $\pm 0.014$ & 3.207 & 3.178 & 3.234 \\
 &  & H & \textbf{1.035} $\pm 0.033$ & 1.465 $\pm 0.258$ & 1.220 & 1.127 & 1.133 \\
 & \multirow{3}{*}{Tourism} & Y & \textbf{3.951} $\pm 0.371$ & 3.976 $\pm 0.494$ & 2.956 & 3.460 & 4.135 \\
 &  & Q & \textbf{1.963} $\pm 0.127$ & 2.062 $\pm 0.219$ & 1.924 & 1.701 & 1.750 \\
 &  & M & 2.089 $\pm 0.073$ & \textbf{2.026} $\pm 0.069$ & 1.817 & 1.615 & 1.623 \\
\midrule
\multirow{15}{*}{\model{LSTM}} & \multirow{3}{*}{M1} & Y & \textbf{2.477} $\pm 0.044$ & 2.607 $\pm 0.124$ & 3.046 & 2.476 & 2.632 \\
 &  & Q & \textbf{2.000} $\pm 0.023$ & 2.042 $\pm 0.016$ & 2.527 & 1.899 & 1.966 \\
 &  & M & \textbf{1.516} $\pm 0.007$ & 1.521 $\pm 0.009$ & 1.859 & 1.338 & 1.333 \\
 & \multirow{3}{*}{M3} & Y & 4.113 $\pm 0.157$ & \textbf{4.104} $\pm 0.717$ & 3.333 & 2.384 & 3.030 \\
 &  & Q & \textbf{1.235} $\pm 0.046$ & 1.246 $\pm 0.031$ & 1.457 & 1.115 & 1.130 \\
 &  & M & \textbf{1.050} $\pm 0.004$ & 1.053 $\pm 0.005$ & 1.243 & 0.932 & 0.901 \\
 & \multirow{6}{*}{M4} & Y & \textbf{3.877} $\pm 0.149$ & 3.909 $\pm 0.168$ & 4.557 & 3.250 & 4.754 \\
 &  & Q & \textbf{1.293} $\pm 0.002$ & 1.303 $\pm 0.009$ & 1.590 & 1.196 & 1.207 \\
 &  & M & \textbf{1.056} $\pm 0.008$ & 1.059 $\pm 0.007$ & 1.301 & 0.956 & 0.938 \\
 &  & W & \textbf{2.780} $\pm 0.041$ & 2.813 $\pm 0.107$ & 2.799 & 2.471 & 2.843 \\
 &  & D & \textbf{3.492} $\pm 0.010$ & 3.494 $\pm 0.009$ & 3.207 & 3.178 & 3.234 \\
 &  & H & 1.087 $\pm 0.074$ & \textbf{1.030} $\pm 0.024$ & 1.220 & 1.127 & 1.133 \\
 & \multirow{3}{*}{Tourism} & Y & 4.238 $\pm 0.531$ & \textbf{4.077} $\pm 0.137$ & 2.956 & 3.460 & 4.135 \\
 &  & Q & \textbf{1.864} $\pm 0.009$ & 1.939 $\pm 0.056$ & 1.924 & 1.701 & 1.750 \\
 &  & M & 1.827 $\pm 0.009$ & \textbf{1.826} $\pm 0.015$ & 1.817 & 1.615 & 1.623 \\
\bottomrule
\end{tabular}
\end{table}

% Generated by src/scripts/render_updated_m_supervised_appendix.py. Do not edit by hand.
\begin{table}[p]
\centering
\caption{Best-checkpoint WQL by benchmark and frequency in the supervised experiment. Neural values are mean $\pm$ sample standard deviation across four seeds. \ours uses scale-invariant MAE and \SCL uses scale-contaminated MAE, both with the same input-window standard scaling. References are fitted at every rolling origin on the same eligible series. Lower is better, and bold marks the better loss within each matched pair.}
\label{tab:supervised-mseries-benchmark-frequency-wql}
\tiny
\setlength{\tabcolsep}{3pt}
\begin{tabular}{lllcc||ccc}
\toprule
\textbf{Architecture} & \textbf{Benchmark} & \textbf{Frequency} & \ours & \SCL & \SNaive & \ETS & \ARIMA \\
\midrule
\multirow{15}{*}{\model{N-HiTS}} & \multirow{3}{*}{M1} & Y & 0.119 $\pm 0.003$ & \textbf{0.117} $\pm 0.007$ & 0.219 & 0.144 & 0.204 \\
 &  & Q & \textbf{0.096} $\pm 0.002$ & \textbf{0.096} $\pm 0.001$ & 0.139 & 0.094 & 0.106 \\
 &  & M & 0.083 $\pm 0.001$ & \textbf{0.081} $\pm 0.003$ & 0.144 & 0.088 & 0.086 \\
 & \multirow{3}{*}{M3} & Y & \textbf{0.105} $\pm 0.005$ & 0.107 $\pm 0.006$ & 0.105 & 0.093 & 0.100 \\
 &  & Q & \textbf{0.081} $\pm 0.001$ & 0.082 $\pm 0.001$ & 0.092 & 0.079 & 0.081 \\
 &  & M & \textbf{0.083} $\pm 0.001$ & \textbf{0.083} $\pm 0.001$ & 0.111 & 0.089 & 0.088 \\
 & \multirow{6}{*}{M4} & Y & \textbf{0.120} $\pm 0.001$ & 0.123 $\pm 0.004$ & 0.146 & 0.118 & 0.152 \\
 &  & Q & \textbf{0.099} $\pm 0.000$ & \textbf{0.099} $\pm 0.000$ & 0.118 & 0.097 & 0.100 \\
 &  & M & \textbf{0.077} $\pm 0.000$ & \textbf{0.077} $\pm 0.000$ & 0.102 & 0.080 & 0.080 \\
 &  & W & \textbf{0.061} $\pm 0.002$ & \textbf{0.061} $\pm 0.002$ & 0.066 & 0.064 & 0.064 \\
 &  & D & \textbf{0.026} $\pm 0.000$ & \textbf{0.026} $\pm 0.000$ & 0.027 & 0.026 & 0.027 \\
 &  & H & 0.048 $\pm 0.001$ & \textbf{0.046} $\pm 0.001$ & 0.039 & 0.042 & 0.050 \\
 & \multirow{3}{*}{Tourism} & Y & \textbf{0.146} $\pm 0.004$ & 0.156 $\pm 0.016$ & 0.125 & 0.138 & 0.168 \\
 &  & Q & \textbf{0.113} $\pm 0.001$ & 0.115 $\pm 0.001$ & 0.130 & 0.120 & 0.152 \\
 &  & M & 0.140 $\pm 0.002$ & \textbf{0.139} $\pm 0.003$ & 0.171 & 0.141 & 0.146 \\
\midrule
\multirow{15}{*}{\model{N-BEATS}} & \multirow{3}{*}{M1} & Y & 0.120 $\pm 0.005$ & \textbf{0.118} $\pm 0.007$ & 0.219 & 0.144 & 0.204 \\
 &  & Q & \textbf{0.093} $\pm 0.002$ & 0.094 $\pm 0.002$ & 0.139 & 0.094 & 0.106 \\
 &  & M & \textbf{0.082} $\pm 0.001$ & 0.083 $\pm 0.002$ & 0.144 & 0.088 & 0.086 \\
 & \multirow{3}{*}{M3} & Y & \textbf{0.104} $\pm 0.003$ & 0.113 $\pm 0.009$ & 0.105 & 0.093 & 0.100 \\
 &  & Q & \textbf{0.081} $\pm 0.001$ & \textbf{0.081} $\pm 0.001$ & 0.092 & 0.079 & 0.081 \\
 &  & M & \textbf{0.082} $\pm 0.000$ & \textbf{0.082} $\pm 0.000$ & 0.111 & 0.089 & 0.088 \\
 & \multirow{6}{*}{M4} & Y & \textbf{0.119} $\pm 0.001$ & 0.126 $\pm 0.010$ & 0.146 & 0.118 & 0.152 \\
 &  & Q & \textbf{0.098} $\pm 0.000$ & \textbf{0.098} $\pm 0.000$ & 0.118 & 0.097 & 0.100 \\
 &  & M & \textbf{0.077} $\pm 0.000$ & \textbf{0.077} $\pm 0.000$ & 0.102 & 0.080 & 0.080 \\
 &  & W & \textbf{0.062} $\pm 0.002$ & \textbf{0.062} $\pm 0.001$ & 0.066 & 0.064 & 0.064 \\
 &  & D & \textbf{0.026} $\pm 0.000$ & \textbf{0.026} $\pm 0.000$ & 0.027 & 0.026 & 0.027 \\
 &  & H & 0.047 $\pm 0.001$ & \textbf{0.046} $\pm 0.002$ & 0.039 & 0.042 & 0.050 \\
 & \multirow{3}{*}{Tourism} & Y & \textbf{0.142} $\pm 0.005$ & 0.153 $\pm 0.003$ & 0.125 & 0.138 & 0.168 \\
 &  & Q & \textbf{0.115} $\pm 0.002$ & 0.116 $\pm 0.002$ & 0.130 & 0.120 & 0.152 \\
 &  & M & \textbf{0.138} $\pm 0.002$ & \textbf{0.138} $\pm 0.002$ & 0.171 & 0.141 & 0.146 \\
\midrule
\multirow{15}{*}{\PatchTST} & \multirow{3}{*}{M1} & Y & \textbf{0.116} $\pm 0.010$ & 0.117 $\pm 0.010$ & 0.219 & 0.144 & 0.204 \\
 &  & Q & 0.100 $\pm 0.005$ & \textbf{0.097} $\pm 0.001$ & 0.139 & 0.094 & 0.106 \\
 &  & M & \textbf{0.103} $\pm 0.002$ & 0.104 $\pm 0.005$ & 0.144 & 0.088 & 0.086 \\
 & \multirow{3}{*}{M3} & Y & \textbf{0.165} $\pm 0.015$ & 0.185 $\pm 0.031$ & 0.105 & 0.093 & 0.100 \\
 &  & Q & \textbf{0.090} $\pm 0.003$ & 0.092 $\pm 0.002$ & 0.092 & 0.079 & 0.081 \\
 &  & M & \textbf{0.092} $\pm 0.001$ & \textbf{0.092} $\pm 0.001$ & 0.111 & 0.089 & 0.088 \\
 & \multirow{6}{*}{M4} & Y & \textbf{0.139} $\pm 0.006$ & 0.143 $\pm 0.012$ & 0.146 & 0.118 & 0.152 \\
 &  & Q & \textbf{0.110} $\pm 0.003$ & \textbf{0.110} $\pm 0.003$ & 0.118 & 0.097 & 0.100 \\
 &  & M & \textbf{0.084} $\pm 0.001$ & \textbf{0.084} $\pm 0.001$ & 0.102 & 0.080 & 0.080 \\
 &  & W & \textbf{0.072} $\pm 0.001$ & \textbf{0.072} $\pm 0.002$ & 0.066 & 0.064 & 0.064 \\
 &  & D & \textbf{0.034} $\pm 0.001$ & \textbf{0.034} $\pm 0.001$ & 0.027 & 0.026 & 0.027 \\
 &  & H & 0.120 $\pm 0.015$ & \textbf{0.117} $\pm 0.029$ & 0.039 & 0.042 & 0.050 \\
 & \multirow{3}{*}{Tourism} & Y & \textbf{0.133} $\pm 0.019$ & 0.142 $\pm 0.021$ & 0.125 & 0.138 & 0.168 \\
 &  & Q & \textbf{0.173} $\pm 0.004$ & 0.174 $\pm 0.005$ & 0.130 & 0.120 & 0.152 \\
 &  & M & \textbf{0.186} $\pm 0.011$ & \textbf{0.186} $\pm 0.009$ & 0.171 & 0.141 & 0.146 \\
\midrule
\multirow{15}{*}{\model{TCN}} & \multirow{3}{*}{M1} & Y & \textbf{0.111} $\pm 0.015$ & 0.117 $\pm 0.010$ & 0.219 & 0.144 & 0.204 \\
 &  & Q & 0.094 $\pm 0.003$ & \textbf{0.092} $\pm 0.002$ & 0.139 & 0.094 & 0.106 \\
 &  & M & 0.096 $\pm 0.003$ & \textbf{0.094} $\pm 0.001$ & 0.144 & 0.088 & 0.086 \\
 & \multirow{3}{*}{M3} & Y & \textbf{0.123} $\pm 0.009$ & 0.135 $\pm 0.010$ & 0.105 & 0.093 & 0.100 \\
 &  & Q & \textbf{0.083} $\pm 0.002$ & 0.085 $\pm 0.002$ & 0.092 & 0.079 & 0.081 \\
 &  & M & 0.091 $\pm 0.000$ & \textbf{0.089} $\pm 0.000$ & 0.111 & 0.089 & 0.088 \\
 & \multirow{6}{*}{M4} & Y & \textbf{0.133} $\pm 0.005$ & 0.137 $\pm 0.005$ & 0.146 & 0.118 & 0.152 \\
 &  & Q & \textbf{0.101} $\pm 0.001$ & 0.104 $\pm 0.002$ & 0.118 & 0.097 & 0.100 \\
 &  & M & 0.084 $\pm 0.000$ & \textbf{0.083} $\pm 0.001$ & 0.102 & 0.080 & 0.080 \\
 &  & W & \textbf{0.061} $\pm 0.000$ & 0.062 $\pm 0.001$ & 0.066 & 0.064 & 0.064 \\
 &  & D & \textbf{0.026} $\pm 0.000$ & \textbf{0.026} $\pm 0.000$ & 0.027 & 0.026 & 0.027 \\
 &  & H & \textbf{0.043} $\pm 0.001$ & 0.044 $\pm 0.001$ & 0.039 & 0.042 & 0.050 \\
 & \multirow{3}{*}{Tourism} & Y & \textbf{0.159} $\pm 0.014$ & 0.161 $\pm 0.023$ & 0.125 & 0.138 & 0.168 \\
 &  & Q & \textbf{0.123} $\pm 0.007$ & 0.126 $\pm 0.009$ & 0.130 & 0.120 & 0.152 \\
 &  & M & 0.176 $\pm 0.009$ & \textbf{0.172} $\pm 0.007$ & 0.171 & 0.141 & 0.146 \\
\midrule
\multirow{15}{*}{\model{LSTM}} & \multirow{3}{*}{M1} & Y & \textbf{0.121} $\pm 0.020$ & 0.125 $\pm 0.012$ & 0.219 & 0.144 & 0.204 \\
 &  & Q & \textbf{0.098} $\pm 0.002$ & 0.100 $\pm 0.001$ & 0.139 & 0.094 & 0.106 \\
 &  & M & 0.096 $\pm 0.002$ & \textbf{0.095} $\pm 0.001$ & 0.144 & 0.088 & 0.086 \\
 & \multirow{3}{*}{M3} & Y & 0.142 $\pm 0.007$ & \textbf{0.136} $\pm 0.014$ & 0.105 & 0.093 & 0.100 \\
 &  & Q & \textbf{0.089} $\pm 0.003$ & \textbf{0.089} $\pm 0.003$ & 0.092 & 0.079 & 0.081 \\
 &  & M & \textbf{0.098} $\pm 0.000$ & \textbf{0.098} $\pm 0.000$ & 0.111 & 0.089 & 0.088 \\
 & \multirow{6}{*}{M4} & Y & 0.140 $\pm 0.005$ & \textbf{0.139} $\pm 0.005$ & 0.146 & 0.118 & 0.152 \\
 &  & Q & \textbf{0.106} $\pm 0.000$ & \textbf{0.106} $\pm 0.001$ & 0.118 & 0.097 & 0.100 \\
 &  & M & \textbf{0.088} $\pm 0.000$ & \textbf{0.088} $\pm 0.000$ & 0.102 & 0.080 & 0.080 \\
 &  & W & \textbf{0.064} $\pm 0.001$ & \textbf{0.064} $\pm 0.001$ & 0.066 & 0.064 & 0.064 \\
 &  & D & \textbf{0.029} $\pm 0.000$ & \textbf{0.029} $\pm 0.000$ & 0.027 & 0.026 & 0.027 \\
 &  & H & \textbf{0.056} $\pm 0.005$ & 0.057 $\pm 0.003$ & 0.039 & 0.042 & 0.050 \\
 & \multirow{3}{*}{Tourism} & Y & 0.169 $\pm 0.022$ & \textbf{0.161} $\pm 0.006$ & 0.125 & 0.138 & 0.168 \\
 &  & Q & \textbf{0.123} $\pm 0.004$ & 0.126 $\pm 0.006$ & 0.130 & 0.120 & 0.152 \\
 &  & M & \textbf{0.168} $\pm 0.002$ & 0.170 $\pm 0.005$ & 0.171 & 0.141 & 0.146 \\
\bottomrule
\end{tabular}
\end{table}

\clearpage
\section{Masked Reconstruction with MOMENT}
\label{apd:moment-reconstruction}

\begin{figure}[h]
    \centering
    \includegraphics[width=0.55\linewidth]{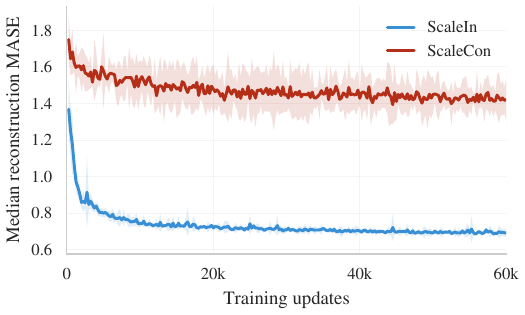}
    \caption{MOMENT reconstruction on held-out GiftEvalPretrain series. Plotted are average per-seed median dataset MASE, with $95\%$ Student-$t$ intervals across 4 seeds.}
    \label{fig:moment-reconstruction}
\end{figure}

We also compare the two loss spaces for masked reconstruction task using \Moment~\citep{goswami2024moment}. The matched runs use $512$-point input, a $30\%$ masking ratio, $60{,}000$ steps, and a batch size of $512$.  The results are averaged across 4 seeds.  Evaluation covers held-out series from GiftEvalPretrain datasets. We measure MAE on masked positions divided by the input window's seasonal-naive MAE (reconstruction MASE). After averaging each dataset across seeds, \ours lowers error on all datasets and reduces median reconstruction MASE from $1.42$ to $0.69$ ($51\%$). Figure~\ref{fig:moment-reconstruction} shows that the advantage is large and persists throughout training.

\end{document}